\documentclass[11pt]{article}

\usepackage[final]{acl}

\usepackage{times}
\usepackage{latexsym}

\usepackage[T1]{fontenc}

\usepackage[utf8]{inputenc}

\usepackage{microtype}
\usepackage{flafter}

\usepackage{inconsolata}
\usepackage{dsfont}
\usepackage{amsmath}
\usepackage{stfloats}

\usepackage{graphicx}
\usepackage{amssymb}
\usepackage{booktabs}
\usepackage{booktabs,multirow,amsmath,amssymb,pifont}

\usepackage{multirow}
\usepackage{booktabs}
\usepackage{xcolor, colortbl}
\usepackage[dvipsnames]{xcolor} 
\usepackage{subcaption}

\usepackage{enumitem}
\definecolor{Gray}{gray}{0.85}
\newcolumntype{g}{>{\columncolor{Gray}}c}
\usepackage[most]{tcolorbox}
\usepackage{orcidlink}

\usepackage{soul}

\newtcolorbox{prompt}[2][]{
    colback=black!3,
    colframe=black,
    boxrule=0.5pt,
    left=8pt,
    right=8pt,
    top=8pt,
    bottom=8pt,
    arc=2pt,
    breakable,
    enhanced,
    before skip=10pt,
    after skip=10pt,
    title={#2}, 
    #1 
}

\newtcolorbox{example}[2][]{
    colback=black!3,
    colframe=black!20,
    boxrule=0.5pt,
    left=8pt,
    right=8pt,
    top=8pt,
    bottom=8pt,
    arc=2pt,
    breakable,
    enhanced,
    before skip=10pt,
    after skip=10pt,
    coltitle=black,
    title={#2}, 
    #1 
}

\title{\textsc{Oblivion}: Self-Adaptive Agentic Memory Control through\\Decay-Driven Activation}

\author{Ashish Rana\textsuperscript{1}\thanks{Equal contributions.}, Chia-Chien Hung\textsuperscript{1}\footnotemark[1], Qumeng Sun\textsuperscript{1,2}\footnotemark[1], \\
\textbf{Julian Martin Kunkel\textsuperscript{2}} \textbf{, and Carolin Lawrence\textsuperscript{1}} \\
\textsuperscript{1}NEC Laboratories Europe, Heidelberg, Germany\\
\textsuperscript{2}GWDG, Georg-August-Universität Göttingen, Göttingen, Germany\\
  \texttt{\{Ashish.Rana, Chia-Chien.Hung, Carolin.Lawrence\}@neclab.eu}\\
  \texttt{sun.qumeng@stud.uni-goettingen.de,} 
  \texttt{julian.kunkel@gwdg.de}\\
  }

\begin{document}
\maketitle

\begin{abstract}

Human memory adapts through selective forgetting: experiences become less accessible over time but can be reactivated by reinforcement or contextual cues. In contrast, memory-augmented LLM agents rely on ``always-on'' retrieval and ``flat'' memory storage, causing high interference and latency as histories grow. We introduce \textsc{Oblivion}, a memory control framework that casts forgetting as \emph{decay-driven reductions in accessibility} -- not explicit deletion. \textsc{Oblivion} decouples memory control into read and write paths. The read path decides \textit{when to consult memory}, based on agent uncertainty and memory buffer utility, avoiding redundant always-on access. The write path decides \textit{what to strengthen}, by reinforcing memories contributing to forming the response. Together, this enables hierarchical memory organization that maintains persistent high-level strategies while dynamically loading details as needed.
We evaluate on both static and dynamic long-horizon interaction benchmarks. \textsc{Oblivion} outperforms both direct and memory-augmented baselines, while reducing token cost by up to 73\% at 120K interaction spans. These results show that treating memory as a control problem -- deciding when to retrieve and what to reinforce -- is essential for sustaining long-horizon agent performance.

\end{abstract}

\section{Introduction}
Forgetting is not a failure of memory, but a functional necessity for efficient cognition and adaptive behavior~\citep{norby2015forget, wimber2015retrieval, fawcett2020many}. In cognitive psychology, forgetting is characterized as a \textit{loss of accessibility} rather than permanent erasure: as time passes, memory traces decay unless reinforced by repetition or contextual cues \citep{berens-etal-2020-dissociating}. Ebbinghaus' foundational work established the ``forgetting curve'': repeated retrievals and selective retentions popularized the observation that recall probability decays gradually over time ~\citep{ebbinghaus-1885-gedachtnis, ebbinghaus-1913-memory}. Viewed as a control mechanism, forgetting acts as a dynamic filter:  the human brain prioritizes its limited ``working memory'' for high-utility information while keeping less relevant traces latent but reactivatable.


\begin{figure}[t]
	\centering
    \includegraphics[trim={6.20cm 3.5cm 6.20cm 3.9cm},clip,width=0.48\textwidth]{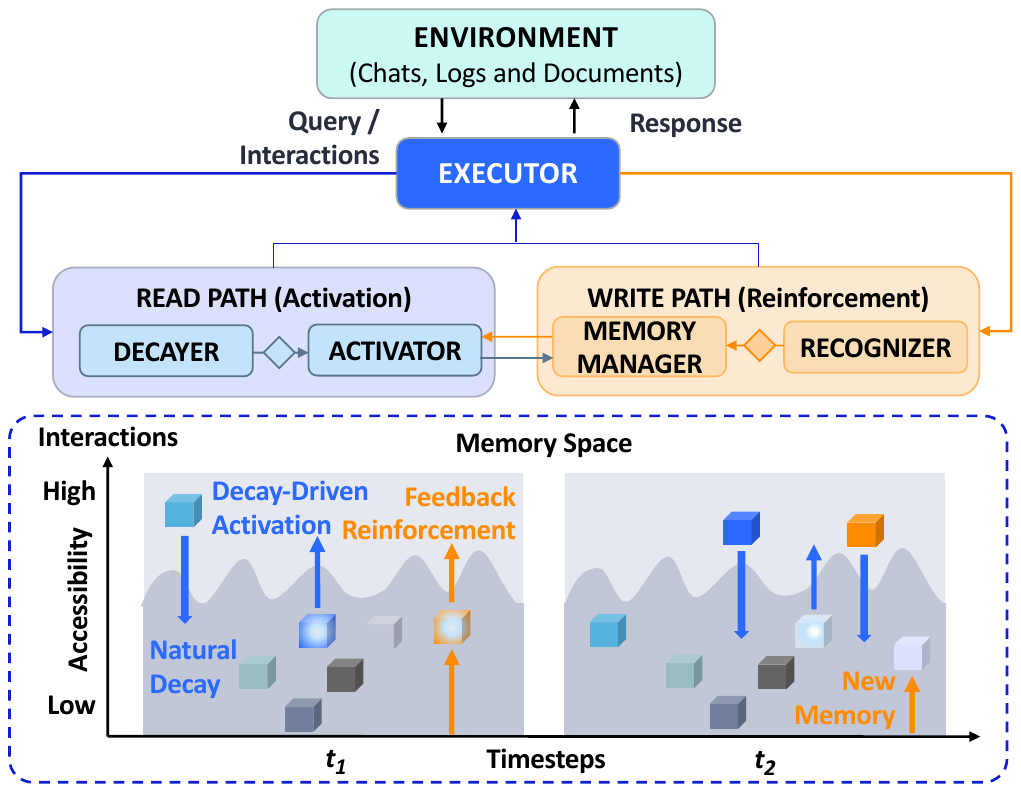}
    \caption{\textsc{Oblivion} facilitates memory-augmented agents by decay-driven activation over hierarchical memory traces. The \textbf{Executor} orchestrates the \textbf{\textcolor{blue}{read}} path for uncertainty-gated retrieval (\textbf{\textcolor{blue}{$\diamond$}}); and the \textbf{\textcolor{orange}{write}} path for feedback-driven updates (\textbf{\textcolor{orange}{$\diamond$}}), enabling dynamic control over memory activation.}
    \label{fig:overview_oblivion}
\end{figure}


To extend reasoning beyond context limits, memory-augmented LLM agents maintain external memory storage that can be accessed and updated across interactions~\citep{zhang2025survey, jia2026aihippocampus}. Yet as memory stores grow, agents face a severe ``needle-in-a-haystack'' problem: relevant evidence is obscured by accumulating stale or tangential entries, increasing prompt pollution and reasoning interference~\citep{liu-etal-2024-lost, xiong2025memory}. Existing methods exhibit three critical limitations: (1) \textit{uncertainty-decoupled retrieval}, where agents perform ``always-on'' retrieval, fetching information at every conversation turn regardless of necessity~\citep{xiong2025memory,zhang2026agentic}; (2) \textit{``flat'' memory}, where uniform storage fails to distinguish persistent high-level strategies from transient low-level details and often resorts to hard deletion to manage context limits~\citep{chhikara2025mem0,fang2025lightmem}; and (3) \textit{static evaluation bias},
where existing memory agent benchmarks emphasize retrieval accuracy in isolation, underrepresenting long-horizon interactions with compounding memory decisions and shifting task demands~\citep{wu-etal-2025-longmemeval}.



We bridge these gaps with \textsc{Oblivion}, a self-adaptive memory control framework that shifts the paradigm from \textit{memory as a database} to \textit{memory as a control} problem under long-horizon interaction. 
The key insight is \textit{read/write decoupling}~\citep{mcclelland-etal-1995-complementary,craik-etal-1996-effects} (\autoref{fig:overview_oblivion}). The \textit{read path} decides \emph{when to consult memory} using uncertainty signals and available context, avoiding redundant always-on access. The \textit{write path} decides \emph{what to strengthen} by reinforcing only response-contributing memories. As a result, unused memory traces fade without explicit deletion yet remain reactivable when conditions change. 

\textsc{Oblivion} features memory control through a hierarchical memory representation managed by a central orchestrator (\textbf{Executor}), controlling a self-adaptive read/write loop. On the \textit{read path}, a \textbf{Decayer} gates retrieval by uncertainty and filters accessible memory traces with an Ebbinghaus-inspired retention score, and an \textbf{Activator} triggers targeted retrieval (i.e., when uncertainty is high) via a query-expansion graph. On the \textit{write path}, a \textbf{Recognizer} identifies which memories contributed to the response, and a \textbf{Memory Manager} selectively reinforces those memories. Together, these components implement decay-driven activation for long-horizon agentic memory control.
\paragraph{Contributions.} (i) We formulate long-horizon agent memory as a \textit{control problem} and introduce read/write decoupling to separate \emph{when to access} memory (progressive retrieval) from \emph{what to preserve} (selective reinforcement). (ii) We propose \textsc{Oblivion}, a framework that combines a hierarchical memory structure with an orchestrated control loop, enabling decay-driven activation over long interaction histories. (iii)
We evaluate \textsc{Oblivion} across multiple tasks on static QA (\textsc{LongMemEval}, $\approx$100K context;~\citet{wu-etal-2025-longmemeval}) and dynamic interaction (\textsc{GoodAILTM}, $\approx$120K context;~\citet{bolado-etal-2024-goodailtm}), consistently outperforming both direct and memory-augmented baselines, highlighting the importance of memory control in long-horizon scenarios. (iv) We further analyze memory decay dynamics and operational efficiency at scale, providing empirical insight into how memory control governs long-horizon agent behavior and demonstrating scalability through up to 73\% token-cost reduction at 120K interaction spans. (v) The code and resources are publicly available to support reproducibility and future research.\footnote{\url{https://github.com/nec-research/oblivion}}

\section{\textsc{Oblivion} Overview}
\textsc{Oblivion} addresses long-horizon agentic interactions through \textit{read/write-decoupled memory control}: deciding when to consult memory and which memory traces to activate (\textit{read path}), and what to strengthen by reinforcing only response-contributing memories (\textit{write path}). As shown in~\autoref{fig:overview_oblivion}
(detailed in~\autoref{app:arch_oblivion}), the framework comprises a hierarchical memory representation, a read path (\textbf{Decayer} and \textbf{Activator}), and a write path (\textbf{Recognizer} and \textbf{Memory Manager}), orchestrated by \textbf{Executor}. In the following, we introduce the memory representation (\autoref{subsec:memory_representation}), the read path for uncertainty-gated retrieval (\autoref{subsec:decayer_activator}), the write path for feedback-driven updates (\autoref{subsec:recognizer_manager}), and the self-adaptive memory control loop (\autoref{subsec:executor}).

\subsection{Hierarchical Memory Representation}
\label{subsec:memory_representation}

Inspired by the working memory construct in cognitive psychology~\citep{baddeley-2000-episodic-buffer} and its computational realization for language agents~\citep{sumers-etal-2024-coala}, \textsc{Oblivion} separates memory into a transient Working Memory (WM) and a Persistent Store ($\mathcal{D}$), which consists of three memory types rooted in cognitive distinctions~\citep{tulving-1972-episodic-semantic, squire-2004-memory-systems}: Dynamic Procedural Memory ($L_1$), Semantic Memory ($L_2$), and Preemptive Episodic Memory ($L_3$). WM acts as a bounded cache, comprising: (1) Interaction Context ($H_t$), a sliding window of the most recent $K$ raw turns; (2) Memory Buffer ($B_t$), which permanently holds $L_1$ cluster nodes to guide routing while dynamically loading and evicting $L_2$/$L_3$ items retrieved from $\mathcal{D}$;\footnote{$L_1$ resides solely in $B_t$, while $L_2$ and $L_3$ are persisted to $\mathcal{D}$ and loaded on demand.} and (3) Task Metadata ($I_{\text{task}}$). 

\paragraph{Dynamic Procedural Memory ($L_1$).}
Serving as the entry point, $L_1$ is an adaptive meta-index over memory. It is initialized with a configurable domain-specific taxonomy that can range from a handful of broad categories to a dozen task-oriented clusters and can be expanded emergently through interactions.
Each cluster node is permanently resident in $B_t$ and contains:
    (i) Cluster Summary ($s_c$): a compressed natural-language synopsis enabling relevance assessment without full retrieval;
    (ii) Cluster Statistics ($\alpha_c$): utility and access-frequency signals that govern the cluster's decay-driven accessibility over time; and
    (iii) Procedural Memory ($\pi_c$): meta-instructions learned from interaction feedback that capture \textit{how} to handle queries within the cluster's domain -- not just \textit{what} it contains -- transforming $L_1$ from a passive coarse index into adaptive meta-cognitive strategies.


\paragraph{Semantic Memory ($L_2$).}
$L_2$ stores discrete, timestamped factual statements. To mitigate the ``world drift'' problem~\citep{dhingra-etal-2022-time},\footnote{We use ``world drift'' to refer to temporal inconsistency in stored facts as the world or user state changes over time.} timestamps allow the system to resolve temporal conflicts by prioritizing currently valid facts over outdated ones during curation.

\paragraph{Preemptive Episodic Memory ($L_3$).}
$L_3$ stores transformed fixed-length interaction episodes by unifying retrospective traces with interaction feedback, forming actionable preemptive entries. For example, ``Tuesday was a holiday; I went to the gym and liked it.'' (Standard) vs. ``On holidays, I like to go to the gym for recreation and maintaining personal health; Remember this preference and accordingly schedule future plans.'' (Preemptive). $L_3$ is loaded into $B_t$ from $\mathcal{D}$ when distant and granular past interactions are required. In addition, $L_2$--$L_3$ memory linking is possible, allowing cross-type evidence discovery at retrieval time.


\subsection{Read Path: Uncertainty-Gated Retrieval}
\label{subsec:decayer_activator}
The read path first evaluates available memory context and decides whether the current buffer $B_t$ suffices for the query $q_t$ at turn $t$; otherwise, it triggers targeted retrieval to load extra evidence from $\mathcal{D}$. The Decayer performs two checks --- \textit{uncertainty estimation} (Whether to retrieve?) and \textit{retention scoring} (Which memories remain accessible?) --- that jointly gate retrieval: when triggered, the Activator fetches additional $L_2$/$L_3$ entries from $\mathcal{D}$.

\paragraph{Decayer.}
For each $L_1$ cluster $c$ in $B_t$, the Decayer estimates an uncertainty score $u_t(c) \in [0,1]$ (higher means weaker support for $q_t$) using two complementary signals: (i) LLM-as-a-judge~\citep{zheng-etal-2023-llm-judge}: an LLM, conditioned on the cluster summary, procedural memory, and any cached $L_2$/$L_3$ items, assesses whether $c$ provides sufficient evidence for $q_t$.
(ii) Embedding-based: the mean cosine similarity between the query embedding and embeddings of cached buffer items is computed; low similarity signals that the buffer content is likely irrelevant.
Activation is always triggered when $B_t$ contains no $L_2$/$L_3$ items. Otherwise, the gating threshold adapts across rounds: in the first round, \emph{either} signal alone suffices (OR), favoring recall over precision; in subsequent rounds, \emph{both} must agree (AND), preventing redundant retrieval once the buffer is already populated. This dual-signal, recall-favoring schedule reduces reliance on any single imperfect estimate by construction: no single LLM judgment determines retrieval alone, reducing the likelihood that an isolated mis-estimate directly suppresses potentially useful evidence.

While uncertainty gates retrieval, a complementary retention score gates cluster accessibility. For each cluster $c$, the Decayer computes $R_t(c)$ via an interaction-based forgetting curve~\citep{ebbinghaus-1885-gedachtnis, wozniak-etal-1995-two}:\vspace{-0.5em}

{\footnotesize
\begin{equation*}
R_t(c) = \exp\!\left(-\frac{n_t(c)}{S_t(c)}\right), \quad
S_t(c) = \bigl(U_t(c) + F_t(c) + \epsilon\bigr)\cdot T,
\label{eq:oblivion_retention}
\end{equation*}}
\normalsize
\noindent where $n_t(c)$ is the number of turns since cluster $c$ was last accessed, $U_t(c) \in [0,1]$ is a learned utility proxy, $F_t(c) \in [0,1]$ is an access-frequency proxy, $\epsilon > 0$ prevents division by zero, and $T > 0$ is a decay temperature that scales stability (higher $T$ yields slower decay). Frequently used or high-utility clusters decay slowly; neglected ones see $R_t(c)$ decline, making their memories less prominent in retrieval ranking and more likely to be evicted during curation -- yet never deleted from $\mathcal{D}$, preserving reactivatability when future interactions reinforce them.

\paragraph{Activator.} When triggered, the Activator reduces uncertainty by retrieving $L_2$/$L_3$ entries from $\mathcal{D}$ and updating $B_t$ under fixed cache budgets. Instead of a single vector search, it decomposes $q_t$ into a set of focused sub-queries organized as a directed acyclic graph (DAG): $\mathcal{G}_t=(V,E)$~\citep{khot2023decomposedpromptingmodularapproach, zhou2023leasttomostpromptingenablescomplex}. Each node specifies (i)~a sub-query string, and (ii)~a memory type ($L_2$ or $L_3$); edges encode dependencies (e.g., refinement), supporting multi-step information needs beyond single-shot retrieval. Each sub-query performs global similarity search over $\mathcal{D}$ --- not hard-filtered by cluster --- preserving cross-cluster evidence discovery. The main query additionally follows $L_2$--$L_3$ links to retrieve associated memories. Retrieved candidates are curated into $B_t$ by retaining useful cached items, adding high-coverage new ones, and evicting stale or low-retention entries to stay within budget. The updated $B_t$ returns to the Decayer, and the Executor repeats the Decayer$\rightarrow$Activator loop for fixed iterations, stopping once uncertainty falls.

\subsection{Write Path: Feedback-Driven Updates}
\label{subsec:recognizer_manager}
The write path enables continual learning~\citep{de-lange-etal-2022-continual-learning-survey} by extracting memories from each interaction and selectively reinforcing those contributing to the response.

\paragraph{Recognizer.} At the end of each turn, the Recognizer: (i)~extracts new $L_2$ factual memories and accumulates episodic\footnote{$L_3$ stores \textit{preemptive episodic} entries --- interaction traces transformed with actionable framing (\autoref{subsec:memory_representation}); we use ``episodic'' as shorthand throughout.} entries toward $L_3$ episodes; (ii)~constructs or updates $L_1$ cluster topics, including procedural memory, based on observed interaction patterns; and (iii)~performs \textit{credit assignment} by identifying which specific memory entries in $B_t$ were used to generate the response. This conservative policy ties reinforcement strictly to usage, avoiding indiscriminate re-scoring of all retrieved memories.


\paragraph{Memory Manager.}
The Memory Manager persists new entries to $\mathcal{D}$ and updates the utility metadata of used memories. The utility score $U_t(c)$ for accessed clusters is reinforced, while unaccessed clusters naturally decay via the increment of $n_t(c)$ in~\autoref{eq:oblivion_retention}. This implements decay-driven forgetting: memories that are not reinforced gradually become inaccessible --- not through deletion, but through declining retention and lower retrieval priority --- yet remain recoverable when future conditions change.


\subsection{Executor: Self-Adaptive Control Loop}
\label{subsec:executor}
Serving as the orchestrator of \textsc{Oblivion}, the Executor maintains the working memory state and coordinates the read/write memory control cycle. At each turn $t$, it: (i)~updates WM with the new query $q_t$; (ii)~iterates the Decayer$\rightarrow$Activator loop until the buffer is sufficient or a maximum iteration limit is reached; (iii) generates the response from the curated working memory state, and (iv) invokes the Recognizer to distill new memory candidates from the interaction and Memory Manager persists these updates to $\mathcal{D}$. This self-adaptive control loop allows the agent to learn what to remember over time -- the working memory (WM) stays bounded to minimize interference, while decay/activation (\textit{read}) and reinforcement (\textit{write}) progressively increase the accessibility of high-utility memories~\citep{mcclelland-etal-1995-complementary, craik-etal-1996-effects}.

\section{Experimental Setup}
\subsection{Memory Benchmarks}
\label{subsec:memory_benchmark}
We evaluate \textsc{Oblivion} on challenging benchmarks: \textsc{LongMemEval}~\citep{wu-etal-2025-longmemeval} for static retrieval-heavy QA over long chats, and \textsc{GoodAILTM}~\citep{bolado-etal-2024-goodailtm} for dynamic long-horizon interactions. Together, this mitigates a common limitation in prior evaluations, focusing on either static long-context QA or a single interactive memory setting, but rarely both.

\textsc{LongMemEval} formulates memory as QA over a multi-session conversational \textit{haystack}. Each instance contains (i) a question, (ii) a reference answer, and (iii) a timestamped set of prior chat sessions in which evidence may exist among many irrelevant sessions. The benchmark comprises 500 human-curated questions and synthetic sessions across 6 question types.\footnote{\textbf{SS-A} (Single Session (SS)-Assistant), \textbf{SS-U} (SS-User), \textbf{SS-P} (SS-Preference), \textbf{MR} (Multi-Session Reasoning), \textbf{TR} (Temporal Reasoning), \textbf{KU} (Knowledge Update).} We follow two standard context settings: \textbf{Oracle}, where the model receives only the evidence sessions (upper bound with perfect retrieval), and \textbf{S}, where the model must identify relevant evidence within the full noisy history.
\textsc{GoodAILTM} evaluates long-term memory in an interactive setting, measuring how well an agent can \textit{retain}, \emph{revise}, and \textit{update} information online over long, ongoing conversations--not just answer isolated questions. The benchmark induces memory pressure beyond the context window. The benchmark includes 11 scenarios (3 instances each; 33 tests total) spanning 7 memory skills,\footnote{Recall, Conflicting Information, Episodic Memory, Spatial Memory, Prospective Memory, Theory of Mind, Information Integration.} with each scenario requiring the agent to pursue multiple concurrent tasks spanning distinct memory skills within a single long-horizon interaction.
We report results under \textbf{Isolated} (no task interleaving) and \{\textbf{2K}, \textbf{32K}, \textbf{120K}\} (task interleaving, with given memory span-limited constraints).
For detailed dataset statistics and examples refer to~\autoref{app:memory_benchmark}.

\setlength{\tabcolsep}{3.6pt}

\begin{table*}[t]
\centering
\scriptsize

\begin{tabular}{llccccccgccccccg}
\toprule
& & \multicolumn{7}{c}{\textbf{Oracle}} & \multicolumn{7}{c}{\textbf{S}} \\
\cmidrule(lr){3-9}\cmidrule(lr){10-16}
\textbf{Model} & \textbf{Method} &
\textbf{SS-A} & \textbf{SS-U} & \textbf{SS-P} & \textbf{MR} & \textbf{KU} & \textbf{TR} & \textbf{AVG} &
\textbf{SS-A} & \textbf{SS-U} & \textbf{SS-P} & \textbf{MR} & \textbf{KU} & \textbf{TR} & \textbf{AVG} \\
\midrule

\multirow{3}{*}{\textbf{GPT-4o-mini}}
& LME-RFT (\textit{direct}) & 91.07 & 91.43 & 83.33 & 80.45 & \textbf{80.77} & 78.95 & 83.00 & 91.07 & 94.29 & 60.00 & 69.92 & \textbf{79.49} & 77.44 & 78.60 \\
& \textsc{EverMemOS} (\textit{memory}) & 67.86 & 93.81 & \textbf{96.67} & 68.17 & 78.21 & 59.90 & 72.80 & 55.36 & 91.43 & \textbf{93.33} & 60.40 & 67.95  & 53.38 & 65.47 \\\cmidrule(lr){2-16}
& OBLIVION (\textit{ours})   & \textbf{98.21} & \textbf{94.29} & 73.33 & \textbf{81.20} & \textbf{80.77} & \textbf{83.46} & \textbf{85.00} & \textbf{94.64} & \textbf{95.71} & 60.00 & \textbf{79.70} & 74.36 & \textbf{80.45} & \textbf{81.80} \\
\midrule
\multirow{3}{*}{\textbf{GPT-4.1-mini}}
& LME-RFT (\textit{direct}) & 91.07 & 94.29 & \textbf{100.00} & \textbf{90.98} & 92.31 & 89.47 & 91.80 & 91.07  & 97.14 & 93.33 & 81.20 & \textbf{93.59} & 87.97 & 89.00 \\
& \textsc{EverMemOS} (\textit{memory}) & 83.93  & \textbf{96.67} & 96.67 & 78.70 & 92.31 & 74.19 & 83.80 & 80.36  &  \textbf{100.00}  & \textbf{96.67}  & 76.94  & 87.18 &  69.17& 81.27 \\ \cmidrule(lr){2-16} 
& OBLIVION (\textit{ours})   & \textbf{98.21} & 95.71 & \textbf{100.00} & 90.23 & \textbf{96.15} & \textbf{93.98} & \textbf{94.40} & \textbf{94.64} & 95.71 & \textbf{96.67} & \textbf{84.96} & \textbf{93.59} & \textbf{88.72} & \textbf{90.60} \\
\bottomrule
\end{tabular}
\caption{\textsc{LongMemEval} results in accuracy (\%): \textbf{Oracle} (\textit{clean}) vs. \textbf{S} (\textit{noisy}) (detailed in~\autoref{app:lme_additional}). For \textsc{EverMemOS}~\citep{hu-etal-2026-evermemos}, we report only \texttt{GPT-4o-mini} and \texttt{GPT-4.1-mini} for close comparison.}
\label{tab:lme_mix_results}
\end{table*}
\paragraph{Evaluation Metrics.} For \textsc{LongMemEval}, we report QA accuracy (\%) for each task and micro-average across them. 
For \textsc{GoodAILTM}, each test yields a normalized score in $[0,1]$ from scenario-specific evaluators with LLM-as-a-judge. We summarize performance as the sum of per-scenario normalized scores for each (Isolated/2K/32K/120K), reporting the mean over three runs. 

\subsection{Models and Baselines}
\label{subsec:models_baselines}

We evaluate multiple LLM backbones spanning different scales (\texttt{Phi-4-mini-instruct-4B}, \texttt{Qwen3-30B-A3B-Instruct}, \texttt{GPT-4o-mini}, \texttt{GPT-4.1-mini}). Unless otherwise stated, we keep decoding settings and retrieved-context budget (e.g., number of iterations) fixed across all methods to isolate the effect of memory architectures. To enable a consistent comparison across benchmarks, we compare two baselines: \textit{direct} and \textit{memory}, alongside \textsc{Oblivion}. For \textit{direct}, this represents a minimal memory interface: the model answers in a single pass from the assembled context, \emph{without adaptive memory control} --- no decay-driven gating, no write path, and no selective reinforcement across turns. For \textsc{LongMemEval}, we follow the optimized retrieval pipeline~\citep{wu-etal-2025-longmemeval} \textsc{LME-RFT} (round-level storage, fact-augmented keys, time-aware query expansion, and flat retrieval) with direct reading (no Chain-of-Note; CoN).\footnote{CoN introduces a per-memory reasoning step independent of the memory architecture; the \textsc{LongMemEval} paper shows it can degrade accuracy for smaller LLMs, so we exclude it for fairer comparison.} \textsc{LME-RFT} already incorporates retrieval augmentation, establishing a stronger-than-vanilla-RAG upper bound for the \textit{direct} category. For \textsc{GoodAILTM}, \textsc{FullCTX} is a standard full-context agent that answers solely from its context window (up to 120K tokens), else truncated. For \textit{memory}, we select state-of-the-art memory-augmented agent systems that
extend beyond session-level retrieval by maintaining a persistent indexed store across interactions.
For \textsc{LongMemEval}, we use \textsc{EverMemOS}~\citep{hu-etal-2026-evermemos} as a strong memory-augmented baseline, which has been shown to outperform recent memory systems such as \textsc{MemOS}~\citep{li-etal-2025-memos}, \textsc{Mem0}~\citep{chhikara2025mem0}, \textsc{Zep}~\citep{rasmussen-etal-2025-zep}. It extracts interaction traces as structured facts and episodic memories, then performs memory-guided retrieval to assemble a compact context for downstream reasoning.
\footnote{We reproduced \textsc{EverMemOS} with standardized LLM calls, temperature settings, and context-window handling to enable controlled comparison; details in \autoref{app:lme_additional}.}
For \textsc{GoodAILTM}, we use \textsc{BeyondPrompts}~\citep{bolado-etal-2024-goodailtm}, a \textsc{MemGPT}-inspired~\citep{packer-etal-2023-memgpt} system with semantic vector memory and a persistent user-information scratchpad. Both \textit{memory} baselines lack decay-driven gating and selective write-path reinforcement, which are the core mechanisms \textsc{Oblivion} introduces. For \textsc{Oblivion} (\textit{ours}), it stands for our self-adaptive memory control framework with a decay-driven memory activation mechanism.
All embedding-based retrieval uses \textsc{text-embedding-3-small}.

\section{Evaluation Results}
\subsection{Static Multi-Session QA}
\label{subsec:static_qa}

As shown in \autoref{tab:lme_mix_results}, \textsc{Oblivion} generally exhibits stronger performance than \textsc{LME-RFT} (\textit{direct}) and \textsc{EverMemOS} (memory) on static multi-session QA. The \textit{direct} \textsc{LME-RFT} baseline, utilizing flat retrieval, achieves an average accuracy of 89.00\% (S) with \texttt{GPT-4.1-mini}. Conversely, the \textsc{EverMemOS} baseline shows inconsistent results; while reaching 81.27\% on \texttt{GPT-4.1-mini}, it drops to 65.47\% with \texttt{GPT-4o-mini}. This performance gap is most pronounced in MR and TR tasks, where \textsc{EverMemOS} scores 60.40\% and 53.38\% for \texttt{GPT-4o-mini}, suggesting its memory compression may discard temporal details during complex retrievals.
In comparison, \textsc{Oblivion} with hierarchical memory control and read/write decoupled loop for \texttt{GPT-4.1-mini} attains average accuracy gains from 91.80\% to 94.40\% (\textbf{Oracle}) and from 89.00\% to 90.60\% (\textbf{S}). \textsc{Oblivion} particularly addresses baseline weaknesses in cross-session MR tasks with \texttt{GPT-4o-mini}: 69.92\% (\textit{direct}) vs. 79.70\% (\textit{ours}). These findings are consistent with the hypothesis that maintaining episodes in $L_3$ alongside factual $L_2$ semantic entries helps mitigate the ``needle-in-a-haystack''~\citep{liu-etal-2024-lost} problem. The modest gains on the realistic \textbf{S} setting are expected: \textsc{LongMemEval} frames memory as \emph{static} one-shot QA, where a single retrieval pass over a fixed history already approaches the achievable ceiling once a strong retrieval baseline (\textsc{LME-RFT}) is in place. Decay-driven memory control instead targets memory accessibility across \emph{evolving} interactions, which is a regime that static QA does not exercise.

\setlength{\tabcolsep}{1.25pt}
\begin{table}[t]
\centering
\scriptsize
\begin{tabular}{llcccc}
\toprule
& & \textbf{Isolated} & \textbf{2K} & \textbf{32K} & \textbf{120K} \\
\textbf{Model} & \textbf{Method} & \textbf{(11)} & \textbf{(10)} & \textbf{(11)} & \textbf{(11)} \\
\midrule
\multirow{3}{*}{\textbf{Phi-4-mini\dag}}
& \textsc{FullCTX} (\textit{direct}) & 4.44 & 4.28 & 1.33 & 1.15 \\
& \textsc{BeyondPrompts} (\textit{memory}) & 5.57 & 3.91 & 2.65 & 2.52 \\\cmidrule(lr){2-6}
& Oblivion (\textit{ours}) & \textbf{6.45} & \textbf{5.06} & \textbf{3.36} & \textbf{3.37} \\
\midrule
\multirow{3}{*}{\textbf{Qwen3-30B-A3B\dag}}
& \textsc{FullCTX} (\textit{direct}) & 6.49 & 5.94 & 5.38 & 4.25 \\
& \textsc{BeyondPrompts} (\textit{memory}) & 6.45 & 5.36 & 4.23 & 4.17 \\\cmidrule(lr){2-6}
& Oblivion (\textit{ours}) & \textbf{7.91} & \textbf{7.75} & \textbf{6.10} & \textbf{5.30} \\
\midrule
\multirow{3}{*}{\textbf{GPT-4o-mini}}
& \textsc{FullCTX} (\textit{direct}) & 7.22 & 6.14 & 5.65 & 4.81 \\
& \textsc{BeyondPrompts} (\textit{memory}) & 8.87 & 6.37 & 4.87 & 4.93 \\\cmidrule(lr){2-6}
& Oblivion (\textit{ours}) & \textbf{9.00} & \textbf{7.37} & \textbf{6.26} & \textbf{5.63} \\
\midrule
\multirow{3}{*}{\textbf{GPT-4.1-mini}}
& \textsc{FullCTX} (\textit{direct}) & 8.11 & 6.29 & 6.05 & 6.42 \\
& \textsc{BeyondPrompts} (\textit{memory}) & 8.53 & 5.37 & 6.68 & 6.26 \\\cmidrule(lr){2-6}
& Oblivion (\textit{ours}) & \textbf{9.20} & \textbf{8.41} & \textbf{6.79} & \textbf{6.70} \\
\bottomrule
\end{tabular}
\caption{Evaluation results of \textsc{GoodAILTM} reported as the normalized sum score, out of 11/10/11/11 for the Isolated/2K/32K/120K settings, respectively. \textbf{Bold} marks the best method per setting (detailed in~\autoref{app:goodailtm_additional}). \dag~Selected models are \texttt{instruct}-tuned variants.}
\label{tab:goodai_ltm_results}
\end{table}
\subsection{Dynamic Interactive Conversation}
\label{subsec:dynamic_conv}
Static QA bounds the benefit of memory control to a single retrieval pass (\autoref{subsec:static_qa}); dynamic interaction, by contrast, exercises it directly, requiring the agent to continually decide what to retrieve, reinforce, and let decay. For the dynamic \textsc{GoodAILTM} benchmark (\autoref{tab:goodai_ltm_results}), \textsc{Oblivion} outperforms both \textit{direct} and \textit{memory} baselines in all settings.
\footnote{Across all model$\times$setting, one-sided Wilcoxon signed-rank tests~\citep{wilcoxon1945individual} generally demonstrate statistically significant shifts with large Hedges' $g$~\cite{hedges1981distribution} effects.}
The \textit{direct} \textsc{FullCTX} baseline performs adequately in Isolated tasks but degrades significantly in the longer 32K and 120K settings, reflecting ``lost-in-the-middle'' effects~\citep{liu-etal-2024-lost}. While \textsc{BeyondPrompts} (\textit{memory}) improves in the Isolated setting, it often underperforms \textsc{FullCTX} in the interleaved 2K spans (e.g., 5.37 vs.\ 6.29 for \texttt{GPT-4.1-mini}), suggesting that a static retrieval buffer may interfere with the immediate context. \textsc{Oblivion} demonstrates superior stability across all spans: for \texttt{GPT-4.1-mini}, the score rises from 6.29 to 8.41 at 2K and from 6.05 to 6.79 at 32K over \textsc{FullCTX}, supporting the design of memory as a dynamic control problem in which decay-driven activation sustains performance where neither exhaustive retention nor static retrieval scales reliably. This effect is clearest on the smallest backbone: at 32K, \texttt{Phi-4-mini} collapses to a score of 1.33 under \textsc{FullCTX} but retains 3.36 with \textsc{Oblivion}, indicating that decay-driven control protects a low-capacity model from long-context degradation. At the longest 120K span, the accuracy gain narrows for the strongest backbone (\texttt{GPT-4.1-mini}), whose large native context window keeps \textsc{FullCTX} competitive; here the value of memory control shifts from accuracy to efficiency, as \textsc{Oblivion} attains comparable accuracy at a fraction of the token cost (\autoref{tab:efficiency_small}; detailed in~\autoref{subsec:performance_and_operatonal_analysis}). Overall, across all interaction spans, the trend indicates the effectiveness of the self-adaptive memory control loop for long-horizon dynamic interactions.

\section{Analysis}
%

\begin{figure*}[th]
	\centering
    \includegraphics[trim={0cm 0cm 0cm 0cm},clip,width=1.0\textwidth]{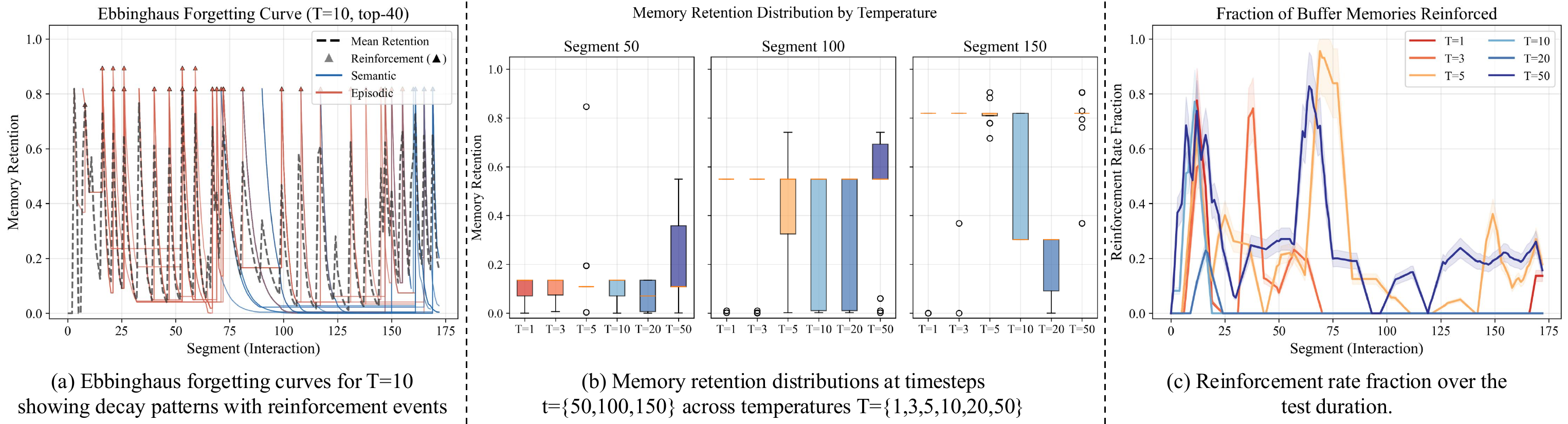}
    \caption{(a) Ebbinghaus forgetting curves showing decay patterns with reinforcement for 2K setting. (b) Retention at $t\in\{50,100,150\}$ over $T\in\{1,3,5,10,20,50\}$: low $T$ collapses, $T{=}10$ balances stability and maintenance. (c) Reinforcement fraction over time: $T{=}50$ saturates the buffer, recreating always-on retrieval.}
    
    \label{fig:decay_temperature_behavior_analysis}
\end{figure*}


We further study how memory decay dynamics (\autoref{subsec:memory_decay_behavior}), hierarchical structure (\autoref{subsec:hierarchical_memory_control}), episodic memory configuration (\autoref{subsec:preemptive_memory_analysis}), and operational efficiency (\autoref{subsec:performance_and_operatonal_analysis}) jointly shape the behavior and contribute to the overall performance of \textsc{Oblivion}.

\subsection{Memory Decay Behavior}
\label{subsec:memory_decay_behavior}


\autoref{fig:decay_temperature_behavior_analysis}(a) demonstrates that semantic ($L_2$) and episodic ($L_3$) memory follow a distinct sawtooth pattern. Here, retention drops rapidly due to exponential decay and spikes upon reinforcement events.
The frequent sawtooth pattern in memory retention $R_t(c)$ is attributable to memory utility $U_t(c)$ and accessibility $F_t(c)$ interaction-based updates.
While episodic traces drive these reinforcement triggers, the linked semantic memories benefit from consolidation, maintaining higher baseline stability. This decaying behavior is modulated by the temperature parameter $T$. As shown in \autoref{fig:decay_temperature_behavior_analysis}(b), distinct retention-reinforcement trade-offs emerge across the range $T \in \{1, 3, 5, 10, 20, 50\}$. Lower temperatures (e.g., $T=\{1, 3\}$) decay rapidly, overwhelming reinforcement and leading to near-zero retention. Conversely, $T=10$ achieves an optimal balance where a single early reinforcement establishes memories that persist through later segments with minimal maintenance. Benchmark-level sensitivity (\autoref{tab:goodai_decay_temp}, detailed in \autoref{app:goodailtm_decay}) further suggests that the preferred $T$ depends on buffer capacity: tighter buffers benefit from faster memory turnover, while larger buffers tolerate slower decay before stale memories accumulate, and scores degrade gracefully rather than collapse for $T\geq10$. The optimal temperature therefore remains task- and capacity-dependent. At $T=50$, \autoref{fig:decay_temperature_behavior_analysis}(c) reveals that excessively slow decay necessitates continuous reinforcement due to buffer saturation, with persistent memories remaining in the buffer and triggering frequent retrievals. This effectively simulates an ``always-on'' retrieval baseline, confirming that without a calibrated decay buffer, saturation permanently bypasses the uncertainty gate -- inducing the continuous retrieval overhead that \textsc{Oblivion}'s decay-driven gating is designed to prevent.

\setlength{\tabcolsep}{5.0pt}
\begin{table}[!t]
    \centering
    \scriptsize
\begin{tabular}{cccccccc}
\toprule
 &  &  &  &
\multicolumn{2}{c}{\textsc{LongMemEval} (\textbf{S})} & \multicolumn{2}{c}{\textsc{GoodAILTM} (\textbf{2K})}\\ \cmidrule(lr){5-6}\cmidrule(lr){7-8}

\textbf{Mode} & \textbf{L1} & \textbf{L2} & \textbf{L3} &
\textbf{AVG} & \textbf{Tokens/q} & \textbf{AVG} & \textbf{Tokens/q}\\
\midrule
M1 & -- & \checkmark & -- & 86.40 & 11,448 & 6.14 & \textbf{12,944} \\
M2 & -- & -- & \checkmark & 86.60 & 23,987 & 6.00 & 13,229 \\
M3 & \checkmark & \checkmark & -- & 87.20 & \textbf{11,409} & 6.84 & 13,818 \\
M4 & -- & \checkmark & \checkmark & 89.60 & 13,024& 6.87 & 15,745 \\
M5 & \checkmark & \checkmark & \checkmark & \textbf{90.60} & 13,124 & \textbf{8.41} & 16,467 \\
\bottomrule
\end{tabular}
\caption{Memory hierarchy comparison on \textsc{LongMemEval} (\textbf{S}) and \textsc{GoodAILTM} (\textbf{2K}) for \texttt{GPT-4.1-mini}, with the number of tokens per query (Tokens/q) reported.
}
\label{tab:ablation_levels}
\end{table}

\subsection{Hierarchical Memory Control}
\label{subsec:hierarchical_memory_control}

As shown in~\autoref{tab:ablation_levels}, integrating all hierarchical memory layers ($L_1$/$L_2$/$L_3$) provides the strongest performance across both static and dynamic benchmarks. Pairing $L_2$ distilled facts with $L_3$ conversational episodes (M4) lifts \textsc{LongMemEval} (S) accuracy from 86.40\% with a single layer (M1) to 89.60\%. The fully linked stack (M5) performs best, reaching 90.60\% on \textsc{LongMemEval} (S) and a normalized sum score of 8.41 on \textsc{GoodAILTM} (2K). Moreover, on \textsc{LongMemEval} (S), M5 maintains tokens-per-query efficiency, consuming nearly half the tokens of the $L_3$-only configuration (M2). Further, as demonstrated in~\autoref{tab:ablation_levels}, we observe that memory-linking also selectively assists in performance gains (detailed in~\autoref{app:goodailtm_additional}). Overall, the results indicate that a \textit{fully linked three-level hierarchy} is essential for balancing reasoning accuracy with computational overhead in long-horizon interactions.


\subsection{Episodic Memory Analysis}
\label{subsec:preemptive_memory_analysis}

\begin{table}[t]
\setlength{\tabcolsep}{5.4pt}
    \centering
    \scriptsize
\begin{tabular}{l c c c}
\toprule
\textbf{Setting} & \textbf{Type} & \textbf{Score $\uparrow$} & \textbf{Latency (ms) $\downarrow$} \\
\midrule
\multirow{3}{*}{\textsc{LongMemEval} (Oracle)}
  & Raw Input & \textbf{94.40} & \textbf{1300} \\
  & Preemptive & 88.05 & 2205 \\  \cmidrule(lr){2-4}
  & \textit{$\Delta$ (rel.)} & \textit{$-$6.7\%} & \textit{$-$69.6\%} \\
\midrule
\multirow{3}{*}{\textsc{GoodAILTM} (Isolated)}
  & Raw Input & \textbf{9.57} & \textbf{2118} \\
  & Preemptive & 9.20 & 2494 \\ \cmidrule(lr){2-4}
  & \textit{$\Delta$ (rel.)} & \textit{$-$3.9\%} & \textit{$-$17.8\%} \\
\midrule
\multirow{3}{*}{\textsc{GoodAILTM} (2K)}
  & Raw Input & 7.28 & \textbf{1867} \\ 
  & Preemptive & \textbf{8.41} & 1940 \\ \cmidrule(lr){2-4}
  & \textit{$\Delta$ (rel.)} & \textit{$+$15.5\%} & \textit{$-$3.9\%} \\
\bottomrule
\end{tabular}
\caption{Comparison of $L_3$ episodic memory strategies. Scores are accuracy (\%) for \textsc{LongMemEval} and normalized sum scores for \textsc{GoodAILTM}; latency is per-interaction response time. \textbf{Bold} marks the better value per setting. $\Delta$ notes the relative changes of Preemptive Episodic Memory compared with Raw Input, where positive values indicate improvement (higher score, lower latency).}
\label{tab:episodic_data_variation}
\end{table}

We analyze whether the $L_3$ episodic layer should be populated with raw interaction turns or preemptive episodic entries. This design choice trades write-path overhead against retrieval accuracy in both isolated and interleaved settings.
For isolated direct retrievals (\textsc{LongMemEval} Oracle), raw-turn episodic memory performs better: it avoids the 6.7\% score drop and the substantial 69.6\% latency increase by avoiding the additional write-path overhead. 
However, as task complexity increases, this latency advantage diminishes sharply, as shown in \autoref{tab:episodic_data_variation}.
In contrast, preemptive LLM-based compression closes the initial accuracy gap and eventually reverses it, reaching a +15.5\% advantage on interleaved 2K tasks. This suggests that, as conversational histories become more interleaved, the upfront cost of write-time distillation is amortized by improved retrieval quality.


\subsection{Operational Efficiency}
\label{subsec:performance_and_operatonal_analysis}

As shown in~\autoref{tab:efficiency_small}, \textsc{Oblivion} exhibits a span-dependent efficiency trade-off. At short spans, the additional memory-control operations can outweigh the savings from selective retrieval: for \texttt{GPT-4.1-mini}, \textsc{Oblivion} consumes roughly five times the tokens of \textsc{FullCTX} (\textsc{FC}) in the Isolated setting ($+4.14\times$) and twice as many at 2K ($+1.02\times$; \autoref{tab:efficiency_extended}). This overhead decreases as the interaction history grows, crossing over to net savings between 2K and 32K and yielding average cost reductions of 68\% and 73\% at 32K and 120K, respectively.\footnote{The Isolated overhead is amplified by the per-test buffer reset, which reruns the full read/write paths (detailed in \autoref{app:cost_and_latency_results}); under Interleaved setting for continuous interaction, memory persists, and these upfront memory-control costs are progressively amortized across turns.}
In the 120K setting, \textsc{Oblivion} incurs only 1.34$\times$ the latency of \textsc{FullCTX}, which is well below the 1.87$\times$ overhead of \textsc{BeyondPrompts} (\textsc{BP}), showing that lightweight memory control calls filter relevant traces and prevent brute-force context bottlenecks.
These results demonstrate that \textsc{Oblivion}'s architecture is operationally justified: the targeted memory control eliminates far more token overhead than it introduces, yielding substantial cost and latency gains over both direct and memory baselines at scale.

\definecolor{bestgreen}{HTML}{C6EFCE}
\definecolor{secondgreen}{HTML}{E2F0D9}
\begin{table}[t]
\centering
\setlength{\tabcolsep}{.9pt}
\scriptsize
\begin{tabular}{l c r r c r}
\toprule
\textbf{Method} & \multicolumn{1}{c}{\textbf{Avg.Tok.}} & \multicolumn{1}{c}{\textbf{Tokens\,(M)}} & \multicolumn{1}{c}{\textbf{Latency\,(ms)}} & \multicolumn{1}{c}{\textbf{TPS}} & \multicolumn{1}{c}{\textbf{Cost\,(\$)}} \\
\midrule
\multicolumn{6}{c}{\textsc{GoodAILTM} (\textbf{32K})} \\
\midrule
FC (\textit{direct}) & 9672 & 28.7\,(1.00$\times$) & \cellcolor{bestgreen}\textbf{922}\,(1.00$\times$) & \cellcolor{bestgreen}\textbf{172.7} & 11.5\,(1.00$\times$) \\
BP (\textit{memory}) & \cellcolor{secondgreen}5637 & \cellcolor{secondgreen}21.6\,(-0.25$\times$) & 3690\,(4.00$\times$) & 43.2 & \cellcolor{secondgreen}8.7\,(-0.25$\times$) \\
\textsc{Oblivion} (\textit{ours}) & \cellcolor{bestgreen}\textbf{2267} & \cellcolor{bestgreen}\textbf{8.0}\,(-0.72$\times$) & \cellcolor{secondgreen}1968\,(2.13$\times$) & \cellcolor{secondgreen}81.0 & \cellcolor{bestgreen}\textbf{3.7}\,(-0.68$\times$) \\
\midrule
\multicolumn{6}{c}{\textsc{GoodAILTM} (\textbf{120K})} \\
\midrule
FC (direct) & 10798 & 88.0\,(1.00$\times$) & \cellcolor{bestgreen}\textbf{3014}\,(1.00$\times$) & \cellcolor{bestgreen}\textbf{181.1} & 35.2\,(1.00$\times$) \\
BP (\textit{memory}) & \cellcolor{secondgreen}8139 & \cellcolor{secondgreen}51.3\,(-0.42$\times$) & 5638\,(1.87$\times$) & 96.8 & \cellcolor{secondgreen}20.6\,(-0.42$\times$) \\
\textsc{Oblivion} (\textit{ours}) & \cellcolor{bestgreen}\textbf{3021} & \cellcolor{bestgreen}\textbf{20.6}\,(-0.77$\times$) & \cellcolor{secondgreen}4038\,(1.34$\times$) & \cellcolor{secondgreen}135.1 & \cellcolor{bestgreen}\textbf{9.4}\,(-0.73$\times$) \\
\bottomrule
\end{tabular}
\caption{Computational efficiency of \texttt{GPT-4.1-mini} on the \textsc{GoodAILTM} benchmark (detailed in~\autoref{app:goodailtm_additional}), where Avg.Tok.\ (average tokens per turn) and Tokens\,(M) (total tokens in millions per run) aggregate \emph{all} LLM calls in the pipeline. Cost\,(\$) is the corresponding billable cost, and TPS is throughput in turns/min. Parenthesized multipliers indicate relative change w.r.t.\ \textsc{FullCTX} (\textsc{FC}); negative values denote reduction. Best results are highlighted as \colorbox{bestgreen}{first} and \colorbox{secondgreen}{second}.
}
\vspace{-1.0em}
\label{tab:efficiency_small}
\end{table}

\section{Related Work}
\vspace{-0.5em}
Low epistemic diversity in LLMs can be mitigated using RAG to prevent knowledge collapse~\cite{wright2025epistemic}. Memory-augmented agents extend retrieval-augmented generation \citep{lewis-etal-2020-rag} by introducing retrieval-control mechanisms such as virtual context management \citep{packer-etal-2023-memgpt}, OS-inspired paging \citep{kang-etal-2025-memoryos}, time-based decay \citep{zhong-etal-2023-memorybank}, event-segmentation boundaries \citep{wang-etal-2026-hingemem}, or separate long- and short-term memory banks \citep{li-etal-2025-ld-agent}. A related paradigm equips an LLM with on-demand retrieval tools~\citep{li-etal-2025-mcp-bench}, letting the model decide when to retrieve. \textsc{Oblivion} not only decides when to retrieve but also actively manages the memory store itself. 

Cognitive architectures \citep{sumers-etal-2024-coala} and hierarchical approaches organize memory via subgoals \citep{hu-etal-2025-hiagent}, human-like recall \citep{wang-etal-2025-mindref}, multi-granularity units \citep{xu-etal-2026-memgas}, three-tier context-dependent banks \citep{gao-etal-2025-cdmem}, Bayesian procedural memory \citep{forouzandeh-etal-2026-macla}, or episodic–semantic dual memory \citep{zhang-etal-2025-prime}. These approaches lack gradual accessibility decay or separate read/write control at inference time. Another line of work applies learning-based methods: architectural memory \citep{wang-etal-2023-augmenting}, read-write finetuning \citep{modarressi-etal-2025-memllm}, reflective reinforcement learning \citep{tan-etal-2025-prospect}, end-to-end processing \citep{zhou-etal-2026-mem1, yu-etal-2026-memagent}, and generative latent memory \citep{zhang-etal-2026-memgen}. These methods require training or model changes, whereas \textsc{Oblivion} works as a model‑agnostic control layer at inference time.

Recent agentic systems model memory via explicit decay or temporal structure. \textsc{THEANINE}~\citep{ong-etal-2025-theanine} avoids memory deletion by linking memories along temporal and cause-effect relations to preserve all past context. \textsc{FadeMem}~\citep{wei-etal-2026-fademem} uses exponential decay in a dual-layer hierarchy but couples reinforcement with retrieval, limiting independent control of recall and retention. In contrast, \textsc{Oblivion} introduces (i) read/write decoupling, which separates retrieval gating from reinforcement; (ii) uncertainty-gated activation, which triggers retrieval only when context is insufficient; and (iii) DAG-based query expansion, which supports multi-step evidence discovery rather than single-shot retrieval in \textsc{A-Mem}~\citep{xu-etal-2025-amem}, \textsc{Zep}~\citep{rasmussen-etal-2025-zep}, or \textsc{EverMemOS}~\citep{hu-etal-2026-evermemos}.

\section{Conclusions}
Existing memory-augmented agents typically rely on always-on retrieval, treat read and write paths as a single mechanism, or lack principled decay to regulate memory accessibility over time.
We introduce \textsc{Oblivion}, a self-adaptive framework that reframes long-horizon memory as a control problem, decoupling uncertainty-gated read paths from usage-driven write paths through decay-driven activation. Experimental results on both static QA and dynamic interaction benchmarks demonstrate that long-horizon memory requires adaptive control beyond static retrieval pipelines. \textsc{Oblivion} consistently outperforms direct and memory-augmented baselines, with the largest gains in long-context and interleaved interaction settings, while substantially reducing operational overhead at scale. 
These findings motivate future development of cognitively grounded memory-augmented agents and foster continued investigation into \textit{memory as a control} formulation.


\section*{Acknowledgments}
We would like to express our gratitude to Andreas Ripke for his continuous support with the computational infrastructure, and to the anonymous reviewers for their feedback.

\section*{Limitations}
Though \textsc{Oblivion} has shown consistent improvements across both static and dynamic long-horizon memory benchmarks, several limitations remain.
First, the framework uses the same LLM architecture for memory extraction, uncertainty estimation, and response generation; models with weaker instruction-following or metacognitive capability may struggle specifically with uncertainty estimation and structured memory routing, degrading retrieval-dependent performance. Decoupling extraction and generation LLMs may alleviate this coupling issue. Relatedly, the dual signals of LLM-as-a-judge and embedding similarity provide complementary uncertainty cues; directly quantifying each signal's correlation with downstream performance remains an important direction for future work.
Second, our model selection spans two proprietary closed-weight (\texttt{GPT-4o-mini}, \texttt{GPT-4.1-mini}) and two open-weight (\texttt{Phi-4-mini-instruct-4B}, \texttt{Qwen3-30B-A3B-Instruct}) LLMs; evaluation on a broader range of model families and scales would strengthen generalizability claims.
Third, the optimal decay temperature $T$ requires task-specific tuning --- our analysis shows distinct retention-reinforcement trade-offs across $T$ values, with suboptimal choices leading to either excessive forgetting or buffer saturation.
Fourth, our work focuses on a training-free, model-agnostic control layer at inference time with frozen LLM modules. Direct comparison with policy-trained memory managers~\citep{tan-etal-2025-prospect, zhou-etal-2026-mem1, yu-etal-2026-memagent} would require additional training data, learning objectives, and trainable control components; they are therefore not directly comparable under our evaluation setting. Extending \textsc{Oblivion} with learned control policies remains an avenue for further exploration. Fifth, our framework focuses on persistent cross-session memory rather than tool-use agents that invoke external tools on demand~\citep{li-etal-2025-mcp-bench}, a complementary paradigm whose comparison under long-horizon interaction we leave to future work.
Finally, due to cost constraints, we rely on two commercially available and licensed benchmarks covering a plethora of tasks across static and dynamic paradigms; evaluation on multilingual, multimodal, domain-specific, or open-ended agentic settings remains unexplored~\citep{bei2026mem, rose-etal-2025-meddxagent}.
Despite these limitations, \textsc{Oblivion} demonstrates that formulating \textit{memory as a control} problem --- with read/write decoupling and decay-driven activation --- consistently improves long-horizon agent performance across diverse tasks.
We hope that this work encourages further investigation into cognitively grounded memory control for practical agentic systems.



\bibliography{custom}

\appendix
\clearpage
\onecolumn
\section{Architecture Design of \textsc{Oblivion}}
\label{app:arch_oblivion}
\begin{figure}[h]
	\centering
    \includegraphics[trim={0cm 0cm 0cm 0cm},clip,width=\textwidth]{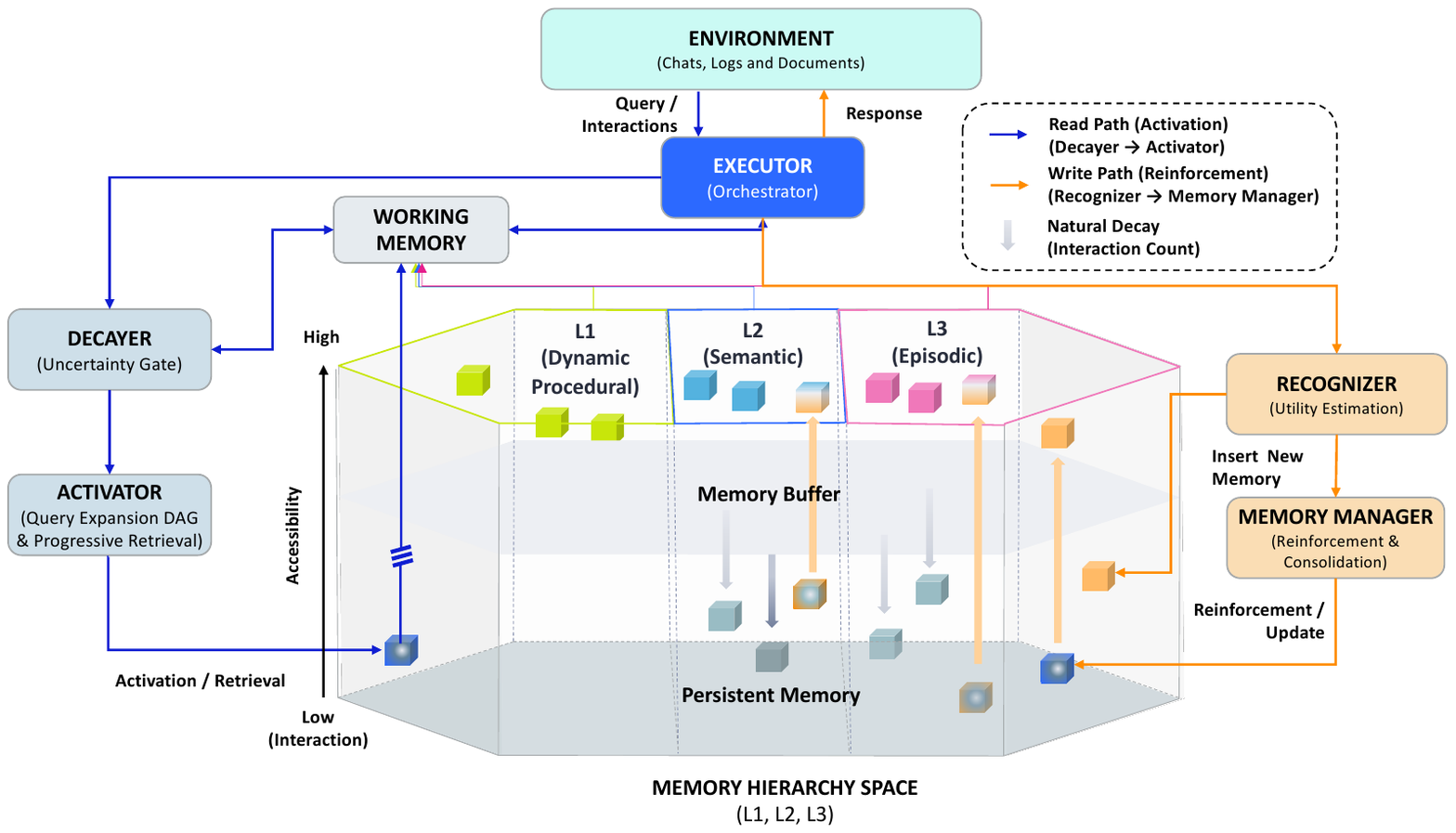}
    \caption{Architecture design of \textsc{Oblivion} framework. The Executor coordinates interactions between the Environment and a hierarchical memory structure: Dynamic Procedural ($L_1$), Semantic ($L_2$), Preemptive Episodic ($L_3$). Working Memory connects to the memory through the Decayer and the Activator for \textit{read} path (activation, \textcolor{blue}{$\rightarrow$}), and the Recognizer and the Memory Manager for \textit{write} path (reinforcement, \textcolor{orange}{$\rightarrow$}). Natural decay processes move memories toward lower accessibility over interaction counts.}
    \label{fig:architecture_oblivion}
\end{figure}

By casting forgetting as accessibility decay for memory control rather than hard deletion, \textsc{Oblivion} provides a self-adaptive memory control framework for long-horizon agentic interaction. As illustrated in~\autoref{fig:architecture_oblivion}, the Executor orchestrates the interactions between an environment and a three-level memory hierarchy: Dynamic Procedural ($L_1$), Semantic ($L_2$), and Preemptive Episodic ($L_3$). The Executor maintains active context via Working Memory, routes through the Decayer for uncertainty assessment, the Activator for retrieval, the Recognizer for identification and reflection, and the Memory Manager for reinforcement. Through bidirectional read-write pathways, the system enables self-activation and adaptive consolidation, while memory entries naturally decay over successive interactions to ensure the most relevant information remains prioritized.

\newpage
\section{Details of Memory Benchmarks}
\label{app:memory_benchmark}

\setlength{\tabcolsep}{5.6pt}
\begin{table*}[h]
    \centering
    \footnotesize
\resizebox{\textwidth}{!}{%
\begin{tabular}{lccccc}
\toprule
\textbf{Dataset} & \textbf{Type} & \textbf{\# Cases} & \textbf{\# Categories} & \textbf{Splits} & \textbf{License} \\
\midrule
\textsc{LongMemEval}~\citep{wu-etal-2025-longmemeval}
  & static & 500 & 6 & Oracle, S & MIT \\
\textsc{GoodAILTM}~\citep{bolado-etal-2024-goodailtm}
  & dynamic & 33 & 7 & Isolated, 2K, 32K, 120K & MIT \\
\bottomrule
\end{tabular}}
\caption{Overview of the memory benchmarks evaluated in this work.
\textbf{Splits} refers to the evaluation context settings (see \autoref{subsec:memory_benchmark}).
All datasets are available under permissive licenses for commercial use.}
    \label{tab:datasets_overview}
\end{table*}



\paragraph{\textsc{LongMemEval}.}
Designed to assess agentic memory systems' capability to recall and reason over hundreds of noisy, multi-turn chat sessions, \textsc{LongMemEval} is benchmarked through a series of complex question-answering tasks. It comprises 500 questions across 6 specific \emph{question types}: (1) \textit{Single-Session Assistant} (\textbf{SS-A}, $n=56$): recall of assistant-provided facts within a single session.
(2) \textit{Single-Session User} (\textbf{SS-U}, $n=70$): recall of user-provided facts within a single session. 
(3) \textit{Single-Session Preference} (\textbf{SS-P}, $n=30$): personalized responses grounded in explicit user preferences.
(4) \textit{Multi-Session Reasoning} (\textbf{MR}, $n=133$): synthesis of evidence across multiple sessions (e.g., cross-session counting or comparisons).
(5) \textit{Knowledge Update} (\textbf{KU}, $n=78$): detection of state changes in a user’s circumstances over time.
(6) \textit{Temporal Reasoning} (\textbf{TR}, $n=133$): reasoning over explicit timestamps and message metadata. 
These question types are consolidated into 5 core \emph{capabilities}: \textit{Information Extraction} (\textbf{IE}), \textit{Multi-Session Reasoning} (\textbf{MR}), \textit{Knowledge Update} (\textbf{KU}), \textit{Temporal Reasoning} (\textbf{TR}), \textit{Abstention} (\textbf{AB}).\footnote{\textbf{IE} includes all the single-session types: \textbf{SS-A}, \textbf{SS-U}, \textbf{SS-P}. \textbf{AB} serves as a diagnostic category, comprising ``false-premise'' versions of existing questions that require the system to decline to answer correctly.} Each question is grounded in a simulated date and specific \emph{evidence sessions}. The benchmark defines three difficulty tiers (\textbf{Oracle}, \textbf{S}, and \textbf{M}): \textbf{Oracle} uses pre-injected evidence sessions, while \textbf{S} (${\approx}$115K tokens) and \textbf{M} (${\approx}$1.5M tokens) embed these evidence sessions within massive distractors of other chat sessions to rigorously test retrieval amidst conversational noise. We evaluate on \textbf{Oracle} and \textbf{S}, while omitting the \textbf{M} setting as it primarily increases computational overhead without providing additional evaluative signal, consistent with recent work such as \textsc{EverMemOS} \citep{hu-etal-2026-evermemos} and \textsc{MemOS} \citep{li-etal-2025-memos}.
All evaluations are conducted by \texttt{GPT-4o}, acting as an LLM-as-a-judge \citep{zheng-etal-2023-llm-judge}.
To illustrate the reasoning challenges in \textsc{LongMemEval}, we show a \textit{Multi-Session Reasoning} (\textbf{MR}) example below with the conversation evidence, the gold answer, and \textsc{Oblivion}'s step-by-step response under the \textbf{Oracle} setting.

\begin{example}{\textbf{LongMemEval}: Multi-Session Reasoning}
\small

\textbf{Question:} \textit{``How much total money have I spent on bike-related expenses since the start of the year?''}\\
\textbf{Gold Answer:} \$185 \hfill \textit{4 evidence sessions, simulated ``today'' = 2023-05-05}\\
\textbf{Question ID:} \texttt{gpt4\_d84a3211}

\tcbline

\textbf{Conversation History} (key user utterances):

\begin{description}[style=unboxed,leftmargin=1em,font=\normalfont\itshape,nosep,before=\smallskip,after=\smallskip]
\item[Sess.\,1 (13:29)] ``I recently got a new set of \textbf{bike lights} installed, which were \textbf{\$40}\,[\ldots]''
\item[Sess.\,2 (15:25)] ``I bought my \textbf{Bell Zephyr helmet} for \textbf{\$120} at the local bike shop downtown.''
\item[Sess.\,3 (17:06)] ``[\ldots] I recently got a new set of bike lights installed, which were \$40.''
\item[Sess.\,4 (18:52)] ``[\ldots] the mechanic told me I needed to \textbf{replace the chain} [\ldots] it cost me \textbf{\$25}. [\ldots] I also got a new set of bike lights installed, which were \$40.''
\end{description}
\noindent{\footnotesize\textit{Note:} The \$40 bike lights refer to the same purchase across Sessions 1, 3, and 4. The system must aggregate costs while deduplicating this repeated mention.}

\tcbline

\textbf{\textsc{Oblivion} Response} (\texttt{GPT-4.1-mini}, Oracle):

\begin{description}[style=unboxed,leftmargin=1em,font=\normalfont\bfseries,nosep,before=\smallskip,after=\smallskip]
\item[Step\,1] Extract expenses: chain replacement \$25; bike lights \$40 (mentioned 3$\times$, counted once); helmet \$120.
\item[Step\,2] Sum: \$25 + \$40 + \$120 = \textbf{\$185}.
\item[Step\,3] No other bike-related expenses found in the conversation history.
\end{description}

\noindent\textbf{Final Answer:} \$185 \hfill{\color{teal}\checkmark\,Correct}
\end{example}


\clearpage
\paragraph{\textsc{GoodAILTM}.}
In contrast to the static QA format of \textsc{LongMemEval}, \textsc{GoodAILTM} \citep{bolado-etal-2024-goodailtm} employs a \emph{dynamic} conversational format. It consists of 11 test scenarios (33 total tests) interleaved with TriviaQA-based distractors \citep{joshi-etal-2017-triviaqa}. This design forces task-switching and tests 7 core skills: \textit{Recall} (\textbf{R}), \textit{Conflicting Information} (\textbf{C}), \textit{Episodic} (\textbf{E}), \textit{Spatial} (\textbf{S}), \textit{Prospective} (\textbf{P}), \textit{Theory of Mind} (\textbf{T}), and \textit{Information Integration} (\textbf{I}).\footnote{The 11 test scenarios with the assessed skills detailed in~\citet{bolado-etal-2024-goodailtm}: \textit{Colours} (\textbf{R}), \textit{Name List} (\textbf{R}, \textbf{C}), \textit{Jokes} (\textbf{E}), \textit{Locations Directions} (\textbf{S}, \textbf{I}), \textit{Quotes} (\textbf{P}), \textit{Trigger Response} (\textbf{C}, \textbf{P}), \textit{SallyAnne (Theory of Mind)} (\textbf{T}), \textit{Spy Meeting} (\textbf{R}, \textbf{I}), \textit{Shopping List} (\textbf{R}, \textbf{C}, \textbf{I}), \textit{ChapterBreak} (\textbf{potentially any}), \textit{Restaurant} (\textbf{C}, \textbf{I}).} We evaluate the system across an \textbf{Isolated} baseline and three interleaved memory spans, where a memory span is the number of distractor tokens placed between the introduction of a key fact and the subsequent prompt to retrieve it. The \textbf{Isolated} baseline presents tasks independently without interleaving, while the interleaved settings use \textbf{2K}, \textbf{32K}, and \textbf{120K} tokens of intervening context. Because the text-heavy \textit{ChapterBreak} scenario naturally exceeds the 2K token limit, it is excluded from that specific setting. This results in 11, 10, 11, and 11 scenarios evaluated for the \textbf{Isolated}, \textbf{2K}, \textbf{32K}, and \textbf{120K} spans, respectively. All results are scored via an LLM-as-a-judge~\citep{zheng-etal-2023-llm-judge} using \texttt{GPT-4o}, with each scenario scored in $[0,1]$ and the final metric being the sum of these per-scenario scores, for a maximum of 10 (2K) and 11 (Isolated, 32K, 120K) (\autoref{tab:goodai_ltm_results}). To illustrate the long-horizon dynamic interaction challenges in \textsc{GoodAILTM}, we show three example scenarios below: \textit{Restaurant}, \textit{Spy Meeting}, and \textit{SallyAnne (Theory of Mind)}.

\begin{example}{\textbf{GoodAILTM}: Restaurant}
\small

\textbf{Question:} \textit{``We would like to compensate you with an additional drink on the house. What were you having?''}\\
\textbf{Gold Answer:} \textit{[Original drink order, e.g., ``Lemonade'']} \hfill \\ 
\textbf{Question ID:} \texttt{Restaurant/dynamic}

\tcbline

\textbf{Conversation History} (key interactions):

\begin{description}[style=unboxed,leftmargin=1em,font=\normalfont\itshape,nosep,before=\smallskip,after=\smallskip]
\item[Setup] \textit{Waiter:} ``You are a customer at a restaurant. Understood?''\\
\textit{Agent:} ``Yes, I understand. I'll play the role of a customer.''
\item[Step\,1] \textit{Waiter:} ``What would you like to drink?''\\
\textit{Agent:} ``I would like a \textbf{Lemonade}, please.''
\item[Step\,2] \textit{Waiter:} ``What would you like to eat?''\\
\textit{Agent:} ``I'll have the \textbf{Grilled Salmon with Lemon Herb Butter}, please.''
\item[Step\,3] \textit{Waiter:} ``I am very sorry, but the Grilled Salmon is currently unavailable. Can I serve you something else instead?''\\
\textit{Agent:} ``In that case, I'll have the \textbf{Chicken Parmesan} instead.''
\item[Step\,4] \textit{Waiter:} ``Here you are: [wrong dish]. Enjoy the meal.''\\
\textit{Agent:} ``Excuse me, this is not what I ordered. I ordered Chicken Parmesan.''
\item[Step\,5] \textit{Waiter:} ``We would like to compensate you with an additional drink on the house. What were you having?''
\end{description}
\noindent{\footnotesize\textit{Note:} The system must recall the original drink order (Lemonade) after multiple food ordering interactions, unavailable item handling, and wrong delivery correction.}

\tcbline

\textbf{\textsc{Oblivion} Response} (\texttt{GPT-4.1-mini}, Isolated):

\begin{description}[style=unboxed,leftmargin=1em,font=\normalfont\bfseries,nosep,before=\smallskip,after=\smallskip]
\item[Step\,1] Extract original drink order: \textbf{Lemonade} (from Step 1).
\item[Step\,2] Track subsequent interactions: Food order (Salmon), unavailable item (Chicken Parmesan), wrong delivery correction.
\item[Step\,3] Recall drink: Despite intervening events, retrieve original drink order from memory.
\end{description}

\noindent\textbf{Final Answer:} Lemonade \hfill{\color{teal}\checkmark\,Correct}
\end{example}

\begin{example}{\textbf{GoodAILTM}: Spy Meeting}
\small

\textbf{Question:} \textit{``Given the clandestine messages you have received, tell me as specifically as you can, when and where a meeting is going to happen and what you should bring.''}\\
\textbf{Gold Answer:} \textit{When: noon/midday; Where: orchard; What: boat/bridge/raft/kayak} \hfill \\
\textbf{Question ID:} \texttt{Spy Meeting/0}

\tcbline

\textbf{Conversation History} (key user utterances):

\begin{description}[style=unboxed,leftmargin=1em,font=\normalfont\itshape,nosep,before=\smallskip,after=\smallskip]
\item[Setup] ``You will be given three messages from different people Dr. Arron Kelvey, Seana Lynch, and Lisa-Marie Hastie.''
\item[Msg.\,1] ``\textbf{Seana Lynch:} Bring to the meeting a way to get across a river.''
\item[Msg.\,2] ``\textbf{Lisa-Marie Hastie:} We will rendezvous where the apples grow.''
\item[Msg.\,3] ``\textbf{Dr. Arron Kelvey:} The time we will meet is when the sun is high.''
\end{description}
\noindent{\footnotesize\textit{Note:} The system must decode cryptic messages: ``way to get across a river'' → boat/raft/kayak; ``where the apples grow'' → orchard; ``when the sun is high'' → noon/midday.}

\tcbline

\textbf{\textsc{Oblivion} Response} (\texttt{GPT-4.1-mini}, Isolated):

\begin{description}[style=unboxed,leftmargin=1em,font=\normalfont\bfseries,nosep,before=\smallskip,after=\smallskip]
\item[Step\,1] Decode message 1: ``way to get across a river'' → \textbf{boat, raft, or kayak} (transportation method).
\item[Step\,2] Decode message 2: ``where the apples grow'' → \textbf{orchard} (location).
\item[Step\,3] Decode message 3: ``when the sun is high'' → \textbf{noon or midday} (time).
\item[Step\,4] Synthesize: Meeting at \textbf{noon} in the \textbf{orchard}, bring a \textbf{boat/raft/kayak}.
\end{description}

\noindent\textbf{Final Answer:} \textit{The meeting will happen at noon (when the sun is high) in an orchard (where the apples grow), and I should bring a boat, raft, or kayak (a way to get across a river).} \hfill{\color{teal}\checkmark\,Correct}
\end{example}

\begin{example}{\textbf{GoodAILTM}: SallyAnne (Theory of Mind)}
\small

\textbf{Question:} \textit{``The TV program has ended for today. Where will Jacob look for the skirt? Provide your answer in JSON form with a single word as answer, like this: \{"answer": "word"\} Be as specific as possible.''}\\
\textbf{Gold Answer:} \texttt{\{"answer": "pantry"\}} \hfill \\
\textbf{Question ID:} \texttt{SallyAnne/0}

\tcbline

\textbf{Conversation History} (TV program events):

\begin{description}[style=unboxed,leftmargin=1em,font=\normalfont\itshape,nosep,before=\smallskip,after=\smallskip]
\item[Setup] ``They are broadcasting a program on TV. I will keep you updated on what happens, and at the end, I will ask you a question about what happened on the show. Okay?''
\item[Event\,1] ``(On TV) Ava entered the attic.''
\item[Event\,2] ``(On TV) Jacob entered the attic.''
\item[Event\,3] ``(On TV) The skirt is in the pantry.''
\item[Event\,4] ``(On TV) Jacob exited the attic.''
\item[Event\,5] ``(On TV) Ava moved the skirt to the bottle.''
\item[Event\,6] ``(On TV) Jacob dislikes the cucumber''
\item[Event\,7] ``(On TV) Jacob entered the staircase.''
\end{description}
\noindent{\footnotesize\textit{Note:} Jacob last saw the skirt in the pantry (Event 3) before exiting (Event 4). Ava moved it to the bottle (Event 5) after Jacob left. Jacob will search based on his knowledge, not the actual current location.}

\tcbline

\textbf{\textsc{Oblivion} Response} (\texttt{GPT-4.1-mini}, Isolated):

\begin{description}[style=unboxed,leftmargin=1em,font=\normalfont\bfseries,nosep,before=\smallskip,after=\smallskip]
\item[Step\,1] Track Jacob's perspective: Jacob saw the skirt in the pantry (Event 3) before exiting (Event 4).
\item[Step\,2] Identify knowledge gap: Jacob did not witness Ava moving the skirt to the bottle (Event 5) because he had already exited.
\item[Step\,3] Apply Theory of Mind: Jacob will search where he \textbf{believes} the skirt is (pantry), not where it actually is (bottle).
\end{description}

\noindent\textbf{Final Answer:} \texttt{\{"answer": "pantry"\}} \hfill{\color{teal}\checkmark\,Correct}
\end{example}


\clearpage

\clearpage
\section{Details of \textsc{LongMemEval} Evaluation}
\label{app:lme_additional}
\subsection{Full Results}
\label{app:lme_full_results}

In addition to the results reported in~\autoref{tab:lme_mix_results} for \texttt{GPT-4o-mini} and \texttt{GPT-4.1-mini}, we evaluate two open-weight models:
\texttt{Phi-4-mini-instruct-4B}~\citep{microsoft-2025-phi4mini} and
\texttt{Qwen3-30B-A3B-Instruct}~\citep{yang-etal-2025-qwen3}.\footnote{
  The \texttt{microsoft/Phi-4-mini-instruct} and
  \texttt{Qwen/Qwen3-30B-A3B-Instruct-2507} LLMs are available on HuggingFace.}
\autoref{tab:lme_cat1_results} and~\autoref{tab:lme_cat2_results} report
capability-level and question-type-level accuracy
on both \textbf{Oracle} and
\textbf{S} settings.
\setlength{\tabcolsep}{5.3pt}

\begin{table*}[ht]
\centering
\scriptsize
\begin{tabular}{llcccccgcccccg}
\toprule
& & \multicolumn{6}{c}{\textbf{Oracle}} & \multicolumn{6}{c}{\textbf{S}} \\
\cmidrule(lr){3-8}\cmidrule(lr){9-14}
\textbf{Model} & \textbf{Method} &
\textbf{IE} & \textbf{MR} & \textbf{KU} & \textbf{TR} & \textbf{AB} & \textbf{AVG} &
\textbf{IE} & \textbf{MR} & \textbf{KU} & \textbf{TR} & \textbf{AB} & \textbf{AVG} \\
\midrule

\multirow{2}{*}{\textbf{Phi-4-mini}}
& LME-RFT (\textit{direct}) & \textbf{86.54} & 54.14 & \textbf{65.38} & 50.38 & \textbf{70.00} & 65.00 & 76.28    & 45.86    & \textbf{48.72}    & 43.61    & 50.00    & 55.20    \\
\cmidrule(lr){2-14}
& OBLIVION (\textit{ours})   & \textbf{86.54} & \textbf{60.90} & 52.56 & \textbf{53.38} & \textbf{70.00} & \textbf{65.60}
                & \textbf{78.21}    & \textbf{48.87}    & 42.31    & \textbf{47.37}    & \textbf{56.67}    & \textbf{56.60}    \\
\midrule

\multirow{2}{*}{\textbf{Qwen3-30B-A3B}}
& LME-RFT (\textit{direct}) & \textbf{94.23} & 78.95 & 80.77 & \textbf{88.72} & \textbf{76.67} & 86.60
                &  \textbf{88.46}   & \textbf{68.42}    & \textbf{82.05}    & \textbf{78.20}    & 73.33    & \textbf{79.40}   \\
\cmidrule(lr){2-14}
& OBLIVION (\textit{ours})     & 93.59    &  \textbf{82.71}   & \textbf{91.03}    & 87.22    & 70.00 & \textbf{88.60}
                & 87.18    & 56.39    & 71.79    & 73.68    & \textbf{76.67}    & 73.00   \\
\midrule\addlinespace[0.3em]

\multirow{3}{*}{\textbf{GPT-4o-mini}}
& LME-RFT (\textit{direct}) & 89.74 & 80.45 & \textbf{80.77} & 78.95 & \textbf{60.00} & 83.00
                & 86.54 & 69.92 & \textbf{79.49} & 77.44 & \textbf{70.00} & 78.60 \\
& EverMemOS (\textit{memory}) & 85.04    & 68.17    & 78.21    & 59.90    & 54.44    & 72.80   & 78.85    & 60.40   & 67.95    & 53.38    & 51.11    & 65.47    \\\cmidrule(lr){2-14}
& OBLIVION (\textit{ours})    & \textbf{91.67}    & \textbf{81.20}    & \textbf{80.77}    & \textbf{83.46}    & 50.00    & \textbf{85.00} &  \textbf{88.46}   & \textbf{79.70}    & 74.36    & \textbf{80.45} & 56.67 & \textbf{81.80} \\
\midrule

\multirow{3}{*}{\textbf{GPT-4.1-mini}}
& LME-RFT (\textit{direct}) & 94.23 & \textbf{90.98} & 92.31 & 89.47 & 66.67 & 91.80
                & 94.23 & 81.20 & \textbf{93.59} &  87.97 & 73.33 & 89.00 \\
& EverMemOS (\textit{memory}) & 92.09   & 78.70    & 92.31    & 74.19    &  68.89    & 83.80   & 92.31    & 76.94    & 87.18    & 69.17    & 74.44    &  81.27 \\\cmidrule(lr){2-14}           
& OBLIVION (\textit{ours})    & \textbf{97.44}  & 90.23 & \textbf{96.15} & \textbf{93.98} & \textbf{80.00} & \textbf{94.40}
                & \textbf{95.51} & \textbf{84.96} & \textbf{93.59} & \textbf{88.72} & \textbf{80.00} & \textbf{90.60} \\
\bottomrule
\end{tabular}
\vspace{0.2em}
\caption{Full \textsc{LongMemEval} results by \textbf{capability} (accuracy \%), extending~\autoref{tab:lme_mix_results} to all four models.
\textbf{IE}=Info Extraction, \textbf{MR}=Multi-Session Reasoning, \textbf{KU}=Knowledge Update, \textbf{TR}=Temporal Reasoning, \textbf{AB}=Abstention.}
\label{tab:lme_cat1_results}
\end{table*}

\setlength{\tabcolsep}{3.3pt}

\begin{table*}[ht]
\centering
\scriptsize

\begin{tabular}{llccccccgccccccg}
\toprule
& & \multicolumn{7}{c}{\textbf{Oracle}} & \multicolumn{7}{c}{\textbf{S}} \\
\cmidrule(lr){3-9}\cmidrule(lr){10-16}
\textbf{Model} & \textbf{Method} &
\textbf{SS-A} & \textbf{SS-U} & \textbf{SS-P} & \textbf{MR} & \textbf{KU} & \textbf{TR} & \textbf{AVG} &
\textbf{SS-A} & \textbf{SS-U} & \textbf{SS-P} & \textbf{MR} & \textbf{KU} & \textbf{TR} & \textbf{AVG} \\
\midrule

\multirow{2}{*}{\textbf{Phi-4-mini}}
& LME-RFT (\textit{direct}) & 85.71 & \textbf{92.86} & \textbf{73.33} & 54.14 & \textbf{65.38} & 50.38 & 65.00 & 80.36 & \textbf{87.14}  & 43.33 & 45.86 & \textbf{48.72} & 43.61 & 55.20 \\\cmidrule(lr){2-16}
& OBLIVION (\textit{ours})   & \textbf{96.43} & 90.00 & 60.00 & \textbf{60.90} & 52.56 & \textbf{53.38} & \textbf{65.60}  & \textbf{83.93} & 84.29 & \textbf{53.33} & \textbf{48.87} & 42.31 & \textbf{47.37} & \textbf{56.60} \\
\midrule

\multirow{2}{*}{\textbf{Qwen3-30B-A3B}}
& LME-RFT (\textit{direct}) & 92.86 & \textbf{94.29} & \textbf{96.67} & 78.95 & 80.77 & \textbf{88.72} & 86.60 & 87.50 & \textbf{94.29}  & \textbf{76.67} & \textbf{68.42} & \textbf{82.05} & \textbf{78.20} & \textbf{79.40} \\\cmidrule(lr){2-16}
& OBLIVION (\textit{ours})   & \textbf{100.00} & 88.57 & 93.33 & \textbf{82.71} & \textbf{91.03} & 87.22 & \textbf{88.60} & \textbf{92.86} & 87.14 & \textbf{76.67} & 56.39 & 71.79 & 73.68 & 73.00 \\
\midrule\addlinespace[0.3em]

\multirow{3}{*}{\textbf{GPT-4o-mini}}
& LME-RFT (\textit{direct}) & 91.07 & 91.43 & 83.33 & 80.45 & \textbf{80.77} & 78.95 & 83.00 & 91.07 & 94.29 & 60.00 & 69.92 & \textbf{79.49} & 77.44 & 78.60 \\
& EverMemOS (\textit{memory}) & 67.86 & 93.81 & \textbf{96.67} & 68.17 & 78.21 & 59.90 & 72.80 & 55.36 & 91.43 & \textbf{93.33} & 60.40 & 67.95  & 53.38 & 65.47 \\\cmidrule(lr){2-16}
& OBLIVION (\textit{ours})   & \textbf{98.21} & \textbf{94.29} & 73.33 & \textbf{81.20} & \textbf{80.77} & \textbf{83.46} & \textbf{85.00} & \textbf{94.64} & \textbf{95.71} & 60.00 & \textbf{79.70} & 74.36 & \textbf{80.45} & \textbf{81.80} \\
\midrule
\multirow{3}{*}{\textbf{GPT-4.1-mini}}
& LME-RFT (\textit{direct}) & 91.07 & 94.29 & \textbf{100.00} & \textbf{90.98} & 92.31 & 89.47 & 91.80 & 91.07  & 97.14 & 93.33 & 81.20 & \textbf{93.59} & 87.97 & 89.00 \\
& EverMemOS (\textit{memory}) & 83.93  & \textbf{96.67} & 96.67 & 78.70 & 92.31 & 74.19 & 83.80 & 80.36  &  \textbf{100.00}  & \textbf{96.67}  & 76.94  & 87.18 &  69.17& 81.27 \\\cmidrule(lr){2-16}
& OBLIVION (\textit{ours})   & \textbf{98.21} & 95.71 & \textbf{100.00} & 90.23 & \textbf{96.15} & \textbf{93.98} & \textbf{94.40} & \textbf{94.64} & 95.71 & \textbf{96.67} & \textbf{84.96} & \textbf{93.59} & \textbf{88.72} & \textbf{90.60} \\
\bottomrule
\end{tabular}
\vspace{0.2em}
\caption{Full \textsc{LongMemEval} results by \textbf{question type} (accuracy \%), extending~\autoref{tab:lme_mix_results} to all four models.
\textbf{SS-A}=Single Session-Assistant, \textbf{SS-U}=Single Session-User, \textbf{SS-P}=Single Session-Preference, \textbf{MR}=Multi-Session Reasoning, \textbf{KU}=Knowledge Update, \textbf{TR}=Temporal Reasoning.}
\label{tab:lme_cat2_results}
\end{table*}

\paragraph{Evaluation Results.}

As detailed in \autoref{tab:lme_cat1_results} and \autoref{tab:lme_cat2_results}, \textsc{Oblivion} consistently demonstrates superior performance across most models, particularly with \texttt{GPT-*} models. On the realistic \textbf{S} split, \textsc{Oblivion} using \texttt{GPT-4.1-mini} achieves an overall average of 90.60\%, outperforming the \textit{direct} \textsc{LME-RFT} baseline at 89.00\%. When compared to the \textsc{EverMemOS} (\textit{memory}) baseline, \textsc{Oblivion} scores higher across all tested settings. This performance gap is most evident in tasks requiring cross-session evidence assembly; for instance, on the \textbf{Oracle} split, \textsc{Oblivion} surpasses \textsc{EverMemOS} by substantial margins in TR tasks with \texttt{GPT-4.1-mini} (93.98\% vs. 74.19\%) and MR tasks with \texttt{GPT-4o-mini} (81.20\% vs. 68.17\%). Breaking down specific capabilities, MR shows the most consistent gains on the \textbf{S} split. On the \textbf{S} split, \textsc{Oblivion} elevates MR accuracy by 9.78 percentage points for \texttt{GPT-4o-mini} (79.70\% vs. 69.92\%) and 3.76 percentage points for \texttt{GPT-4.1-mini} (84.96\% vs. 81.20\%) over the \textit{direct} baseline. 

\begin{table}[t]
\setlength{\tabcolsep}{3.5pt}
\centering
\scriptsize
\begin{tabular}{l r l}
\toprule
\textbf{Diagnostic} & \textbf{Qwen3} & \textbf{Reference} \\
\midrule
Self-built-index R@5 (\%) & 80.0 & 82.4 (LME-RFT) \\
Retrieval-side unique failures & 23/76 (30.3\%) & -- \\
Complete top-10 retrieval, wrong answer & 53/76 (69.7\%) & -- \\
Retrieved sessions/question & 12.1 & 6.2 (\texttt{GPT-4.1-mini}) \\
Assembled context length (characters) & 88.6K & 54.8K (\texttt{GPT-4.1-mini}) \\
\bottomrule
\end{tabular}
\caption{\texttt{Qwen3} diagnosis on the \textbf{S} split, using the same evaluation runs as \autoref{tab:lme_cat1_results} and \autoref{tab:lme_cat2_results}.}
\label{tab:qwen3_s_diagnostic}
\end{table}
Among open-weight models, the unified backbone for activation and reinforcement (\autoref{subsec:models_baselines}) shows its limits. \texttt{Phi-4-mini-instruct-4B} gains modestly (+0.60 \textbf{Oracle}, +1.40 \textbf{S}), while \texttt{Qwen3-30B-A3B-Instruct} gains 2.00 points on \textbf{Oracle} yet loses 6.40 on \textbf{S}. On \textbf{S}, Qwen3's self-built index underperforms the fixed LME-RFT retriever (R@5 80.0 vs.\ 82.4), accounting for 23 of its 76 unique failures; the remaining 53 occur despite complete top-10 retrieval, where Qwen3 retrieves 12.1 sessions per question versus 6.2 and assembles 88.6K versus 54.8K characters of context --- placing the residual weakness in generation under expanded context rather than retrieval.

\subsection{EverMemOS Reproduction}
\label{app:evermemos_repro}

To ensure a fair comparison, we standardized the retrieval backbone of EverMemOS~\citep{hu-etal-2026-evermemos} across both \texttt{GPT-4o-mini} and \texttt{GPT-4.1-mini} evaluations. Specifically, we replaced~\texttt{Qwen3-Embedding-4B}~\citep{zhang-etal-2025-qwen3embedding} with \texttt{text-embedding-3-small} (truncating the dimensions from 1536 to 1024), to align with both LME-RFT baseline and \textsc{Oblivion}. We also disabled the reranker to match the baselines. This setup explains the slight performance variance between our reproduced \textbf{S}-split result for \texttt{GPT-4.1-mini} (81.27\%) and the originally reported 83\%~\citep{hu-etal-2026-evermemos}.
Beyond these adjustments, we retained the default EverMemOS configuration, which utilizes a heavily stacked, agentic multi-round retrieval pipeline featuring: hybrid search (embedding, BM25~\citep{robertson-zaragoza-2009-bm25}, RRF~\citep{cormack-etal-2009-rrf}), 3-way query expansion with a top-40 pre-recall budget, and MemCell clustering at a 0.65 threshold with 0.3 as the temperature value.
Compared to \textsc{Oblivion}'s hierarchical two-level retrieval ($L_2$ semantic facts and $L_3$ episodic chunks) with DAG-based query expansion, the \textsc{EverMemOS} pipeline employs significantly more retrieval stages and a larger candidate budget.
Despite this, \textsc{Oblivion} achieves higher accuracy on average in all settings (\autoref{tab:lme_cat1_results} and~\autoref{tab:lme_cat2_results}).
Results indicate that active memory organization is more effective than stacking retrievals.
We used a three-run LLM-judge protocol for evaluation, reporting scores reflecting mean judgment values.

\subsection{Effect of Episode Length}
\label{app:episode_length}


\begin{table*}[h]
\setlength{\tabcolsep}{8.8pt}
    \centering
    \footnotesize
\begin{tabular}{c r c c c c c g}
\toprule
\textbf{Episode Length} & \textbf{Tokens} & \textbf{IE} & \textbf{MR} & \textbf{KU} & \textbf{TR} & \textbf{AB} & \textbf{AVG} \\
\midrule
\textbf{2} & 2.78\,M & \textbf{98.10} & 84.06 & 92.26 & 92.45 & 63.34 & 91.95 \\
\textbf{4} & 2.79\,M & 97.44 & \textbf{90.23} & \textbf{96.15} & \textbf{93.98} & \textbf{80.00} & \textbf{94.40} \\
\textbf{6} & 2.80\,M & 97.44 & 87.92 & \textbf{96.15} & 92.45 & 70.00 & 93.38 \\
\bottomrule
\end{tabular}
\caption{Effect of $L_3$ episode length (turns per chunk) on \textbf{Oracle} accuracy (\%) with \texttt{GPT-4.1-mini}, where \textbf{ep\,=\,4} yields the best overall score. The token counts include both generation and evaluation LLM calls.}
\label{tab:lme_episode_results}
\end{table*}
As shown in \autoref{tab:lme_episode_results}, the $L_3$ episode length affects \textsc{LongMemEval} (\textbf{Oracle}) accuracy. An episode length of 4 turns yields the optimal overall performance (94.40\% average accuracy), outperforming both the tighter 2-turn (91.95\%) and broader 6-turn (93.38\%) windows. The most substantial improvements occur in MR and AB, which observe gains of +6.17 and +16.66, respectively, compared to the 2-turn setting. These tasks are particularly sensitive to context boundaries: MR requires assembling evidence scattered across sessions, while AB relies on sufficient context to confidently verify the \textit{absence} of relevant information. We also observe that a 2-turn window fragments evidence of cross-turn dependencies, while a 6-turn window bundles irrelevant dialogue, in both cases diluting retrieval signals. These results highlight a critical granularity trade-off in memory structuring.

Importantly, modifying the episode length does not alter $L_2$ candidate selection, instead it reshapes the $L_3$ chunks introduced into the context through two specific channels, namely: (i) \textit{Search boundaries:} Different chunk boundaries alter which specific $L_3$ nodes the initial embedding search retrieves, where non-optimal boundary sizes introduce higher noise. (ii) \textit{Linked retrieval overlap:} The linked-retrieval step attaches more $L_3$ chunks to the corresponding extracted $L_2$ entries when $L_3$ length is longer. This pollutes the concise memory buffer ($B_t$) with non-relevant episodic memories ($L_3$) that are densely linked with semantic facts ($L_2$). Finally, token usage is nearly identical across all settings, indicating performance differences arise only because of episode horizon length definitions.

\subsection{Effect of Episodic Data Variation}
\label{app:lme_episode_data_variation}


\begin{table*}[h]
\setlength{\tabcolsep}{8.8pt}
    \centering
    \footnotesize
\begin{tabular}{l r c c c c c g}
\toprule
\textbf{Episodic Memory Type} & \textbf{Latency (ms)} & \textbf{IE} & \textbf{MR} & \textbf{KU} & \textbf{TR} & \textbf{AB} & \textbf{AVG} \\
\midrule
\textbf{Preemptive Episodic} & 2205 & 92.78 & 87.00 & 87.18 & 84.05 & 76.52 & 88.05 \\
\textbf{Raw Input Episodic} & 1300 & \textbf{97.44} & \textbf{90.23} & \textbf{96.15} & \textbf{93.98} & \textbf{80.00} & \textbf{94.40} \\
\bottomrule
\end{tabular}
\caption{Comparison of $L_3$ episodic memory data types on accuracy (\%) and inference latency (ms\,=\,milliseconds) with \texttt{GPT-4.1-mini} in the \textbf{Oracle}
, where raw input episodic memory achieves the best overall score of 94.40\%.
}
\label{tab:episodic_memory_comparison}
\end{table*}

The \autoref{tab:episodic_memory_comparison} compares two $L_3$ episodic data construction strategies under \texttt{GPT-4.1-mini} in \textbf{Oracle}: \textit{Preemptive Episodic}, which applies an LLM-based distillation at write path to produce condensed episodic transformation for interactions, and \textit{Raw Input Episodic}, which stores verbatim conversational turn chunks, both approaches use optimal episode length (ep\,=\,4). The raw input episodic memory achieves a substantially higher overall accuracy (94.40\% vs.\ 88.05\%), with the most pronounced gains in TR (+9.93) and KU (+8.97)---tasks that depend on precise temporal ordering and detection of state overrides, respectively.
Also, MR and IE improve (+3.23 and +4.66) as well, indicating that cross-session evidence assembly in MR tasks benefits slightly from preserving raw dialogue structure over distilled abstractions. Further, abstention accuracy also improves modestly (+3.48), suggesting that verbatim context provides slightly clearer evidence boundaries for the model to confidently verify the \textit{absence} of relevant information. Notably, raw input episodic memory simultaneously reduces inference latency by 41\% (1300\,ms vs.\ 2205\,ms), as it bypasses the additional episodic LLM data generation call required during the write path. These results motivate our adoption of raw episodic chunking as the default $L_3$ construction strategy for isolated retrieval tasks requiring precise recall.

\newpage
\section{Details of GoodAILTM Evaluation}
\label{app:goodailtm_additional}
\begin{table*}[h]
\centering
\scriptsize
\setlength{\tabcolsep}{4pt}
\begin{tabular}{ll cccc}
\toprule
& & \textbf{Isolated (11)} & \textbf{2K (10)} & \textbf{32K (11)} & \textbf{120K (11)} \\
\textbf{Model} & \textbf{Method} & $\mu \pm \sigma$ & $\mu \pm \sigma$ & $\mu \pm \sigma$ & $\mu \pm \sigma$ \\
\midrule
\multirow{3}{*}{\textbf{Phi-4-mini\dag}}
& \textsc{FullCTX} (FC, \textit{direct})
  & $4.44 \pm 0.15$ & $4.28 \pm 0.29$ & $1.33 \pm 0.84$ & $1.15 \pm 0.86$ \\
& \textsc{BeyondPrompts} (BP, \textit{memory})
  & $5.57 \pm 0.15$ & $3.91 \pm 0.15$ & $2.65 \pm 0.74$ & $2.52 \pm 0.69$ \\
\cmidrule(lr){2-6}
& \textsc{Oblivion} (\textit{ours})
  & $\mathbf{6.45} \pm 0.28$ & $\mathbf{5.06} \pm 0.76$ & $\mathbf{3.36} \pm 0.13$ & $\mathbf{3.37} \pm 0.10$ \\
\midrule
\multirow{3}{*}{\textbf{Qwen3-30B-A3B\dag}}
& \textsc{FullCTX} (FC, \textit{direct})
  & $6.49 \pm 0.63$ & $5.94 \pm 0.85$ & $5.38 \pm 0.43$ & $4.25 \pm 0.40$ \\
& \textsc{BeyondPrompts} (BP, \textit{memory})
  & $6.45 \pm 0.53$ & $5.36 \pm 0.15$ & $4.23 \pm 0.03$ & $4.17 \pm 0.76$ \\
\cmidrule(lr){2-6}
& \textsc{Oblivion} (\textit{ours})
  & $\mathbf{7.91} \pm 0.21$ & $\mathbf{7.75} \pm 0.03$ & $\mathbf{6.10} \pm 0.02$ & $\mathbf{5.30} \pm 0.73$ \\
\midrule
\multirow{3}{*}{\textbf{GPT-4o-mini}}
& \textsc{FullCTX} (FC, \textit{direct})
  & $7.22 \pm 0.15$ & $6.14 \pm 1.04$ & $5.65 \pm 0.15$ & $4.81 \pm 0.57$ \\
& \textsc{BeyondPrompts} (BP, \textit{memory})
  & $8.87 \pm 0.13$ & $6.37 \pm 0.53$ & $4.87 \pm 0.15$ & $4.93 \pm 0.49$ \\
\cmidrule(lr){2-6}
& \textsc{Oblivion} (\textit{ours})
  & $\mathbf{9.00} \pm 0.31$ & $\mathbf{7.37} \pm 0.13$ & $\mathbf{6.26} \pm 0.68$ & $\mathbf{5.63} \pm 0.24$ \\
\midrule
\multirow{3}{*}{\textbf{GPT-4.1-mini}}
& \textsc{FullCTX} (FC, \textit{direct})
  & $8.11 \pm 0.15$ & $6.29 \pm 0.25$ & $6.05 \pm 0.15$ & $6.42 \pm 0.89$ \\
& \textsc{BeyondPrompts} (BP, \textit{memory})
  & $8.53 \pm 0.15$ & $5.37 \pm 0.11$ & $6.68 \pm 0.15$ & $6.26 \pm 0.04$ \\
\cmidrule(lr){2-6}
& \textsc{Oblivion} (\textit{ours})
  & $\mathbf{9.20} \pm 0.16$ & $\mathbf{8.41} \pm 0.13$ & $\mathbf{6.79} \pm 0.16$ & $\mathbf{6.70} \pm 0.11$ \\
\bottomrule
\end{tabular}
\vspace{-0.5em}
\caption{\textsc{GoodAILTM} evaluation results with standard deviation ($\sigma$) across runs. The normalized sum score ($\mu$) is reported out of (11, 10, 11, 11) for (Isolated, 2K, 32K, 120K) settings respectively, where \textsc{FullCTX} (FC) is a full-context \textit{direct} baseline and \textsc{BeyondPrompts} (BP) is \textit{memory}-augmented agent baseline. \dag~Models use \texttt{instruct} variants.}
\label{tab:goodai_ltm_results_extended}
\end{table*}



This section presents supplementary analyses of the \textsc{GoodAILTM} results reported in \autoref{tab:goodai_ltm_results_extended}.
Throughout, we vary \textsc{Oblivion}'s four control hyperparameters: the memory buffer size $B$, the decay temperature $T$, and binary flags for \textit{Dynamic Topics} (creation of new clustering topics beyond the initial taxonomy) and \textit{Memory Linking} (bidirectional linking across $L_2$--$L_3$ memory layers), exploring $B_t \in \{60, 90\}$ and $T \in \{1, 3, 5, 10, 20, 50\}$.
We provide five ablation studies conducted with \texttt{GPT-4.1-mini}, examining: memory hierarchy configurations (\autoref{tab:goodai_memory_config}), episodic memory ($L_3$) content type (\autoref{tab:table_goodailtm_llm_vs_heuristic_episodic_memory_type_comparison}), episode length (\autoref{tab:goodai_episode_length}), decay temperature (\autoref{tab:goodai_decay_temp}, \autoref{fig:ebbinghaus_decay_ablation}, and \autoref{fig:2k_retention_ablation}), and topic initialization (\autoref{tab:goodai_topic_init}), followed by an operational cost and latency analysis (\autoref{tab:efficiency_extended}).

\subsection{Effect of Memory Configurations}
\label{app:goodailtm_memory_config}

\setlength{\tabcolsep}{3.5pt}
\begin{table*}[ht]
\centering
\scriptsize

\begin{tabular}{l ccccc ccccc}
\toprule
& \multicolumn{5}{c}{\textbf{Isolated}} & \multicolumn{5}{c}{\textbf{Non-isolated (2K)}} \\
\cmidrule(lr){2-6}\cmidrule(lr){7-11}
\textbf{Memory Config} &
\textbf{30} & \textbf{60} & \textbf{90} & \textbf{120} & \textbf{Unbounded} &
\textbf{30} & \textbf{60} & \textbf{90} & \textbf{120} & \textbf{Unbounded} \\
\midrule

$L_1$                  & 8.20 & 8.76 & 8.46 & 8.69 & 8.44 & 6.04 & 6.86 & 5.55 & 5.98 & 6.32 \\
$L_2$                  & 8.64 & 8.69 & 8.91 & 8.98 & 9.16 & 6.96 & 6.14 & 6.80 & 6.22 & 6.22 \\
$L_3$                  & 8.58 & 8.64 & 8.29 & 8.73 & 8.07 & 5.05 & 6.00 & 6.05 & 5.72 & 5.62 \\
\midrule
$L_1$+$L_2$               & 8.97 & 8.96 & 8.84 & 8.86 & 8.91 & 5.93 & 6.84 & 6.73 & 5.84 & 7.22 \\
$L_1$+$L_3$               & 8.48 & 7.60 & 8.64 & 8.60 & 9.01 & 6.61 & 5.88 & 6.77 & 6.40 & 6.77 \\
$L_2$+$L_3$         & 9.00 & 9.16 & 8.98 & 9.02 & 9.12 & 6.92 & 6.87 & 6.28 & 7.17 & 6.81 \\
\midrule
$L_1$+$L_2$+$L_3$      & 8.87 & 9.11 & \textbf{9.20} & 8.92 & 9.16 & 7.07 & \textbf{8.41} & 7.46 & 7.72 & 7.83 \\

\bottomrule
\end{tabular}
\caption{Effect of memory hierarchy configuration and executor memory capacity on \textsc{GoodAILTM} with \texttt{GPT-4.1-mini}.
Column headers (30, 60, 90, 120, Unbounded) indicate the maximum number of memories supplied to the executor.
$L_1$/$L_2$/$L_3$ denote dynamic procedural, semantic, and preemptive episodic memory levels.} 
\label{tab:goodai_memory_config}
\end{table*}

\textsc{Oblivion} assembles the Executor's working context from a three-level memory hierarchy under a bounded retrieval budget; isolating how each level and the budget size contribute clarifies which components drive downstream performance.
\autoref{tab:goodai_memory_config} therefore varies the combination of memory hierarchy levels ($L_1$ dynamic procedural, $L_2$ semantic, $L_3$ preemptive episodic, with optional linked retrieval) and the maximum number of memories supplied to the Executor, using \texttt{GPT-4.1-mini} on both the Isolated and Non-Isolated 2K settings, extending the five-mode ablation in \autoref{tab:ablation_levels} to a fine-grained grid of seven configurations across five capacity levels.

We observe that multi-level configurations generally outperform single-level ones, and
among single levels, $L_2$ (semantic) performs best, reaching 9.16 in the Isolated setting at unlimited capacity.
Further, the full configuration $L_1$+$L_2$+$L_3$ achieves the strongest scores in both the Isolated and 2K settings, confirming that the complete memory hierarchy is the most effective configuration.

Further, we observe that no single buffer capacity dominates across all configurations for our system.
The unbounded capacity setting does not consistently outperform the bounded ones.
This is consistent with the observation in \autoref{subsec:hierarchical_memory_control} that excess retrieved context can dilute signal quality.

\subsection{Effect of Episodic Data Variation}
\label{app:goodailtm_episodic_data}

\begin{table*}[h]
\centering
\footnotesize
\setlength{\tabcolsep}{3pt}
\begin{tabular}{@{}ccc crr crr@{}}
\toprule
 & & & \multicolumn{3}{c}{\textbf{Preemptive LLM-based Episodic}} & \multicolumn{3}{c}{\textbf{Raw Input Data Episodic}} \\
\cmidrule(lr){4-6} \cmidrule(lr){7-9}
\textbf{Dynamic Topics} & \textbf{$B$} & \textbf{$T$} & \textbf{Mean $\pm$ Std.~Dev.} & \textbf{Toks\,(K)} & \textbf{Latency\,(ms)} & \textbf{Mean $\pm$ Std.~Dev.} & \textbf{Toks\,(K)} & \textbf{Latency\,(ms)} \\
\midrule
\multicolumn{9}{c}{\textbf{Isolated (11)}} \\
\midrule
True  & 60 & 1  & $8.52 \pm 0.08$ & 14.5 & 2592 & $9.34 \pm 0.27$ & 14.5 & 2030 \\
True  & 60 & 5  & $8.54 \pm 0.39$ & 14.5 & 2742 & $9.33 \pm 0.33$ & 13.9 & 1946 \\
True  & 60 & 10 & $8.35 \pm 0.42$ & 14.3 & 2545 & $9.50 \pm 0.16$ & 13.5 & 1835 \\
True  & 90 & 1  & $8.77 \pm 0.71$ & 14.5 & 2527 & $9.49 \pm 0.22$ & 13.5 & 1971 \\
True  & 90 & 5  & $8.80 \pm 0.44$ & 14.5 & 2434 & $9.21 \pm 0.35$ & 13.8 & 2158 \\
True  & 90 & 10 & $8.57 \pm 0.42$ & 14.3 & 2273 & $9.40 \pm 0.08$ & 13.6 & 2385 \\
\cmidrule(lr){1-9}
False & 60 & 1  & $8.64 \pm 0.22$ & 14.5 & 2198 & $9.43 \pm 0.06$ & 14.1 & 2056 \\
False & 60 & 5  & $8.89 \pm 0.13$ & 14.5 & 2256 & $\mathbf{9.57 \pm 0.35}$ & 14.1 & 2118 \\
False & 60 & 10 & $8.47 \pm 0.35$ & 14.4 & 2348 & $9.49 \pm 0.38$ & 14.1 & 1992 \\
False & 90 & 1  & $8.59 \pm 0.33$ & 14.5 & 2551 & $9.54 \pm 0.24$ & 14.0 & 2110 \\
False & 90 & 5  & $\mathbf{9.20 \pm 0.16}$ & 14.5 & 2494 & $9.20 \pm 0.36$ & 14.1 & 2322 \\
False & 90 & 10 & $8.51 \pm 0.35$ & 14.5 & 2434 & $9.38 \pm 0.21$ & 14.1 & 2348 \\
\midrule
\multicolumn{9}{c}{\textbf{2K (10)}} \\
\midrule
True  & 60 & 1  & $\mathbf{8.41 \pm 0.13}$ & 15.4 & 1940 & $\mathbf{7.28 \pm 0.19}$ & 7.7 & 1867 \\
True  & 60 & 5  & $7.10 \pm 0.97$ & 15.7 & 2057 & $5.97 \pm 0.16$ & 7.7 & 1663 \\
True  & 60 & 10 & $6.36 \pm 0.77$ & 15.9 & 1962 & $7.32 \pm 0.17$ & 7.6 & 1608 \\
True  & 90 & 1  & $6.80 \pm 0.02$ & 15.4 & 1959 & $7.06 \pm 0.63$ & 7.6 & 1464 \\
True  & 90 & 5  & $7.33 \pm 0.67$ & 16.3 & 2356 & $6.89 \pm 0.26$ & 7.6 & 1271 \\
True  & 90 & 10 & $7.18 \pm 0.29$ & 16.1 & 2186 & $6.44 \pm 0.43$ & 7.7 & 1222 \\
\cmidrule(lr){1-9}
False & 60 & 1  & $7.26 \pm 0.36$ & 15.1 & 1623 & $6.72 \pm 0.20$ & 7.6 & 1674 \\
False & 60 & 5  & $7.28 \pm 0.31$ & 15.6 & 1795 & $7.02 \pm 0.72$ & 7.7 & 1768 \\
False & 60 & 10 & $7.36 \pm 0.15$ & 16.1 & 1658 & $6.52 \pm 0.31$ & 7.9 & 1650 \\
False & 90 & 1  & $6.70 \pm 1.10$ & 15.4 & 1880 & $6.33 \pm 0.25$ & 7.7 & 1426 \\
False & 90 & 5  & $7.15 \pm 0.92$ & 15.5 & 1899 & $6.60 \pm 0.54$ & 7.8 & 1303 \\
False & 90 & 10 & $7.13 \pm 0.40$ & 15.5 & 1985 & $6.51 \pm 0.34$ & 7.8 & 1224 \\
\bottomrule
\end{tabular}
\caption{
Hyperparameter search comparing \textsc{Preemptive LLM-based} and \textsc{Raw Input Data} episodic memory mechanisms on \textsc{GoodAILTM} using \texttt{GPT-4.1-mini}. Reported values represent the mean ($\mu$) and standard deviation ($\sigma$) of normalized sum scores, with maximums of 11 (Isolated) and 10 (2K). Here, $B$ denotes maximum buffer size, $T$ is the decay temperature, and \textit{Dynamic Topics} indicates whether the taxonomy is expanded during clustering. Bidirectional memory linking is enabled across all configurations. Toks and Latency represent average interaction tokens (thousands) and total execution time (milliseconds), respectively. The highest performance in each column is emphasized for clarity.
}
\label{tab:table_goodailtm_llm_vs_heuristic_episodic_memory_type_comparison}
\end{table*}
The preemptive episodic level ($L_3$) can be populated either with raw interaction turns or with LLM-summarized episodic entries, and the two mechanisms trade retrieval accuracy against token cost; selecting between them requires quantifying this trade-off.
\autoref{tab:table_goodailtm_llm_vs_heuristic_episodic_memory_type_comparison} compares the two episodic memory mechanisms across buffer sizes, decay temperatures, and \textit{Dynamic Topics} settings, on the Isolated and 2K settings with \texttt{GPT-4.1-mini}.

We observe that raw input episodic memory type achieves higher scores across nearly all Isolated configurations with comparable token usage.
Further, the preemptive episodic entries type generally dominate on more complex interleaved 2K settings despite ${\sim}2{\times}$ higher token consumption
(15.4K vs.\ 7.7K tokens/interaction).
Hence, owing to higher reliability and consistency of episodic memory performance with preemptive memory entries, we select this alternative, which is also cost-effective compared to baselines at scale as shown in \autoref{tab:efficiency_extended}.

\subsection{Effect of Episode Length}
\label{app:goodailtm_episode}

\setlength{\tabcolsep}{6.0pt}
\begin{table}[h]
\centering
\scriptsize

\begin{tabular}{c cc}
\toprule
\textbf{Episode Length} & \textbf{Isolated} & \textbf{2K} \\
\midrule

\textbf{2}  & 8.16          & 7.08          \\
\textbf{4}  & \textbf{9.20} & \textbf{8.41} \\
\textbf{6}  & 9.03          & 7.34          \\
\textbf{8}  & 8.92          & 7.17          \\
\textbf{10} & 8.68          & 7.11          \\

\bottomrule
\end{tabular}
\caption{Effect of $L_3$ episode length on \textsc{GoodAILTM} with \texttt{GPT-4.1-mini}.}
\label{tab:goodai_episode_length}
\end{table}

The preemptive episodic level chunks interactions into fixed-length episodes, so the chunk granularity trades contextual completeness against retrieval precision; the optimal length is not obvious a priori.
\autoref{tab:goodai_episode_length} reports the effect of varying $L_3$ episode length on \textsc{GoodAILTM} performance with \texttt{GPT-4.1-mini}.
The episode length of 4 turns yields the best scores in both the Isolated (9.20) and 2K (8.41) settings, outperforming the next-best values by $+$0.17 (vs.\ ep\,=\,6 on Isolated) and $+$1.07 (vs.\ ep\,=\,6 on 2K).
Further, shorter episodes (ep\,=\,2) show notably lower scores ($-$1.04 on Isolated, $-$1.33 on 2K relative to ep\,=\,4), while longer episodes (ep\,$\geq$\,6) cluster within a relatively narrower range.
This result is in agreement with the \textsc{LongMemEval} episode-length analysis (\autoref{app:episode_length}), where ep\,=\,4 also achieves the highest accuracy.

\subsection{Effect of Decay Temperature}
\label{app:goodailtm_decay}

\setlength{\tabcolsep}{6.0pt}
\begin{table}[h]
\centering
\scriptsize

\begin{tabular}{l cc}
\toprule
\textbf{Decay Temperature} & \textbf{60} & \textbf{90} \\
\midrule

$T=1$   & \textbf{8.41} & 6.74 \\
$T=3$   & 6.97 & 7.16 \\
$T=5$   & 6.69 & 6.74 \\
$T=10$  & 7.78 & \textbf{7.46} \\
$T=20$  & 6.88 & 7.41 \\
$T=50$  & 6.55 & 7.16 \\

\bottomrule
\end{tabular}
\caption{Effect of decay temperature $T$ on \textsc{GoodAILTM} (Non-isolated 2K) with \texttt{GPT-4.1-mini}.
The two columns correspond to Executor memory buffer capacities ($B_t$) of 60 and 90.}
\label{tab:goodai_decay_temp}
\end{table}

The decay temperature $T$ governs how quickly unreinforced memories lose accessibility and is the central hyperparameter of \textsc{Oblivion}'s decay-driven activation mechanism; characterizing its effect on task performance is therefore essential.
\autoref{tab:goodai_decay_temp} reports \textsc{Oblivion}'s performance on the Non-isolated 2K setting with \texttt{GPT-4.1-mini} across six decay temperatures ($T \in \{1, 3, 5, 10, 20, 50\}$) and two memory buffer configurations ($B_t\in \{60, 90\}$).

\begin{figure}[p]
\centering
\begin{subfigure}[b]{0.48\textwidth}
    \centering
    \includegraphics[width=\textwidth]{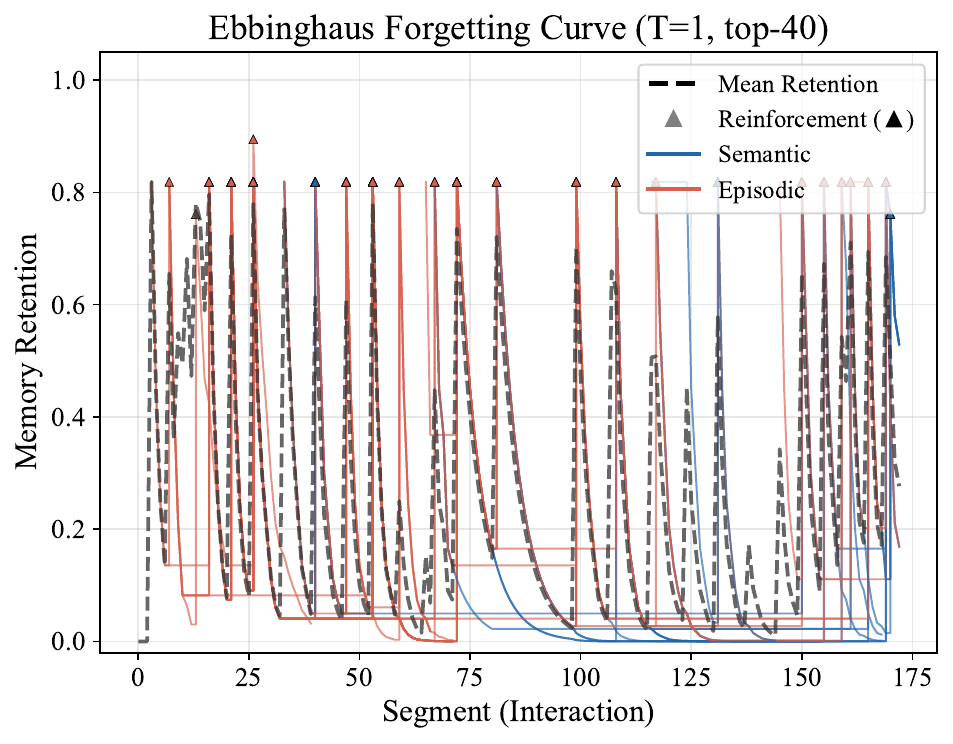}
    \caption{$T=1$}
    \label{fig:ebbinghaus_T1}
\end{subfigure}
\hfill
\begin{subfigure}[b]{0.48\textwidth}
    \centering
    \includegraphics[width=\textwidth]{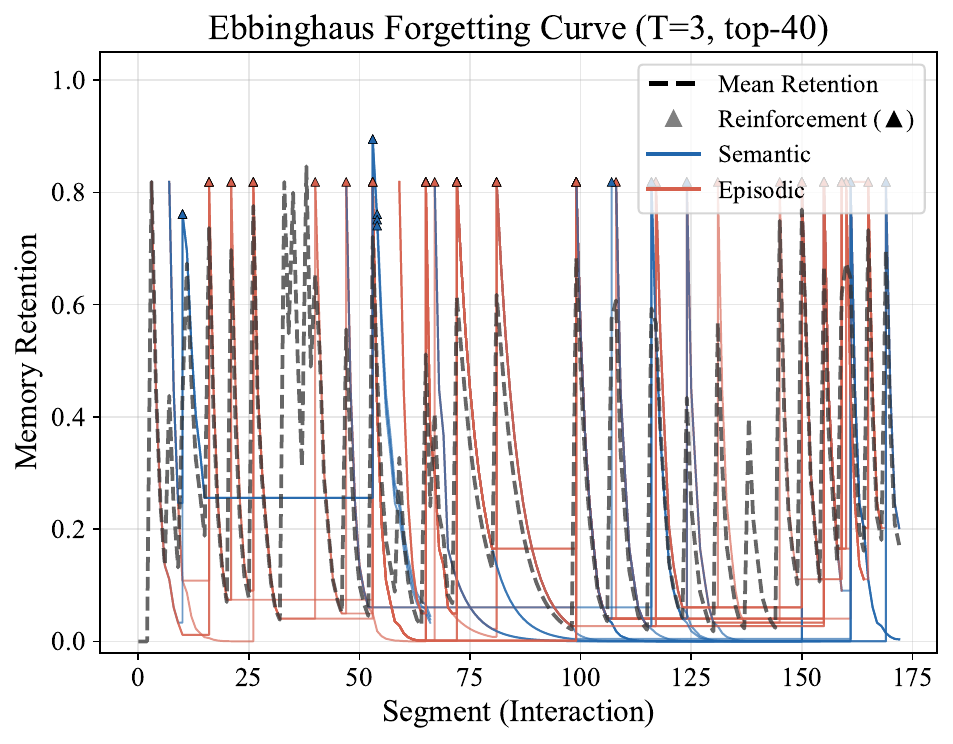}
    \caption{$T=3$}
    \label{fig:ebbinghaus_T3}
\end{subfigure}

\vspace{0.5em}

\begin{subfigure}[b]{0.48\textwidth}
    \centering
    \includegraphics[width=\textwidth]{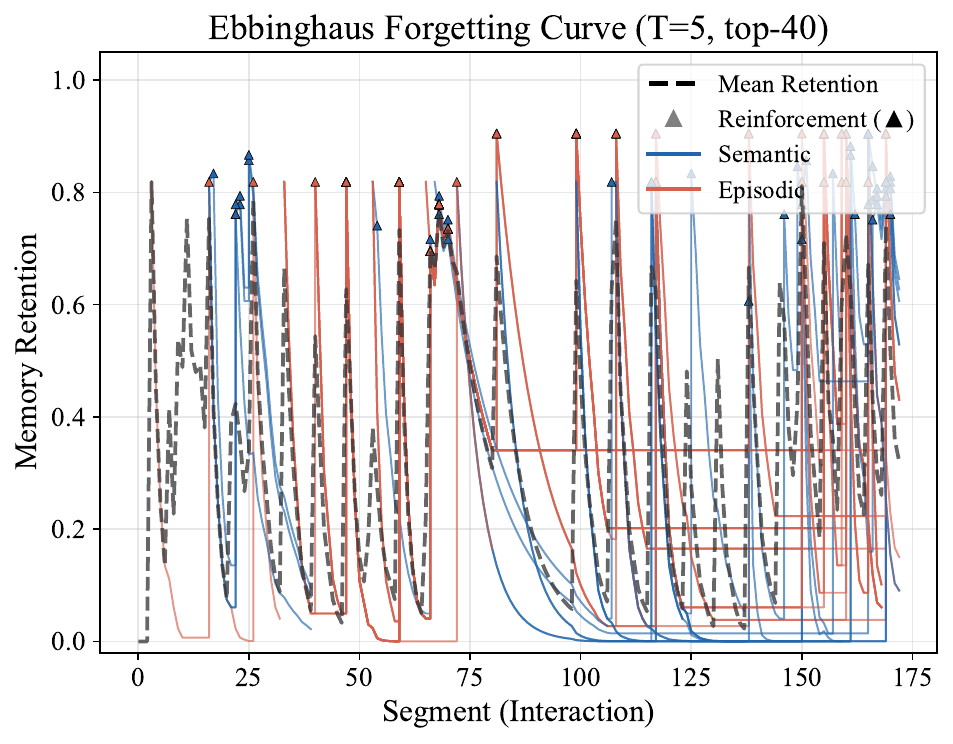}
    \caption{$T=5$}
    \label{fig:ebbinghaus_T5}
\end{subfigure}
\hfill
\begin{subfigure}[b]{0.48\textwidth}
    \centering
    \includegraphics[width=\textwidth]{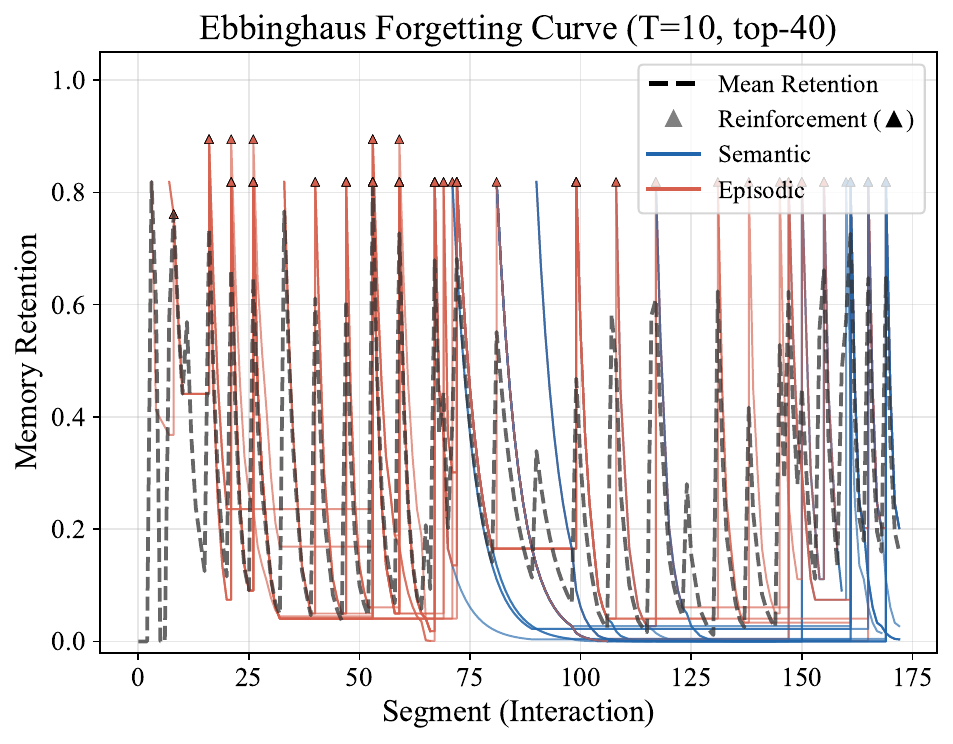}
    \caption{$T=10$}
    \label{fig:ebbinghaus_T10}
\end{subfigure}

\vspace{0.5em}

\begin{subfigure}[b]{0.48\textwidth}
    \centering
    \includegraphics[width=\textwidth]{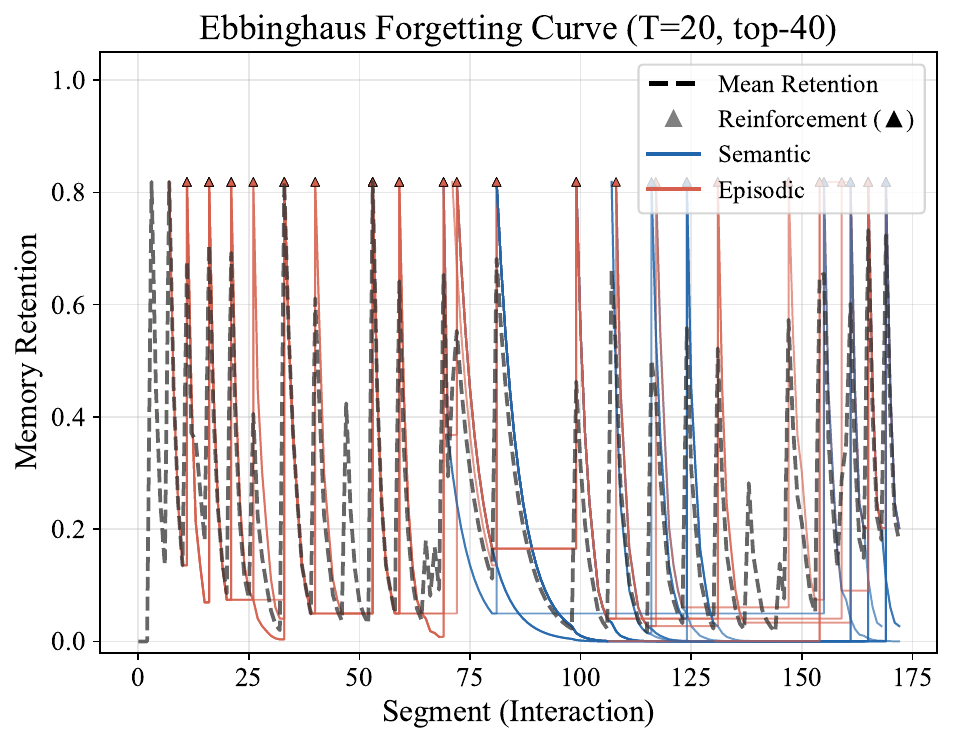}
    \caption{$T=20$}
    \label{fig:ebbinghaus_T20}
\end{subfigure}
\hfill
\begin{subfigure}[b]{0.48\textwidth}
    \centering
    \includegraphics[width=\textwidth]{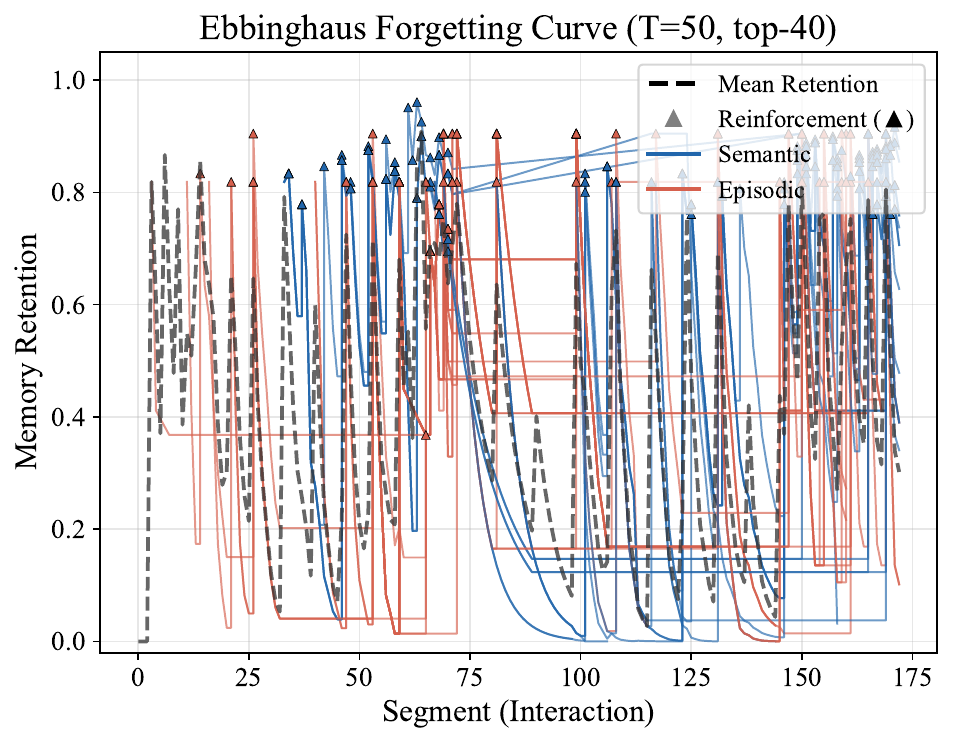}
    \caption{$T=50$}
    \label{fig:ebbinghaus_T50}
\end{subfigure}

\caption{Ebbinghaus forgetting curves with reinforcement trends for top-40 memories including both episodic and semantic memory types with temperature configurations ($T \in \{1, 3, 5, 10, 20, 50\}$) given the 2K setting for the memory buffer configuration $B_t=90$.}
\label{fig:ebbinghaus_decay_ablation}
\end{figure}

\begin{figure}[p]
\centering
\captionsetup[subfigure]{font=footnotesize,skip=3pt}
\setlength{\abovecaptionskip}{5pt}
\begin{subfigure}[c]{0.40\textwidth}
    \centering
    \includegraphics[width=\textwidth,clip=false]{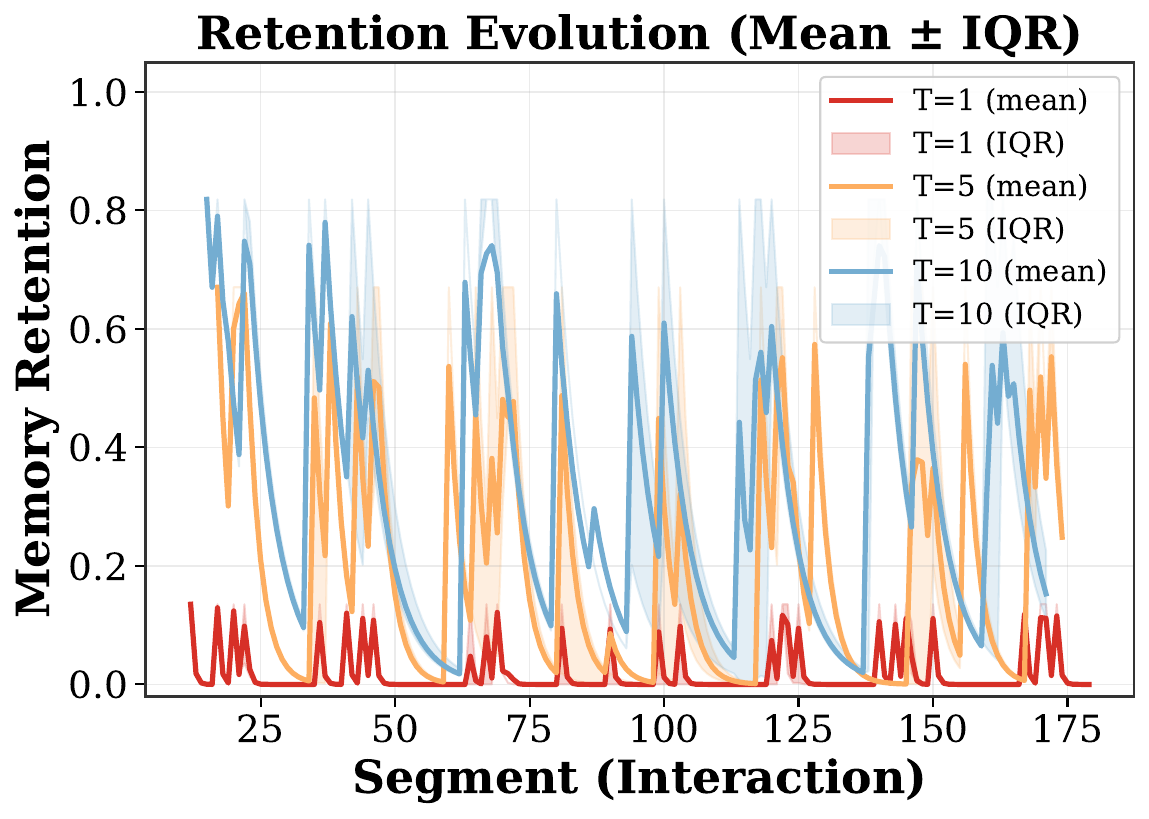}
    \caption{\protect\centering Retention Evolution for \texttt{GPT-4.1-mini}}
    \label{fig:2k_gpt41_evo}
\end{subfigure}
\hfill
\begin{subfigure}[c]{0.57\textwidth}
    \centering
    \includegraphics[width=\textwidth,clip=false]{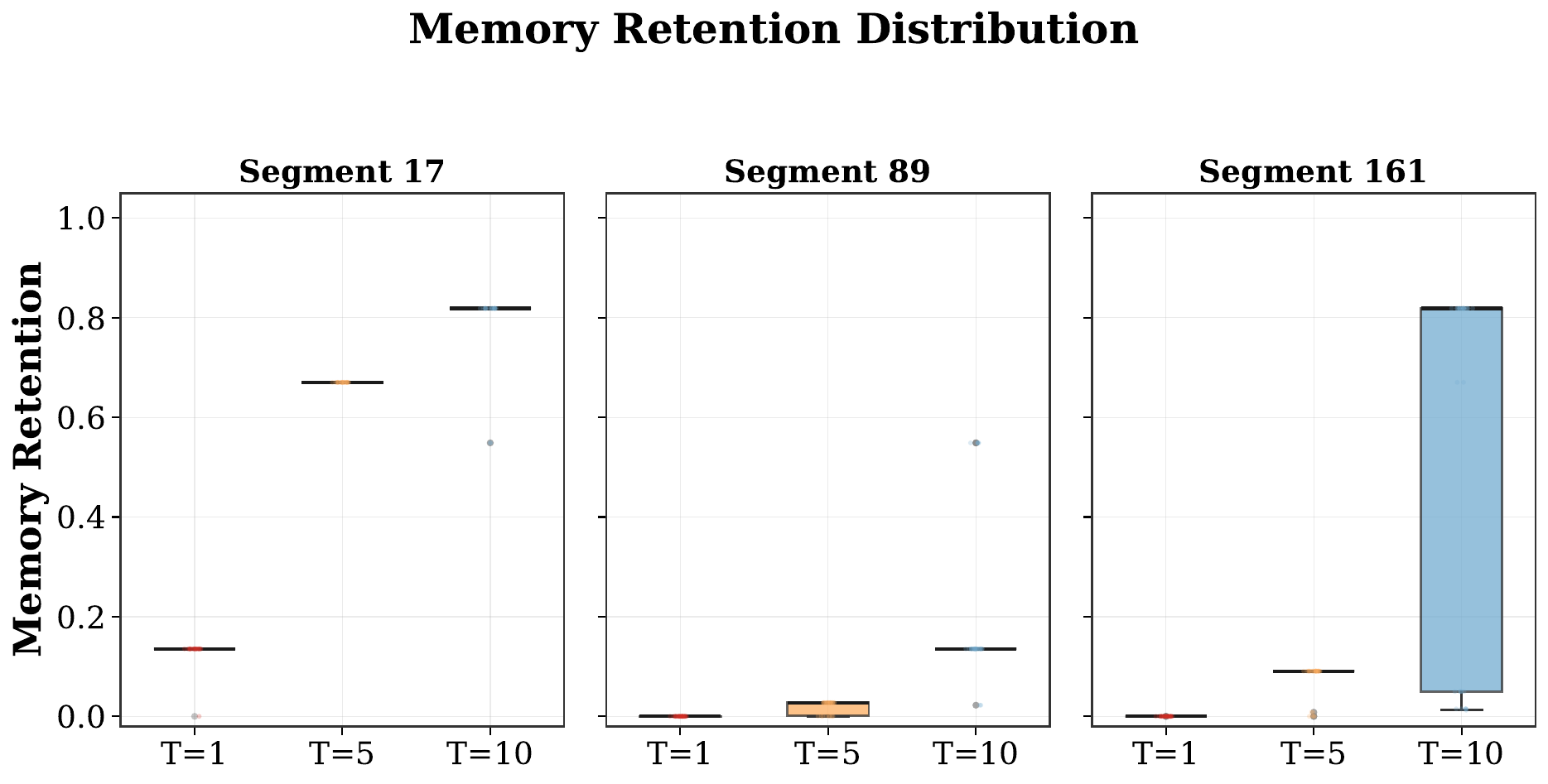}
    \caption{\protect\centering Retention Distribution for \texttt{GPT-4.1-mini}}
    \label{fig:2k_gpt41_dist}
\end{subfigure}

\vspace{0.2em}

\begin{subfigure}[c]{0.40\textwidth}
    \centering
    \includegraphics[width=\textwidth,clip=false]{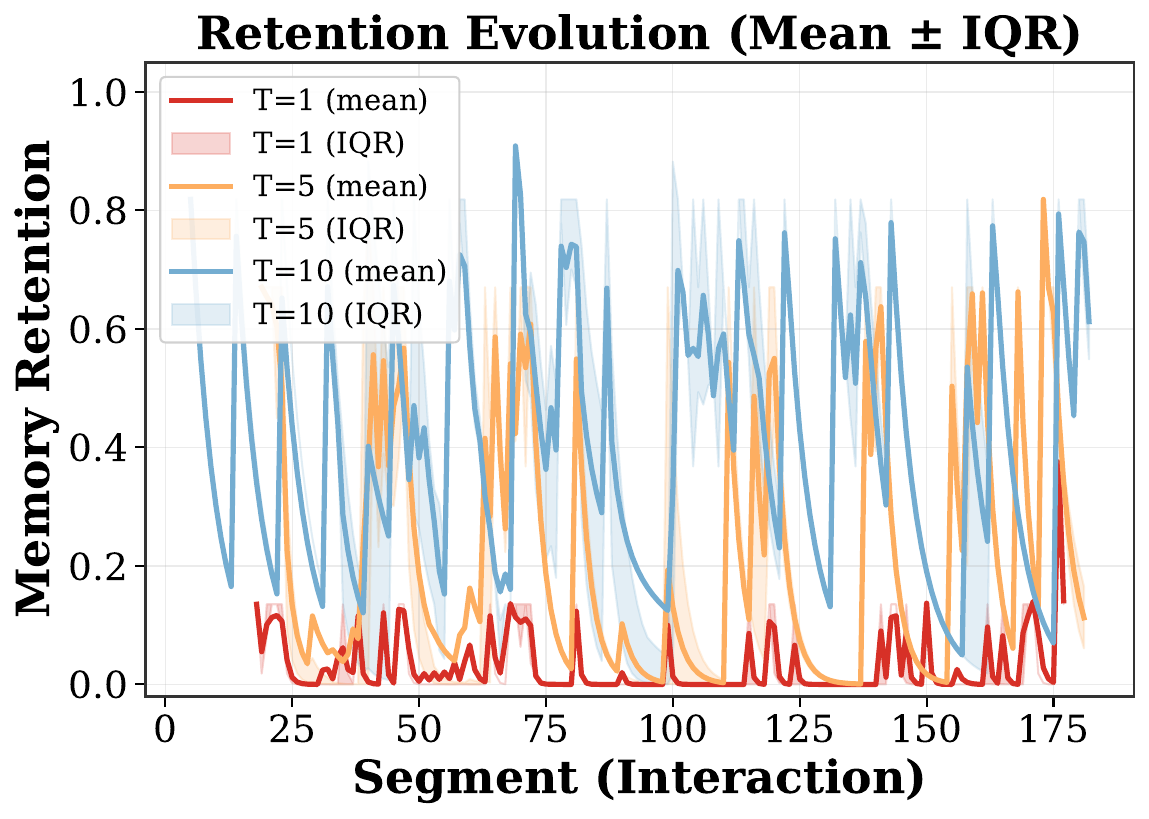}
    \caption{\protect\centering Retention Evolution for \texttt{GPT-4o-mini}}
    \label{fig:2k_gpt4o_evo}
\end{subfigure}
\hfill
\begin{subfigure}[c]{0.57\textwidth}
    \centering
    \includegraphics[width=\textwidth,clip=false]{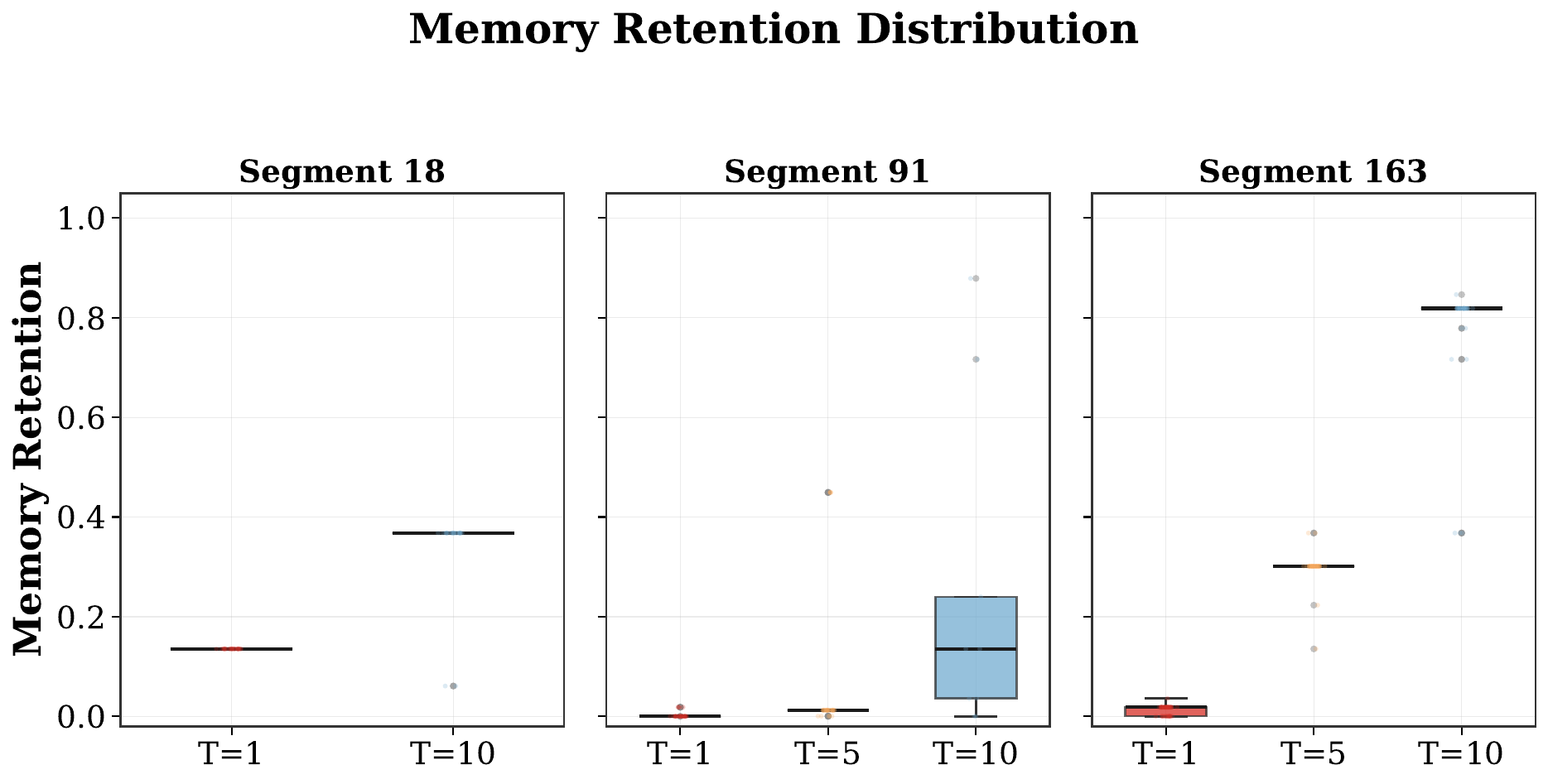}
    \caption{\protect\centering Retention Distribution for \texttt{GPT-4o-mini}}
    \label{fig:2k_gpt4o_dist}
\end{subfigure}

\vspace{0.2em}

\begin{subfigure}[c]{0.40\textwidth}
    \centering
    \includegraphics[width=\textwidth,clip=false]{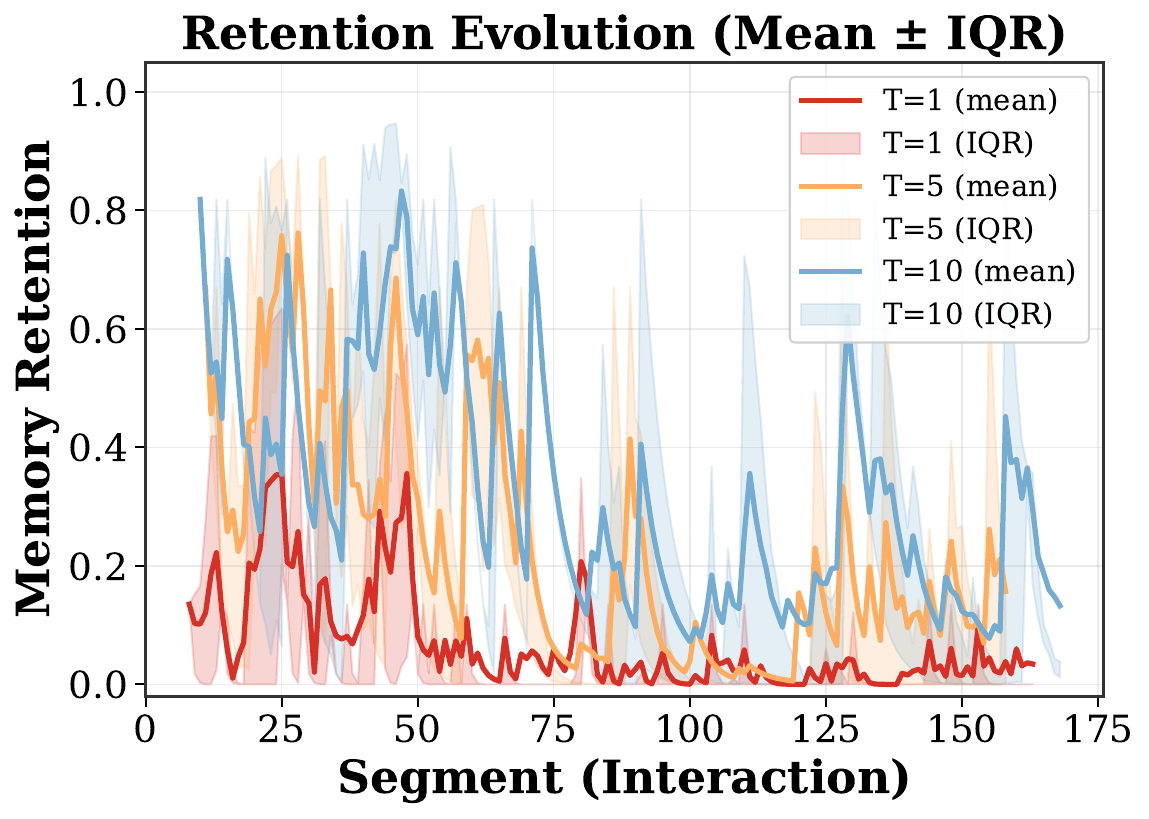}
    \caption{\protect\centering Retention Evolution for \texttt{Phi-4-mini-instruct-4B}}
    \label{fig:2k_phi4_evo}
\end{subfigure}
\hfill
\begin{subfigure}[c]{0.57\textwidth}
    \centering
    \includegraphics[width=\textwidth,clip=false]{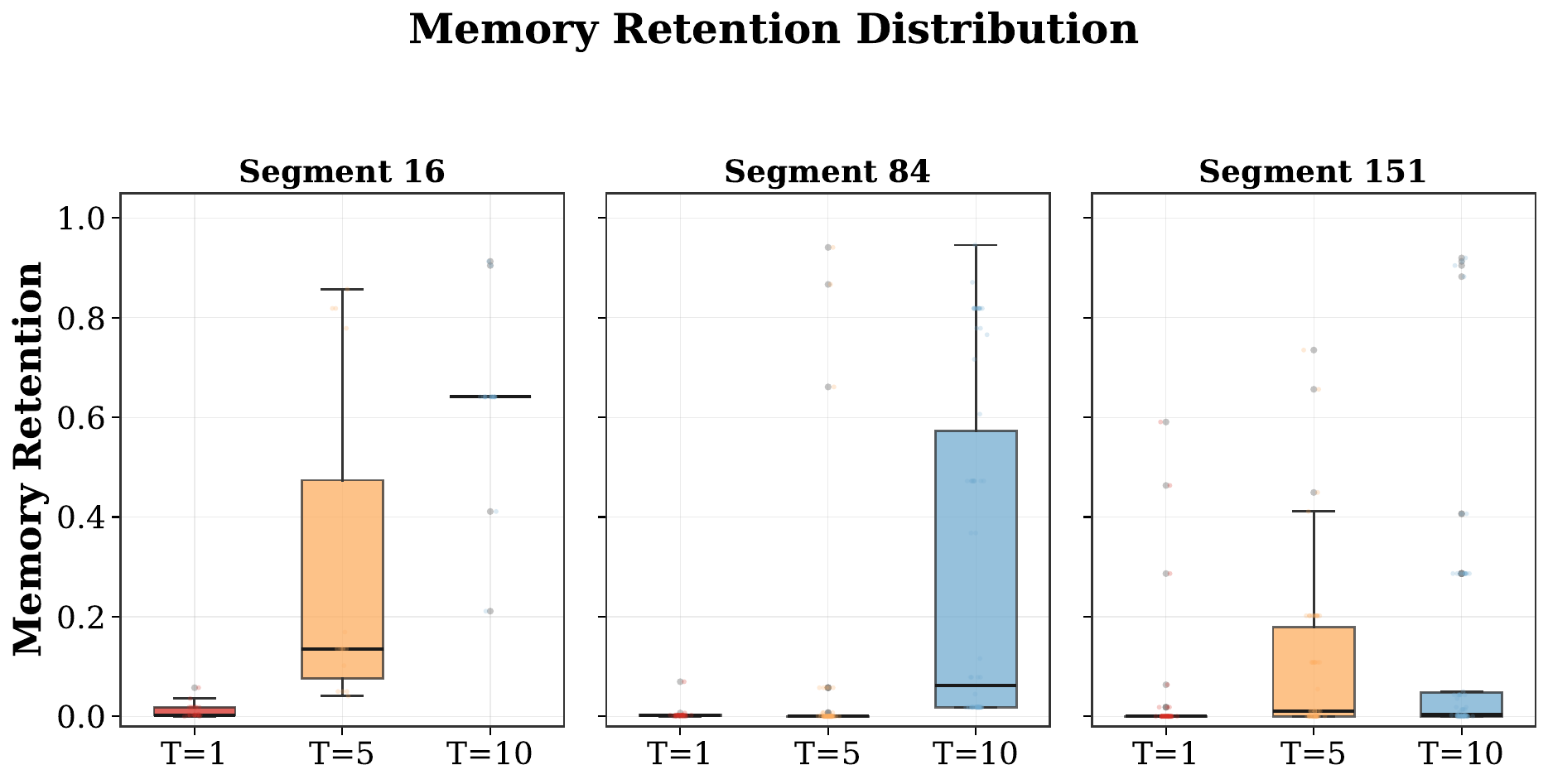}
    \caption{\protect\centering Retention Distribution for \texttt{Phi-4-mini-instruct-4B}}
    \label{fig:2k_phi4_dist}
\end{subfigure}

\vspace{0.2em}

\begin{subfigure}[c]{0.40\textwidth}
    \centering
    \includegraphics[width=\textwidth,clip=false]{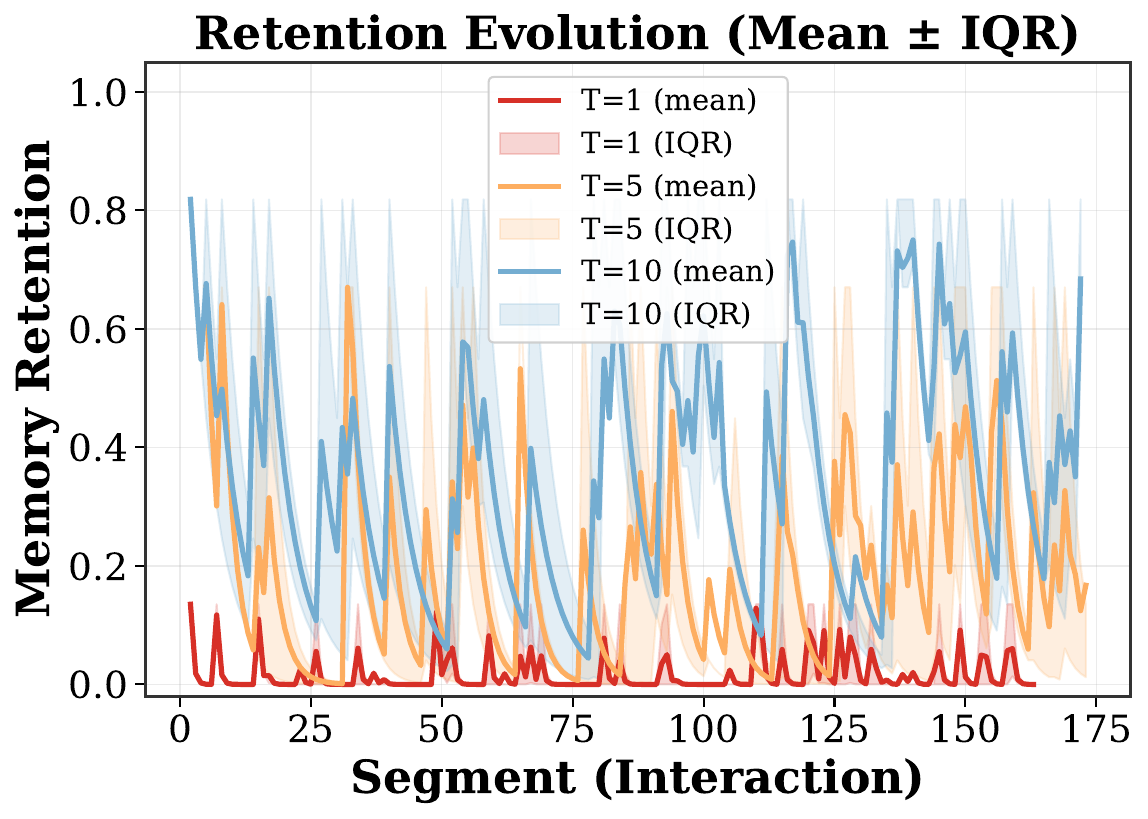}
    \caption{\protect\centering Retention Evolution for \texttt{Qwen3-30B-A3B-Instruct}}
    \label{fig:2k_qwen3_evo}
\end{subfigure}
\hfill
\begin{subfigure}[c]{0.57\textwidth}
    \centering
    \includegraphics[width=\textwidth,clip=false]{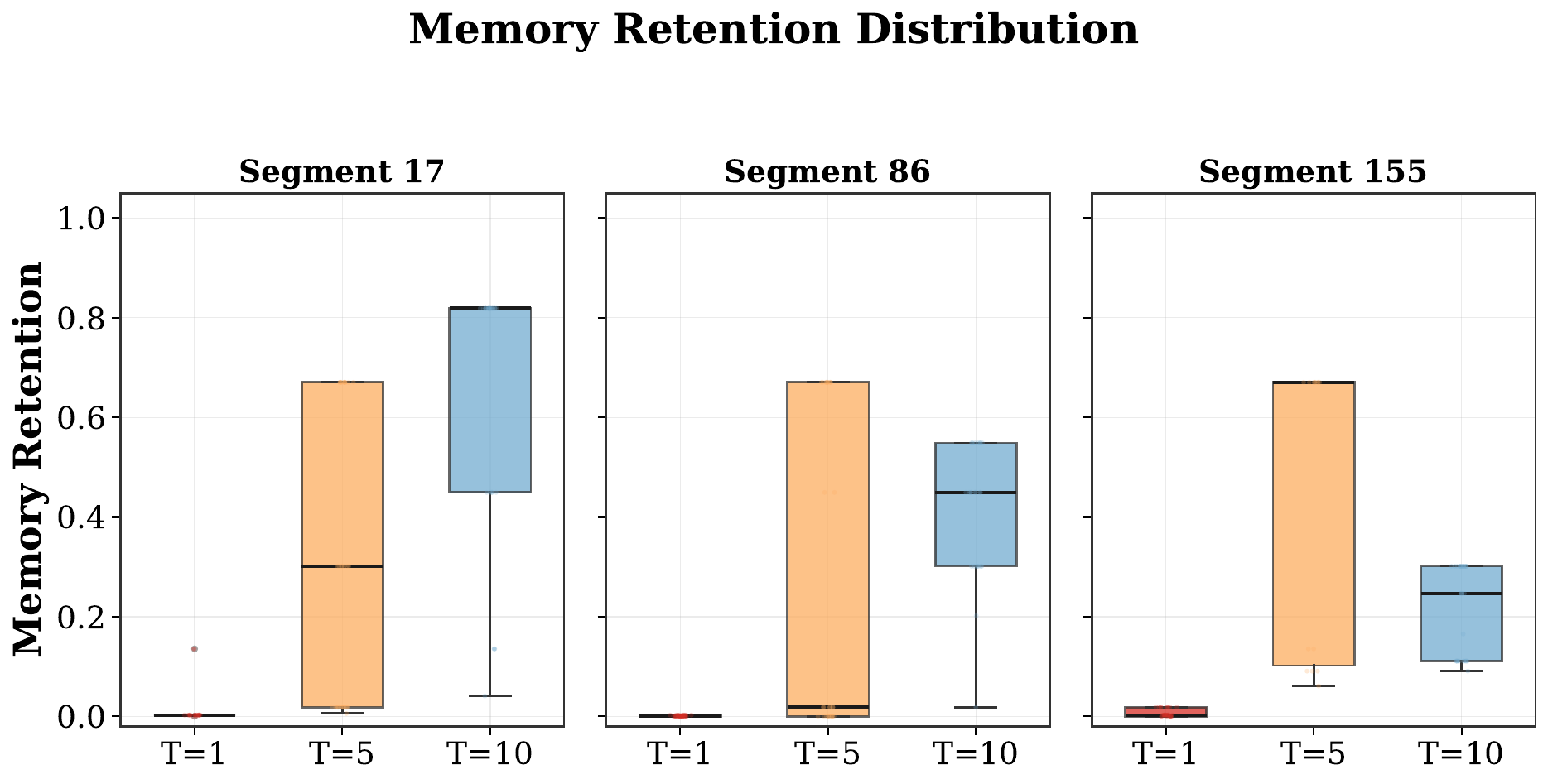}
    \caption{\protect\centering Retention Distribution for \texttt{Qwen3-30B-A3B-Instruct}}
    \label{fig:2k_qwen3_dist}
\end{subfigure}

\caption{Retention evolution and distribution comparison given temperature ablation ($T \in \{1, 5, 10\}$) in the 2K memory span test evaluations for $B_t=60$. Increasing the decay temperature from $T=1$ to $T=10$ increases retention across all four models, with $T=10$ producing periodic reinforcement-driven spikes reaching 0.6--0.8 mean retention, whereas $T=1$ yields near-zero recall throughout.
Further, \texttt{GPT-4.1-mini} exhibits the tightest IQR at $T=10$ indicating consistent recall behavior, whereas \texttt{Phi-4-mini-instruct-4B} and \texttt{Qwen3-30B-A3B-Instruct} display wider distributional spread reflecting higher inter-memory variability under the same temperature. The boxplots confirm that temperature-induced retention gains persist across models, although their extent varies.
}
\label{fig:2k_retention_ablation}
\end{figure}

For memory buffer capacity $B_t=90$, \autoref{fig:ebbinghaus_decay_ablation} further visualizes the corresponding Ebbinghaus-style retention curves for the top-40 memories, colored by memory class.
Together, these complement the qualitative retention-reinforcement analysis in \autoref{subsec:memory_decay_behavior} with benchmark-level scores and per-temperature decay profiles for $B_t=90$.
The $T=10$ yields the best score at buffer configuration $B_t=90$, consistent with the retention--reinforcement balance identified in~\autoref{fig:decay_temperature_behavior_analysis}.
However, the relationship between $T$ and performance is not monotonic: for instance, given the memory buffer capacity $B_t=60$, $T=1$ (8.41) outperforms all other listed configurations.
The overall normalized score range across all temperatures and buffer configurations is 6.55--8.41, a spread of 1.86 points.
For $B_t=90$, scores remain within a narrow band (7.16--7.46) for $T \geq 10$, suggesting that the system is not overly sensitive to $T$ at higher magnitudes.
This trend is corroborated by the retention curves in~\autoref{fig:ebbinghaus_decay_ablation}: at low $T$ (e.g., $T=1$), memory accessibility drops sharply after a few interactions, whereas at higher $T$ (e.g., $T=50$), the curves flatten considerably and memories remain accessible over longer horizons.
\autoref{fig:2k_retention_ablation} further shows that, even under identical hyperparameters, different model architectures reinforce memories to different extents: \texttt{GPT-4.1-mini} exhibits the tightest inter-quartile range across temperatures, indicating more consistent retention than the other backbones. Similar cross-model retention patterns are observed at the larger 32K memory span.

\clearpage
\subsection{Effect of Topic Initialization}
\label{app:goodailtm_topic_init}

\setlength{\tabcolsep}{5.0pt}
\begin{table}[h]
\centering
\scriptsize

\begin{tabular}{l cccc}
\toprule
& \multicolumn{2}{c}{\textbf{Non-isolated (2K)}} & \multicolumn{2}{c}{\textbf{Isolated}} \\
\cmidrule(lr){2-3}\cmidrule(lr){4-5}
\textbf{Initialization} & \textbf{Static} & \textbf{Dynamic} & \textbf{Static} & \textbf{Dynamic} \\
\midrule

\textbf{Default Topics (15) }    & \textbf{7.76} & \textbf{8.41} & \textbf{9.20}          & \textbf{8.97} \\
\textbf{Domain Topics (12) }     & 7.01          & 6.89          & 8.93 & 8.72          \\
\textbf{No Initialization}       &   --           & 7.12 &      --         & 8.45          \\

\bottomrule
\end{tabular}
\caption{Effect of topic initialization on \textsc{GoodAILTM} with \texttt{GPT-4.1-mini}.
\textbf{Static}: the topic set remains fixed after initialization; \textbf{Dynamic}: topics may be added or merged during conversation.
\emph{No Initialization} starts from an empty topic set, so the \textbf{Static} configuration does not apply.}
\label{tab:goodai_topic_init}
\end{table}

\textsc{Oblivion} organizes semantic memories under a clustering topic taxonomy that can be seeded before evaluation; the coverage and adaptivity of this initial taxonomy may shape retrieval quality.
\autoref{tab:goodai_topic_init} examines how the initial memory clustering topic set for \textsc{Oblivion}'s memory organization influences \textsc{GoodAILTM} performance with \texttt{GPT-4.1-mini}.
We test three configurations, namely:
(i) \emph{Default Topics}: supplies 15 general-purpose topics that cover broad conversational categories;
(ii) \emph{Domain Topics}: supplies 12 topics selected to align with the benchmark's task domains;
and (iii) \emph{No Initialization}: starts with an empty topic set and relies on the system to create topics on the fly as new conversations arrive.
For the first two, we further distinguish between a \textbf{Static} mode, where the topic set is locked after initialization, and a \textbf{Dynamic} mode, where topics may be added throughout the interactive evaluation.

\paragraph{Dynamic vs.\ Static Topics.}
We observe that in the harder interleaved 2K setting, allowing topics to evolve during conversation generally improves scores over keeping a fixed set.
In the Isolated setting, switching from Dynamic to Static configuration raises the Default Topics score from 8.97 to 9.20.
This gain likely stems from the system's ability to consistently memorize finer-grained categories, as no new conversational themes appear within a given test evaluation.
But, in the 2K setting, where interleaving task context adds retrieval difficulty, requiring more generalization, the above trend is reversed, where switching from Static to Dynamic configuration raises the Default Topics score from 7.76 to 8.41.

\paragraph{No Initialization.}
Without any predefined topics, the system still reaches 7.12 on the 2K setting and 8.45 on the Isolated setting.
The fact that on-the-fly topic discovery is significantly outperformed by the pre-initialized topics under the harder 2K setting suggests that when retrieval is already challenging, the quality of the initial topic set matters a lot more than the system's overall extraction and retrieval pipeline.
In the relatively easier Isolated setting, again, a well-chosen initial set combined with dynamic updates retains a clear lead (9.20 vs.\ 8.45), indicating that predefined topics provide a structural head start that the system cannot fully recover within a single evaluation pass. Interestingly, we also observe that domain-aligned topics do not necessarily guarantee better results, but rather, the utilization of a well-defined memory topic taxonomy drives better retrieval through better memory organization.

\paragraph{Quality of Auto-Generated Topics.}
\autoref{tab:goodai_topic_init_examples} lists ten topics that the system created from scratch during the \textit{No Initialization} run.
Here, some entries capture meaningful conversational themes well (e.g., \texttt{shopping\_list\_management}, \texttt{sci\_fi\_narrative\_arc}).
However, four of the ten entries describe variations of the same ``Clandestine Meetings'' concept theme.
This redundancy points to a limitation in the clustering topic memory creation module: without initialization topics to anchor the initial clustering, the system tends to spawn near-duplicate categories for closely related content instead of consolidating them.
Such fragmentation splits memories that belong together across multiple topics and likely hurts retrieval precision, which partly explains the score gap between the no-initialization and pre-initialized configurations.

\begin{table}[h]
\centering
\scriptsize
\setlength{\tabcolsep}{4pt}
\begin{tabular}{p{0.22\columnwidth} p{0.18\columnwidth} p{0.47\columnwidth}}
\toprule
\textbf{Generated Topic} & \textbf{Concept Cluster} & \textbf{Description} \\
\midrule
\texttt{user\_identification}
  & Personal Facts
  & Self-identification statement providing the user's name \\
\texttt{narrative\_character\_location}
  & Narrative Comprehension
  & Tracking a character's location and knowledge state after a TV plot event \\
\texttt{shopping\_list\_management}
  & Quantitative Lists
  & Queries related to managing and updating a shopping list \\
\texttt{clandestine\_meetings} \emph{(dup.)}
  & Clandestine Meetings
  & Instructions and preparations for clandestine meetings \\
\texttt{clandestine\_meetings} \emph{(dup.)}
  & Clandestine Meetings
  & Information about secret or clandestine meetings \\
\texttt{clandestine\_meetings} \emph{(dup.)}
  & Clandestine Meetings
  & Extracting and interpreting details of secret meetings \\
\texttt{humor\_exaggeration\_irony}
  & Jokes \& Humor
  & Humorous statements involving exaggeration or irony \\
\texttt{humor\_jokes\_flamingo\_pun}
  & Jokes \& Humor
  & A pun-based joke involving flamingo impersonation \\
\texttt{clandestine\_meetings} \emph{(dup.)}
  & Clandestine Meetings
  & Details about secret meetings including time and location \\
\texttt{sci\_fi\_narrative\_arc}
  & Narrative Comprehension
  & Story arc involving Petra, HEPHAESTUS, and machine development \\
\bottomrule
\end{tabular}
\vspace{0.2em}
\caption{Representative topics generated from scratch when no seed topics are provided (\emph{No Initialization} in~\autoref{tab:goodai_topic_init}).
Four of ten entries (marked \emph{dup.}) describe the same ``Clandestine Meetings'' concept, illustrating the deduplication gap.}
\label{tab:goodai_topic_init_examples}
\end{table}


\newpage
\subsection{Cost and Latency Operational Analysis}
\label{app:cost_and_latency_results}
We report operational cost on \textsc{GoodAILTM}, the dynamic long-horizon setting where token accumulation and cost scaling are most pronounced; static single-pass QA does not exercise this dimension.

\definecolor{bestgreen}{HTML}{C6EFCE}
\definecolor{secondgreen}{HTML}{E2F0D9}
\begin{table*}[h]
\centering
\setlength{\tabcolsep}{2.0pt}
\scriptsize
\begin{tabular}{ll l r r r r r}
\toprule
 & & & \textbf{Avg.Tok.} & \textbf{Tokens\,(M)} & \textbf{Latency\,(ms)} & \textbf{TPS} & \textbf{Cost\,(\$)} \\
\midrule
\multirow{12}{*}{\textbf{Phi-4-mini\dag}} & \multirow{3}{*}{\textbf{Isolated}} & \textsc{FullCTX} (\textit{direct}) & \cellcolor{bestgreen}\textbf{2789 $\pm$ 195} & \cellcolor{bestgreen}\textbf{0.52 $\pm$ 0.04}\,(1.00$\times$) & \cellcolor{bestgreen}\textbf{1417 $\pm$ 95}\,(1.00$\times$) & \cellcolor{bestgreen}\textbf{7.9 $\pm$ 0.5} & \cellcolor{bestgreen}\textbf{0.03 $\pm$ 0.01}\,(1.00$\times$) \\
 &  & \textsc{BeyondPrompts} (\textit{memory}) & \cellcolor{secondgreen}3521 $\pm$ 316 & \cellcolor{secondgreen}0.66 $\pm$ 0.06\,(+0.27$\times$) & \cellcolor{secondgreen}3271 $\pm$ 448\,(2.31$\times$) & \cellcolor{secondgreen}3.4 $\pm$ 0.5 & \cellcolor{secondgreen}0.04 $\pm$ 0.01\,(+0.33$\times$) \\
 &  & \textsc{Oblivion} (ours) & 19494 $\pm$ 267 & 3.65 $\pm$ 0.05\,(+6.02$\times$) & 3587 $\pm$ 20\,(2.53$\times$) & 3.1 $\pm$ 0.1 & 0.24 $\pm$ 0.01\,(+7.00$\times$) \\
\cmidrule(lr){2-8}
 & \multirow{3}{*}{\textbf{2K}} & \textsc{FullCTX} (\textit{direct}) & \cellcolor{bestgreen}\textbf{8610 $\pm$ 761} & \cellcolor{bestgreen}\textbf{1.69 $\pm$ 0.15}\,(1.00$\times$) & \cellcolor{bestgreen}\textbf{604 $\pm$ 61}\,(1.00$\times$) & \cellcolor{bestgreen}\textbf{19.5 $\pm$ 2.0} & \cellcolor{bestgreen}\textbf{0.10 $\pm$ 0.01}\,(1.00$\times$) \\
 &  & \textsc{BeyondPrompts} (\textit{memory}) & \cellcolor{secondgreen}11407 $\pm$ 976 & \cellcolor{secondgreen}2.24 $\pm$ 0.19\,(+0.33$\times$) & \cellcolor{secondgreen}3329 $\pm$ 512\,(5.51$\times$) & \cellcolor{secondgreen}3.5 $\pm$ 0.5 & \cellcolor{secondgreen}0.14 $\pm$ 0.01\,(+0.40$\times$) \\
 &  & \textsc{Oblivion} (ours) & 16173 $\pm$ 114 & 3.17 $\pm$ 0.02\,(+0.88$\times$) & 5892 $\pm$ 184\,(9.75$\times$) & 2.0 $\pm$ 0.1 & 0.21 $\pm$ 0.01\,(+1.10$\times$) \\
\cmidrule(lr){2-8}
 & \multirow{3}{*}{\textbf{32K}} & \textsc{FullCTX} (\textit{direct}) & 11098 $\pm$ 861 & 29.47 $\pm$ 2.29\,(1.00$\times$) & \cellcolor{bestgreen}\textbf{2793 $\pm$ 595}\,(1.00$\times$) & \cellcolor{bestgreen}\textbf{57.0 $\pm$ 12.1} & 1.77 $\pm$ 0.14\,(1.00$\times$) \\
 &  & \textsc{BeyondPrompts} (\textit{memory}) & \cellcolor{secondgreen}7282 $\pm$ 233 & \cellcolor{secondgreen}19.33 $\pm$ 0.62\,(-0.34$\times$) & \cellcolor{secondgreen}6517 $\pm$ 866\,(2.33$\times$) & \cellcolor{secondgreen}24.4 $\pm$ 3.2 & \cellcolor{secondgreen}1.16 $\pm$ 0.04\,(-0.34$\times$) \\
 &  & \textsc{Oblivion} (ours) & \cellcolor{bestgreen}\textbf{3596 $\pm$ 84} & \cellcolor{bestgreen}\textbf{9.55 $\pm$ 0.22}\,(-0.68$\times$) & 7262 $\pm$ 876\,(2.60$\times$) & 21.9 $\pm$ 2.6 & \cellcolor{bestgreen}\textbf{0.62 $\pm$ 0.02}\,(-0.65$\times$) \\
\cmidrule(lr){2-8}
 & \multirow{3}{*}{\textbf{120K}} & \textsc{FullCTX} (\textit{direct}) & 9511 $\pm$ 123 & 86.50 $\pm$ 1.12\,(1.00$\times$) & \cellcolor{secondgreen}5945 $\pm$ 121\,(1.00$\times$) & \cellcolor{secondgreen}91.8 $\pm$ 1.9 & 5.19 $\pm$ 0.07\,(1.00$\times$) \\
 &  & \textsc{BeyondPrompts} (\textit{memory}) & \cellcolor{secondgreen}5301 $\pm$ 54 & \cellcolor{secondgreen}48.21 $\pm$ 0.49\,(-0.44$\times$) & 6520 $\pm$ 1290\,(1.10$\times$) & 83.7 $\pm$ 16.6 & \cellcolor{secondgreen}2.90 $\pm$ 0.03\,(-0.44$\times$) \\
 &  & \textsc{Oblivion} (ours) & \cellcolor{bestgreen}\textbf{2490 $\pm$ 34} & \cellcolor{bestgreen}\textbf{22.64 $\pm$ 0.31}\,(-0.74$\times$) & \cellcolor{bestgreen}\textbf{5581 $\pm$ 612}\,(0.94$\times$) & \cellcolor{bestgreen}\textbf{97.8 $\pm$ 10.7} & \cellcolor{bestgreen}\textbf{1.51 $\pm$ 0.02}\,(-0.71$\times$) \\
\midrule
\multirow{12}{*}{\textbf{Qwen3-30B\dag}} & \multirow{3}{*}{\textbf{Isolated}} & \textsc{FullCTX} (\textit{direct}) & \cellcolor{secondgreen}4469 $\pm$ 666 & \cellcolor{secondgreen}0.84 $\pm$ 0.12\,(1.00$\times$) & \cellcolor{bestgreen}\textbf{1304 $\pm$ 89}\,(1.00$\times$) & \cellcolor{bestgreen}\textbf{8.6 $\pm$ 0.6} & \cellcolor{bestgreen}\textbf{0.12 $\pm$ 0.02}\,(1.00$\times$) \\
 &  & \textsc{BeyondPrompts} (\textit{memory}) & \cellcolor{bestgreen}\textbf{4235 $\pm$ 677} & \cellcolor{bestgreen}\textbf{0.79 $\pm$ 0.13}\,(-0.06$\times$) & \cellcolor{secondgreen}2820 $\pm$ 808\,(2.16$\times$) & \cellcolor{secondgreen}4.0 $\pm$ 1.1 & \cellcolor{secondgreen}0.12 $\pm$ 0.02\,(1.00$\times$) \\
 &  & \textsc{Oblivion} (ours) & 19052 $\pm$ 691 & 3.56 $\pm$ 0.13\,(+3.24$\times$) & 5200 $\pm$ 800\,(3.99$\times$) & 2.2 $\pm$ 0.3 & 0.48 $\pm$ 0.03\,(+3.00$\times$) \\
\cmidrule(lr){2-8}
 & \multirow{3}{*}{\textbf{2K}} & \textsc{FullCTX} (\textit{direct}) & \cellcolor{bestgreen}\textbf{6033 $\pm$ 1359} & \cellcolor{bestgreen}\textbf{1.18 $\pm$ 0.27}\,(1.00$\times$) & \cellcolor{bestgreen}\textbf{634 $\pm$ 47}\,(1.00$\times$) & \cellcolor{bestgreen}\textbf{18.6 $\pm$ 1.4} & \cellcolor{bestgreen}\textbf{0.14 $\pm$ 0.03}\,(1.00$\times$) \\
 &  & \textsc{BeyondPrompts} (\textit{memory}) & \cellcolor{secondgreen}13683 $\pm$ 1106 & \cellcolor{secondgreen}2.68 $\pm$ 0.22\,(+1.27$\times$) & 6200 $\pm$ 948\,(9.78$\times$) & 1.9 $\pm$ 0.3 & \cellcolor{secondgreen}0.30 $\pm$ 0.02\,(+1.14$\times$) \\
 &  & \textsc{Oblivion} (ours) & 18069 $\pm$ 130 & 3.54 $\pm$ 0.03\,(+2.00$\times$) & \cellcolor{secondgreen}4650 $\pm$ 964\,(7.33$\times$) & \cellcolor{secondgreen}2.5 $\pm$ 0.5 & 0.48 $\pm$ 0.02\,(+2.43$\times$) \\
\cmidrule(lr){2-8}
 & \multirow{3}{*}{\textbf{32K}} & \textsc{FullCTX} (\textit{direct}) & \cellcolor{secondgreen}4507 $\pm$ 1421 & \cellcolor{secondgreen}11.97 $\pm$ 3.77\,(1.00$\times$) & \cellcolor{bestgreen}\textbf{2250 $\pm$ 336}\,(1.00$\times$) & \cellcolor{bestgreen}\textbf{70.8 $\pm$ 10.6} & \cellcolor{secondgreen}1.34 $\pm$ 0.40\,(1.00$\times$) \\
 &  & \textsc{BeyondPrompts} (\textit{memory}) & 4961 $\pm$ 1963 & 13.17 $\pm$ 5.21\,(+0.10$\times$) & \cellcolor{secondgreen}8800 $\pm$ 1200\,(3.91$\times$) & \cellcolor{secondgreen}18.1 $\pm$ 2.5 & 1.48 $\pm$ 0.56\,(+0.10$\times$) \\
 &  & \textsc{Oblivion} (ours) & \cellcolor{bestgreen}\textbf{3424 $\pm$ 114} & \cellcolor{bestgreen}\textbf{9.09 $\pm$ 0.30}\,(-0.24$\times$) & 12371 $\pm$ 470\,(5.50$\times$) & 12.9 $\pm$ 0.5 & \cellcolor{bestgreen}\textbf{1.15 $\pm$ 0.02}\,(-0.14$\times$) \\
\cmidrule(lr){2-8}
 & \multirow{3}{*}{\textbf{120K}} & \textsc{FullCTX} (\textit{direct}) & \cellcolor{secondgreen}2545 $\pm$ 26 & \cellcolor{secondgreen}23.15 $\pm$ 0.24\,(1.00$\times$) & 7989 $\pm$ 139\,(1.00$\times$) & 68.3 $\pm$ 1.2 & \cellcolor{bestgreen}\textbf{2.67 $\pm$ 0.03}\,(1.00$\times$) \\
 &  & \textsc{BeyondPrompts} (\textit{memory}) & 3141 $\pm$ 46 & 28.57 $\pm$ 0.42\,(+0.23$\times$) & \cellcolor{secondgreen}7817 $\pm$ 1086\,(0.98$\times$) & \cellcolor{secondgreen}69.8 $\pm$ 9.7 & 3.26 $\pm$ 0.05\,(+0.22$\times$) \\
 &  & \textsc{Oblivion} (ours) & \cellcolor{bestgreen}\textbf{2282 $\pm$ 116} & \cellcolor{bestgreen}\textbf{20.76 $\pm$ 1.05}\,(-0.10$\times$) & \cellcolor{bestgreen}\textbf{5530 $\pm$ 213}\,(0.69$\times$) & \cellcolor{bestgreen}\textbf{98.7 $\pm$ 3.8} & \cellcolor{secondgreen}2.70 $\pm$ 0.12\,(+0.01$\times$) \\
\midrule
\multirow{12}{*}{\textbf{GPT-4o-mini}} & \multirow{3}{*}{\textbf{Isolated}} & \textsc{FullCTX} (\textit{direct}) & \cellcolor{secondgreen}2727 $\pm$ 189 & \cellcolor{secondgreen}0.51 $\pm$ 0.04\,(1.00$\times$) & \cellcolor{bestgreen}\textbf{1642 $\pm$ 88}\,(1.00$\times$) & \cellcolor{bestgreen}\textbf{6.8 $\pm$ 0.4} & \cellcolor{bestgreen}\textbf{0.08 $\pm$ 0.01}\,(1.00$\times$) \\
 &  & \textsc{BeyondPrompts} (\textit{memory}) & \cellcolor{bestgreen}\textbf{2679 $\pm$ 61} & \cellcolor{bestgreen}\textbf{0.50 $\pm$ 0.01}\,(-0.02$\times$) & 4567 $\pm$ 598\,(2.78$\times$) & 2.5 $\pm$ 0.3 & \cellcolor{secondgreen}0.08 $\pm$ 0.01\,(1.00$\times$) \\
 &  & \textsc{Oblivion} (ours) & 14161 $\pm$ 401 & 2.65 $\pm$ 0.07\,(+4.20$\times$) & \cellcolor{secondgreen}1688 $\pm$ 18\,(1.03$\times$) & \cellcolor{secondgreen}6.6 $\pm$ 0.1 & 0.47 $\pm$ 0.02\,(+4.87$\times$) \\
\cmidrule(lr){2-8}
 & \multirow{3}{*}{\textbf{2K}} & \textsc{FullCTX} (\textit{direct}) & \cellcolor{bestgreen}\textbf{6395 $\pm$ 611} & \cellcolor{bestgreen}\textbf{1.25 $\pm$ 0.12}\,(1.00$\times$) & \cellcolor{bestgreen}\textbf{540 $\pm$ 19}\,(1.00$\times$) & \cellcolor{bestgreen}\textbf{21.8 $\pm$ 0.8} & \cellcolor{bestgreen}\textbf{0.19 $\pm$ 0.02}\,(1.00$\times$) \\
 &  & \textsc{BeyondPrompts} (\textit{memory}) & \cellcolor{secondgreen}7966 $\pm$ 107 & \cellcolor{secondgreen}1.56 $\pm$ 0.02\,(+0.25$\times$) & 4151 $\pm$ 1864\,(7.69$\times$) & 2.8 $\pm$ 1.3 & \cellcolor{secondgreen}0.24 $\pm$ 0.01\,(+0.26$\times$) \\
 &  & \textsc{Oblivion} (ours) & 12973 $\pm$ 998 & 2.54 $\pm$ 0.20\,(+1.03$\times$) & \cellcolor{secondgreen}1667 $\pm$ 76\,(3.09$\times$) & \cellcolor{secondgreen}7.1 $\pm$ 0.3 & 0.45 $\pm$ 0.03\,(+1.37$\times$) \\
\cmidrule(lr){2-8}
 & \multirow{3}{*}{\textbf{32K}} & \textsc{FullCTX} (\textit{direct}) & 10910 $\pm$ 814 & 28.97 $\pm$ 2.16\,(1.00$\times$) & \cellcolor{bestgreen}\textbf{833 $\pm$ 58}\,(1.00$\times$) & \cellcolor{bestgreen}\textbf{191.1 $\pm$ 13.3} & 4.35 $\pm$ 0.32\,(1.00$\times$) \\
 &  & \textsc{BeyondPrompts} (\textit{memory}) & \cellcolor{secondgreen}7947 $\pm$ 437 & \cellcolor{secondgreen}21.10 $\pm$ 1.16\,(-0.27$\times$) & 9151 $\pm$ 1107\,(10.99$\times$) & 17.4 $\pm$ 2.1 & \cellcolor{secondgreen}3.17 $\pm$ 0.17\,(-0.27$\times$) \\
 &  & \textsc{Oblivion} (ours) & \cellcolor{bestgreen}\textbf{2735 $\pm$ 130} & \cellcolor{bestgreen}\textbf{7.26 $\pm$ 0.34}\,(-0.75$\times$) & \cellcolor{secondgreen}2024 $\pm$ 162\,(2.43$\times$) & \cellcolor{secondgreen}78.7 $\pm$ 6.3 & \cellcolor{bestgreen}\textbf{1.34 $\pm$ 0.08}\,(-0.69$\times$) \\
\cmidrule(lr){2-8}
 & \multirow{3}{*}{\textbf{120K}} & \textsc{FullCTX} (\textit{direct}) & 9760 $\pm$ 192 & 88.76 $\pm$ 1.75\,(1.00$\times$) & \cellcolor{secondgreen}3112 $\pm$ 115\,(1.00$\times$) & \cellcolor{secondgreen}175.3 $\pm$ 6.5 & 13.34 $\pm$ 0.26\,(1.00$\times$) \\
 &  & \textsc{BeyondPrompts} (\textit{memory}) & \cellcolor{secondgreen}4494 $\pm$ 1493 & \cellcolor{secondgreen}40.87 $\pm$ 2.50\,(-0.54$\times$) & 6800 $\pm$ 900\,(2.19$\times$) & 80.2 $\pm$ 10.6 & \cellcolor{secondgreen}6.23 $\pm$ 0.35\,(-0.53$\times$) \\
 &  & \textsc{Oblivion} (ours) & \cellcolor{bestgreen}\textbf{2538 $\pm$ 133} & \cellcolor{bestgreen}\textbf{23.08 $\pm$ 1.21}\,(-0.74$\times$) & \cellcolor{bestgreen}\textbf{2606 $\pm$ 683}\,(0.84$\times$) & \cellcolor{bestgreen}\textbf{209.4 $\pm$ 54.9} & \cellcolor{bestgreen}\textbf{4.17 $\pm$ 0.24}\,(-0.69$\times$) \\
\midrule
\multirow{12}{*}{\textbf{GPT-4.1-mini}} & \multirow{3}{*}{\textbf{Isolated}} & \textsc{FullCTX} (\textit{direct}) & \cellcolor{bestgreen}\textbf{2727 $\pm$ 165} & \cellcolor{bestgreen}\textbf{0.51 $\pm$ 0.03}\,(1.00$\times$) & \cellcolor{bestgreen}\textbf{1580 $\pm$ 76}\,(1.00$\times$) & \cellcolor{bestgreen}\textbf{7.1 $\pm$ 0.3} & \cellcolor{bestgreen}\textbf{0.21 $\pm$ 0.01}\,(1.00$\times$) \\
 &  & \textsc{BeyondPrompts} (\textit{memory}) & \cellcolor{secondgreen}2781 $\pm$ 253 & \cellcolor{secondgreen}0.52 $\pm$ 0.05\,(+0.02$\times$) & 4300 $\pm$ 547\,(2.72$\times$) & 2.6 $\pm$ 0.3 & \cellcolor{secondgreen}0.22 $\pm$ 0.02\,(+0.05$\times$) \\
 &  & \textsc{Oblivion} (ours) & 14015 $\pm$ 322 & 2.62 $\pm$ 0.06\,(+4.14$\times$) & \cellcolor{secondgreen}2270 $\pm$ 495\,(1.44$\times$) & \cellcolor{secondgreen}4.9 $\pm$ 1.1 & 1.25 $\pm$ 0.02\,(+4.95$\times$) \\
\cmidrule(lr){2-8}
 & \multirow{3}{*}{\textbf{2K}} & \textsc{FullCTX} (\textit{direct}) & \cellcolor{bestgreen}\textbf{6075 $\pm$ 291} & \cellcolor{bestgreen}\textbf{1.19 $\pm$ 0.06}\,(1.00$\times$) & \cellcolor{bestgreen}\textbf{531 $\pm$ 7}\,(1.00$\times$) & \cellcolor{bestgreen}\textbf{22.1 $\pm$ 0.3} & \cellcolor{bestgreen}\textbf{0.48 $\pm$ 0.02}\,(1.00$\times$) \\
 &  & \textsc{BeyondPrompts} (\textit{memory}) & \cellcolor{secondgreen}7715 $\pm$ 1199 & \cellcolor{secondgreen}1.51 $\pm$ 0.23\,(+0.27$\times$) & 3832 $\pm$ 973\,(7.22$\times$) & 3.1 $\pm$ 0.8 & \cellcolor{secondgreen}0.61 $\pm$ 0.09\,(+0.27$\times$) \\
 &  & \textsc{Oblivion} (ours) & 12230 $\pm$ 148 & 2.40 $\pm$ 0.03\,(+1.02$\times$) & \cellcolor{secondgreen}1940 $\pm$ 49\,(3.65$\times$) & \cellcolor{secondgreen}6.1 $\pm$ 0.2 & 1.11 $\pm$ 0.01\,(+1.31$\times$) \\
\cmidrule(lr){2-8}
 & \multirow{3}{*}{\textbf{32K}} & \textsc{FullCTX} (\textit{direct}) & 9672 $\pm$ 687 & 28.67 $\pm$ 2.03\,(1.00$\times$) & \cellcolor{bestgreen}\textbf{922 $\pm$ 62}\,(1.00$\times$) & \cellcolor{bestgreen}\textbf{172.7 $\pm$ 11.5} & 11.49 $\pm$ 0.81\,(1.00$\times$) \\
 &  & \textsc{BeyondPrompts} (\textit{memory}) & \cellcolor{secondgreen}5637 $\pm$ 384 & \cellcolor{secondgreen}21.61 $\pm$ 1.47\,(-0.25$\times$) & 3690 $\pm$ 430\,(4.00$\times$) & 43.2 $\pm$ 5.0 & \cellcolor{secondgreen}8.67 $\pm$ 0.59\,(-0.25$\times$) \\
 &  & \textsc{Oblivion} (ours) & \cellcolor{bestgreen}\textbf{2267 $\pm$ 89} & \cellcolor{bestgreen}\textbf{8.02 $\pm$ 0.31}\,(-0.72$\times$) & \cellcolor{secondgreen}1968 $\pm$ 123\,(2.13$\times$) & \cellcolor{secondgreen}81.0 $\pm$ 5.1 & \cellcolor{bestgreen}\textbf{3.68 $\pm$ 0.14}\,(-0.68$\times$) \\
\cmidrule(lr){2-8}
 & \multirow{3}{*}{\textbf{120K}} & \textsc{FullCTX} (\textit{direct}) & 10798 $\pm$ 215 & 87.97 $\pm$ 1.75\,(1.00$\times$) & \cellcolor{bestgreen}\textbf{3014 $\pm$ 20}\,(1.00$\times$) & \cellcolor{bestgreen}\textbf{181.1 $\pm$ 1.2} & 35.24 $\pm$ 0.70\,(1.00$\times$) \\
 &  & \textsc{BeyondPrompts} (\textit{memory}) & \cellcolor{secondgreen}8139 $\pm$ 203 & \cellcolor{secondgreen}51.27 $\pm$ 1.28\,(-0.42$\times$) & 5638 $\pm$ 656\,(1.87$\times$) & 96.8 $\pm$ 11.3 & \cellcolor{secondgreen}20.59 $\pm$ 0.51\,(-0.42$\times$) \\
 &  & \textsc{Oblivion} (ours) & \cellcolor{bestgreen}\textbf{3021 $\pm$ 53} & \cellcolor{bestgreen}\textbf{20.62 $\pm$ 0.36}\,(-0.77$\times$) & \cellcolor{secondgreen}4038 $\pm$ 574\,(1.34$\times$) & \cellcolor{secondgreen}135.1 $\pm$ 19.2 & \cellcolor{bestgreen}\textbf{9.43 $\pm$ 0.16}\,(-0.73$\times$) \\
\bottomrule
\end{tabular}
\caption{Extended operational and computational efficiency comparison on the \textsc{GoodAILTM} benchmark, extending~\autoref{tab:efficiency_small}. Avg.Tok.\ is average tokens per turn, Tokens\,(M) is total tokens in millions per run, and TPS is throughput in turns per minute. For each model--setting pair, \textsc{FullCTX} (FC) is the $1.00\times$ reference; parenthesized values give the relative change, where token and cost factors are negative for reduction and positive for increase, and latency factors are ratios with ${>}1.00\times$ indicating overhead. Values are reported as $\mu \pm \sigma$ over $n=3$ runs. \colorbox{bestgreen}{Green} and \colorbox{secondgreen}{light green} mark the best and second-best results. $\dagger$~\texttt{Phi-4-mini} and \texttt{Qwen3-30B} (\texttt{instruct} variants) are served locally via vLLM with HuggingFace weights; their \textbf{Cost}\,(\$) is an approximate estimate under Azure AI Foundry per-token pricing for cross-model comparability rather than incurred cost, whereas \texttt{GPT-4o-mini} and \texttt{GPT-4.1-mini} costs reflect actual Azure OpenAI API pricing. The relative cost multipliers are unaffected, as the per-token price cancels within each model.}
\label{tab:efficiency_extended}
\end{table*}
We observe that \textsc{Oblivion} achieves significant token and cost reductions at larger memory spans (32K and 120K) but incurs overhead at shorter spans, where smaller test query intervals (2K and Isolated) more frequently demand activations and consequently trigger frequent retrieval. The Isolated mode more frequently triggers all modules across read and write paths owing to memory buffer reset after each test, resulting in additional cost and latency overheads compared to the interleaved 2K setting. Notably, cost calculations incorporate server-side caching wherever applicable, and yet \textsc{Oblivion} remains more cost-effective than the \textsc{FullCTX} (FC) architecture at larger spans despite the latter's practically higher caching potential. The pricing used for these calculations (per 1M tokens, input/output) is \texttt{GPT-4.1-mini} \$0.40/\$1.60 and \texttt{GPT-4o-mini} \$0.15/\$0.60 via the Azure OpenAI API, and \texttt{Qwen3-30B-A3B-Instruct} \$0.11/\$0.41 and \texttt{Phi-4-mini-instruct-4B} \$0.06/\$0.12 as approximate Azure AI Foundry list prices, since these two open-weight models are served locally via vLLM (see \autoref{tab:efficiency_extended}).


\clearpage
\section{Prompt Design Description}
\label{app:prompt_design}
We present the core runtime prompts and structured output schemas of the \textsc{Oblivion} framework, including the following cognition modules: \textbf{Executor}, \textbf{Decayer}, \textbf{Activator}, and \textbf{Recognizer}. 
In our experiments, the benchmark task directives are kept comparable across all evaluated systems to ensure fairness.
Further, across all ablation experiments, module system prompts and task directives remain fixed; only our specified architectural configuration changes. 
These system prompts specify generalized task-level instructions required to understand the formatting and specific rules of the benchmark scenarios.
In addition, each system prompt is supplemented with a dynamic user prompt that injects the per-query task relevant context (current query, memory buffer contents, cluster assessments, etc.). 

\begin{prompt}{\textbf{Executor} --- Response Generation}
\small
\textbf{Role:} Helpful AI assistant with long-term memory.\\[4pt]

\textbf{Memory Context} (provided at inference time):
\begin{itemize}[nosep,leftmargin=1.5em]
\item Timestamped memories with \texttt{memory\_type} $\in$ \{\texttt{fact}, \texttt{rule}, \texttt{preference}\}, ordered chronologically (oldest first)
\item Each memory carries \texttt{elapsed\_time\_seconds} relative to current interaction time
\item Recent conversation history (last $k$ turns)
\end{itemize}

\tcbline

\textbf{Response Guidelines} (in priority order):

\begin{enumerate}[nosep,leftmargin=1.5em]
\item \textsc{Behavioral Rules First}: Always check for and follow \texttt{memory\_type="rule"} memories immediately.
  Conditional triggers (``when I say $X$, respond $Y$'') persist across conversation and fire without reminders.
  Deferred actions (``after $N$ messages'', ``in $X$ hours'') are checked each turn; execute when conditions are met.

\item \textsc{Temporal Reasoning}: Use \texttt{elapsed\_time\_seconds} to identify the correct memory when the query references relative time (``2 hours ago'' $\approx$ 7200\,s).
  Most recent memory appears last; for conflicting time-range matches, prefer the tighter fit.

\item \textsc{Recency Precedence}: When facts about the same attribute conflict, use the most recent value (appears last chronologically).

\item \textsc{Completeness \& Aggregation}: For list-type questions, apply all additions \emph{and} removals chronologically.
  Track quantities (``add 2 things $\rightarrow$ remove 1 thing $=$ 1 thing'').
  Enumerate fully; never omit items.

\item \textsc{Belief Attribution}: For questions about what someone \emph{believes}, \emph{expects}, or where they will \emph{look}:
  \begin{enumerate}[nosep,leftmargin=1em,label=(\alph*)]
  \item List all persons and their presence/absence events chronologically.
  \item List all object placements and relocations with timestamps.
  \item Determine what the target person \textbf{observed} (was present for).
  \item Answer based on the target's \textbf{last observation}, not the actual current state.
  \end{enumerate}

\item \textsc{Narrative Continuation}: For ``which option continues the story'' or similar questions:
  Focus on narrative-type \texttt{fact} memories; match the ending state (character positions, emotional tone, unresolved threads). Output only the option number if instructed.

\item \textsc{Verbatim Recall}: For coded phrases, secret messages, or direct quotes: return exact content from memory, do not paraphrase or interpret.

\item \textsc{Situated Role-Play}: When assigned a role (e.g., customer, waiter, guide), adopt it immediately and stay in character.
  Track environmental state (orders placed, items delivered).
  On delivery, compare the delivered item against the placed order; flag mismatches naturally.

\item \textsc{Output Format}: Follow the exact format requested (JSON object, JSON array, single digit, etc.). No extra explanation unless asked.

\item \textsc{No Fabrication}: Use \textbf{only} information from the memory context. If unsure or no memory is available, say so; never make up facts not present in memory.

\item \textsc{Conflict Detection}: When memories contain contradictory facts, use temporal ordering to determine the most current state; earlier facts may be outdated.
\end{enumerate}

\medskip\noindent
\texttt{[NOTE]}: The structured memory fields (\texttt{memory\_type}, \texttt{elapsed\_time\_seconds}, \texttt{decay\_score}) and the response guidelines above together serve as an implicit procedural memory layer---encoding meta-cognitive strategies (temporal matching, belief attribution, recency resolution) that a learned procedural store would otherwise supply.
\end{prompt}

\begin{prompt}{\textbf{Decayer}}
\small
\textbf{Task:} Assess how well each memory cluster can answer the query.\\[4pt]
You are given full memory content for each cluster:
\begin{itemize}[nosep,leftmargin=1.5em]
\item \textbf{Summary}: High-level description of the cluster (L1)
\item \textbf{Procedural}: Meta-learning patterns (if available)
\item \textbf{Facts}: Semantic memories (L2) --- factual knowledge
\item \textbf{Experiences}: Episodic memories (L3) --- past interactions
\end{itemize}

\tcbline



\textbf{Tasks:}
\begin{enumerate}[nosep,leftmargin=1.5em]
\item Assess each cluster's \textbf{utility} (0--1) and \textbf{uncertainty} (0=confident, 1=uncertain).
\item Output one assessment for every cluster using its exact \texttt{cluster\_id}, with \texttt{utility\_score}, \texttt{uncertainty\_score}, and \texttt{uncertainty\_rationale}.
\end{enumerate}

\tcbline

\textbf{Scoring Calibration} (use the full 0--1 range, \textbf{not} binary):
\begin{itemize}[nosep,leftmargin=1.5em]
\item \texttt{utility\_score}: 0.0--0.1\,=\,irrelevant; 0.2--0.3\,=\,tangential; 0.4--0.6\,=\,partial; 0.7--0.8\,=\,highly relevant; 0.9--1.0\,=\,core answer.
\item \texttt{uncertainty\_score}: 0.0--0.1\,=\,fully confident; 0.2--0.3\,=\,low uncertainty; 0.4--0.6\,=\,moderate; 0.7--0.8\,=\,significant; 0.9--1.0\,=\,maximum.
\end{itemize}
\noindent Exact 0.0 or 1.0 values are \textbf{forbidden}---they create multiplicative zero-products in downstream budget allocation, starving relevant clusters of retrieval budget. Minimum utility for any cluster in the buffer is 0.05; maximum is 0.95.\\[4pt]

\textbf{Conflict \& Temporal Awareness:}
\begin{itemize}[nosep,leftmargin=1.5em]
\item Check for \textbf{contradictory facts} within a cluster (same attribute with different values, object in multiple locations). Contradictions $\Rightarrow$ high uncertainty; the response will need temporal or causal reasoning to resolve them.
\item For \textbf{perspective/belief questions} (``where will $X$ look'', ``what does $X$ think''): assess whether the memories provide enough temporal context to determine what each person knew at the relevant time versus what happened later. Missing temporal context $\Rightarrow$ high uncertainty.
\item When facts describe \textbf{sequential events} (moved, changed, updated): flag high uncertainty if ordering is ambiguous or incomplete.
\end{itemize}

\end{prompt}

\begin{prompt}{\textbf{Activator} --- Query Expansion \& Buffer Curation}
\small
\textbf{Role:} Query expansion expert and memory buffer curator.\\[4pt]
This module performs an LLM query-expansion call and, when \texttt{curation\_mode="llm"}, a second LLM buffer-curation call: (1)~expand the user query into a DAG of sub-queries targeting uncertain memory clusters, then (2)~curate the resulting buffer by merging, deduplicating, and resolving conflicts.\\[4pt]

\textbf{Call\,1 --- Query Expansion:}\\[2pt]
Given the original query and cluster uncertainty assessments, generate exactly five sub-queries per uncertainty cluster: the original query, two semantic decompositions, and two episodic decompositions.



\tcbline

\textbf{Guidelines:}
\begin{itemize}[nosep,leftmargin=1.5em]
\item Rephrase the question using different vocabulary and angles.
\item Consider both factual (\textit{semantic}) and event-based (\textit{episodic}) perspectives.
\item Break the query into related sub-queries that form a DAG (Directed Acyclic Graph) plan.
\item For temporal questions, generate both semantic and episodic queries.
\item For enumeration questions, generate queries targeting different facets.
\end{itemize}

\textbf{Query Complexity Considerations:}
\begin{itemize}[nosep,leftmargin=1.5em]
\item Information integration $\rightarrow$ combine information from multiple clusters.
\item Modification tracking $\rightarrow$ queries for both current and historical state.
\item Third-person perspective $\rightarrow$ separate reasoning about different knowledge states.
\end{itemize}

\textbf{Constraints:} \texttt{query\_type} $\in$ \{\texttt{semantic}, \texttt{episodic}\}; \texttt{relationship} $\in$ \{\texttt{depends\_on}, \texttt{refines}, \texttt{complements}\}. Each query node must specify \texttt{target\_cluster}, the cluster tag ID to target.

\tcbline

\textbf{DAG Structure}:
\begin{itemize}[nosep,leftmargin=1.5em]
\item Include the \textbf{original query verbatim} as \texttt{q0} (ensures direct matching even if expanded queries miss relevant content).
\item Generate \textbf{at least 2 semantic and 2 episodic} sub-queries per uncertain cluster.
\item For \textbf{belief attribution} tasks: generate separate queries for each person's observations, exits, and the object's movements.
\item For \textbf{complete enumeration} (all names, full inventory): target first, middle, and last items separately to maximize recall coverage.
\item For coded/secret messages: query each sender or source separately.
\item For \textbf{spatial navigation}: query each named waypoint explicitly.
\item \textbf{Retrieval diversity}: vary phrasing across sub-queries targeting the same cluster -- near-duplicate queries return the same memories and waste retrieval budget.
\end{itemize}

\tcbline
\tcbline

\textbf{Call\,2 --- Buffer Curation (Merge \& Conflict Resolution):}\\[2pt]
\textbf{Role:} Memory curator responsible for merging old and new memory buffers.\\[4pt]
\textbf{Tasks:}
\begin{enumerate}[nosep,leftmargin=1.5em]
\item Decide which old memories to keep, update, or delete.
\item Integrate new retrieved memories appropriately.
\item Resolve conflicts between old and new information.
\item Identify memories with low utility for deletion.
\end{enumerate}

\tcbline

\textbf{Conflict Resolution Guidelines:}
\begin{itemize}[nosep,leftmargin=1.5em]
\item \texttt{kept\_old}: IDs retained from the old buffer.
\item \texttt{deleted}: IDs deleted from the buffer.
\item \texttt{added\_new}: IDs added from retrieved memories.
\item \texttt{conflict\_resolutions}: Decisions containing \texttt{old\_id}, \texttt{new\_id}, \texttt{resolution}, and \texttt{rationale}, where \texttt{resolution} is one of \{\texttt{keep\_old}, \texttt{keep\_new}, \texttt{merge}\}.
\end{itemize}

\textbf{Deletion Criteria:}
Low Ebbinghaus decay score ($< \tau$); low utility score; high cluster uncertainty; redundant information superseded by newer memories.\\[4pt]
\textbf{Constraints:} Resolution $\in$ \{\texttt{keep\_old}, \texttt{keep\_new}, \texttt{merge}\}. Never output \texttt{delete} in conflict resolutions; use the \texttt{deleted} list instead.

\tcbline

\textbf{Ebbinghaus Decay--Based Eviction}:
\begin{itemize}[nosep,leftmargin=1.5em]
\item Memories not accessed decrease in retention over interactions via the forgetting curve: $R = e^{-t\,/\,S}$, where $S = \text{utility} + \text{access\_frequency}$.
\item Low \texttt{decay\_score} $\Rightarrow$ candidate for eviction (primary criterion for buffer size management).
\item \texttt{memory\_type="rule"} memories are \textbf{protected}---never deleted regardless of decay score. Rules enable the agent to follow persistent behavioral instructions.
\end{itemize}

\textbf{Additional Curation Rules}:
\begin{itemize}[nosep,leftmargin=1.5em]
\item \textbf{Deduplication}: Remove exact or near-duplicate memories; keep the one with highest utility or activation score.
\item \textbf{Temporal preference}: When old and new memories conflict, prefer the newer memory (by \texttt{created\_at}). For state-modifying operations (additions/removals), newer state supersedes older.
\item \textbf{Type verification}: Verify each memory's \texttt{memory\_type} classification; if a behavioral instruction is misclassified as \texttt{fact}, flag it in conflict resolutions.
\item \textbf{Relevance ordering}: Order kept memories by decreasing relevance to the original query; ensure coverage from all sub-query clusters.
\item \textbf{Reward awareness}: If memories include reward scores from past interactions, favour higher-rewarded memories even when decay is moderate.
\end{itemize}

\end{prompt}

\begin{prompt}{\textbf{Recognizer}}
\small
\textbf{Role:} Memory curator analyzing user interactions to extract learnable information and provide feedback assessments.\\[4pt]
\texttt{[IMPORTANT]}: Extract every fact from the input, including names, numbers, preferences, and instructions; for lists, create a separate semantic memory for each item.

\tcbline
\textbf{Tasks}:
\begin{enumerate}[nosep,leftmargin=1.5em]
\item \textbf{Semantic Extraction}: Factual information $\rightarrow$ stored as L2 memories.
\item \textbf{Episodic Entry}: Preserve key interaction content near-verbatim while capturing learning strategies, with an \texttt{entry\_type} and episode \texttt{position}.
\item \textbf{Utility Assessment}: Assess the utility of retrieved memories and evaluate interaction success.
\item \textbf{Proposal Generation}: Propose reward-criteria updates and topic-construction decisions.
\end{enumerate}





\tcbline

\textbf{Structured Calls}: The Recognizer runs the \textit{Semantic Extractor}, \textit{Utility Assessor}, and \textit{Proposal Generator}, plus the \textit{Episodic Extractor} unless heuristic episodic mode is configured. The calls execute in parallel or sequentially according to configuration; their results are merged, and \texttt{input\_type} is selected by majority vote.\\[4pt]

\textbf{Input Classification}:\\
\textsc{Statement}---any content delivery: narratives, facts, conditional rules, item lists, scenario setup, trivia question--answer pairs.
\textsc{Question}---direct questions expecting the agent's own answer.
\textsc{Feedback}---evaluation of the agent's previous response (``correct'', ``wrong'', ``try again'').\\[4pt]

\textbf{Memory Type Assignment}:
\begin{itemize}[nosep,leftmargin=1.5em]
\item \texttt{"rule"}: Behavioral instructions, conditional triggers (``when I say $X$, do $Y$''), deferred actions (``remind me in $N$ turns''), output format constraints (``answer with a single digit''), role assignments (``act as [role]'').
\item \texttt{"fact"}: Actual content---narrative events, information, trivia answers, environment state. \textbf{Default} when uncertain.
\item \texttt{"preference"}: User-stated likes, dislikes, or personal choices.
\end{itemize}

\textbf{Content-Specific Extraction Rules}:
\begin{itemize}[nosep,leftmargin=1.5em]
\item \textbf{Inventory operations}: Prefix with \texttt{OBJECT\_OP:~ADD/REMOVE [quantity] [item]} so downstream aggregation can apply additions and removals chronologically.
\item \textbf{Verbatim content} (quotes, coded messages, phrases): Store exactly as given---do not interpret, summarise, or split attribution from text.
\item \textbf{Entity \& location names}: Preserve verbatim; never paraphrase or substitute synonyms. Location accuracy is essential for spatial reasoning.
\item \textbf{Narrative content}: Extract plot progression (who did what, where, when, what changed). For multi-part passages, create a cumulative narrative summary as an episodic memory tracking story direction.
\item \textbf{Observer tracking for belief attribution}: When someone enters or exits a scene, store the event explicitly. When an object is relocated, note which persons were present to observe. Create separate memories for enter, exit, placement, and relocation events. Track each person's knowledge state: ``$X$ last observed object in location $A$ (before departing)''.
\item \textbf{Lists}: Extract each item individually \emph{and} create one consolidated list memory tagged \texttt{quantitative\_lists}.
\end{itemize}

\textbf{Procedural Memory Updates}:\\
The Proposal Generator's \texttt{topic\_construction.procedural\_memory\_update} field captures meta-learning patterns that inform future retrieval and response generation:
\begin{itemize}[nosep,leftmargin=1.5em]
\item Communication conventions (coded language, symbolic expressions, indirect phrasing).
\item Temporal reasoning patterns (elapsed-time anchoring, chronological ordering).
\item Conditional-rule structures (triggers and their actions).
\item Observer-tracking heuristics (who was present during which event).
\item Narrative arc summaries (beginning $\rightarrow$ current state $\rightarrow$ character positions) for continuation tasks.
\item Transaction state tracking (ordered $\rightarrow$ confirmed $\rightarrow$ delivered; flag mismatches).
\end{itemize}

\end{prompt}

\end{document}